\documentclass{article} 
\usepackage{iclr2026_conference,times}

\usepackage{amsmath,amsfonts,bm}

\def\eqref#1{equation~\ref{#1}}

\def\1{\bm{1}}

\DeclareMathAlphabet{\mathsfit}{\encodingdefault}{\sfdefault}{m}{sl}
\SetMathAlphabet{\mathsfit}{bold}{\encodingdefault}{\sfdefault}{bx}{n}

\usepackage{calc}
\usepackage{enumitem}
\usepackage{hyperref}
\usepackage{url}
\usepackage{subcaption}
\usepackage{ulem}
\usepackage{wrapfig}
\usepackage{caption}
\usepackage{graphicx}
\usepackage{multirow}
\usepackage{array}
\usepackage{booktabs}
\usepackage[table]{xcolor}
\usepackage{colortbl}
\usepackage{arydshln}
\usepackage{tabularx}
\usepackage{caption}
\usepackage{xcolor}
\usepackage{amsthm}

\definecolor{myred}{HTML}{F67280}
\definecolor{myblue}{HTML}{31ACD0}
\definecolor{mygreen}{HTML}{E0F9E0}
\definecolor{mypink}{HTML}{FFE8E8}
\hypersetup{
    colorlinks=true,
    linkcolor=myred,     
    citecolor=myblue,      
}
\title{MMR-Life: Piecing Together Real-life Scenes for Multimodal Multi-image Reasoning}

\author{Jiachun Li\textsuperscript{1,2}\thanks{Equal contribution.} , Shaoping Huang\textsuperscript{2,3}\footnotemark[1] , Zhuoran Jin\textsuperscript{1,2}\thanks{Corresponding authors.} ,  Chenlong Zhang\textsuperscript{1,2}, Pengfei Cao\textsuperscript{1,2},\\  \textbf{Yubo Chen\textsuperscript{1,2}, Kang Liu\textsuperscript{1,2}, Jun Zhao\textsuperscript{1,2}}\footnotemark[2] \\ \textsuperscript{1}School of Artificial Intelligence, University of Chinese Academy of Sciences \\ \textsuperscript{2}The Key Laboratory of Cognition and Decision Intelligence for Complex Systems, \\ Institute of Automation, Chinese Academy of Sciences  \\
\textsuperscript{3}School of Advanced Interdisciplinary Sciences,  University of Chinese Academy of Sciences \\
\footnotesize{\texttt{\{jiachun.li, pengfei.cao, zhuoran.jin\}@nlpr.ia.ac.cn }} \\
\textbf{\url{https://mmr-life-bench.github.io/}}
}

\iclrfinalcopy 
\begin{document}

\maketitle
\vspace{-5pt}
\begin{abstract}
Recent progress in the reasoning capabilities of multimodal large language models (MLLMs) has empowered them to address more complex tasks such as scientific analysis and mathematical reasoning.  
Despite their promise, MLLMs' reasoning abilities across different scenarios in real life remain largely unexplored and lack standardized benchmarks for evaluation. To address this gap, we introduce MMR-Life, a comprehensive benchmark designed to evaluate the diverse multimodal multi-image reasoning capabilities of MLLMs across real-life scenarios. MMR-Life consists of 2,646 multiple-choice questions based on 19,108 images primarily sourced from real-world contexts, comprehensively covering seven reasoning types: abductive, analogical, causal, deductive, inductive, spatial, and temporal. Unlike existing reasoning benchmarks, MMR-Life does not rely on domain-specific expertise but instead requires models to integrate information across multiple images and apply diverse reasoning abilities. The evaluation of 37 advanced models highlights the substantial challenge posed by MMR-Life. Even top models like GPT-5 achieve only 58\% accuracy and display considerable variance in performance across reasoning types. Moreover, we analyze the reasoning paradigms of existing MLLMs, exploring how factors such as thinking length, reasoning method, and reasoning type affect their performance. In summary, MMR-Life establishes a comprehensive foundation for evaluating, analyzing, and improving the next generation of multimodal reasoning systems.
\end{abstract}
\vspace{-5pt}
\begin{figure}[htbp]
    \centering
    \includegraphics[width=\linewidth]{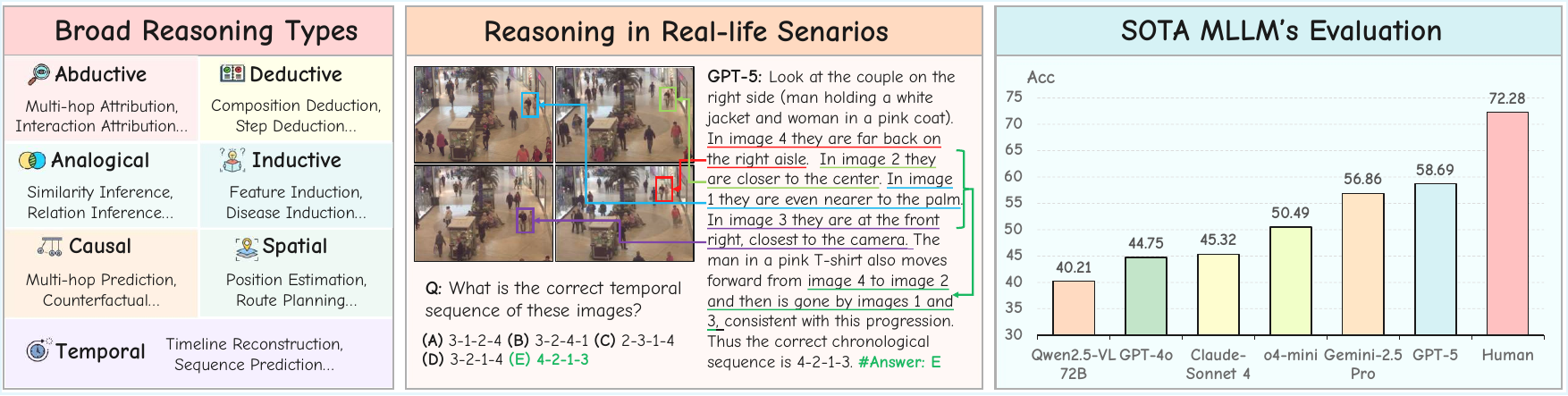}
        \vspace{-10pt}
        \caption{Overview of the \textbf{MMR-Life}. Left: 7 reasoning types and 21 tasks. Middle: A typical example of multi-image reasoning in real-life scenarios. Right: Extensive evaluation reveals a gap between humans and SOTA MLLMs on some real-life reasoning tasks.
        } \label{fig:intro}
    \vspace{-10pt}
\end{figure}
\section{Introduction}\label{sec:1}
Reasoning is the process of generalizing from known premises to new conclusions, and it is considered a key capability for AI systems on the path to artificial general intelligence (AGI) \citep{moss_mir, commonsense_mir,ljc5,ljc2,zhuoran1}. Recently, with the great success of reasoning large language models (RLLMs) in tasks such as mathematical reasoning \citep{deepseek-r1,s1,ljc4}, there has been a widespread exploration of transferring this reasoning-enhanced paradigm to multimodal large language models (MLLMs). Representative models such as Gemini-2.5-Pro \citep{gemini-2.5}, Claude-Sonnet-4 \citep{claude4}, and GPT-5 \citep{gpt-5} leverage long Chain-of-Thought (CoT) \citep{cot} style reasoning to capture key visual information, decompose complex problems, thereby achieving or even surpassing human-level performance in diverse reasoning scenarios.

With the advancement of MLLM reasoning capabilities, there has been an increasing demand for more challenging and realistic multimodal reasoning benchmarks. Recent work mainly evaluates the reasoning ability of MLLMs through two approaches: One line of research collects expert-level domain-specific problems to assess the model's reasoning based on knowledge in areas such as scientific knowledge answering \citep{mmlu-reason,bmmr,mmmu} and math problem solving \citep{mv-math, olympiadbench}.
The other line of research attempts to separate knowledge from reasoning by using synthetic problems like symbolic puzzles to assess reasoning capabilities across different difficulty levels \citep{visualpuzzle, mme-reasoning, puzzlevqa}.

Despite significant progress, current benchmarks still exhibit a considerable deviation from real-life reasoning scenarios. \textbf{(1) From the task design perspective, the tasks in existing benchmarks are not commonly encountered in everyday reasoning.} Both knowledge-intensive tasks and synthesized puzzle-based tasks remain misaligned with the authentic reasoning demands that arise in everyday situations. For the former, daily reasoning seldom relies on expert-level knowledge, whereas for the latter, the symbolic input images differ substantially from those encountered in real-world scenarios. \textbf{(2) From the perspective of input images, current benchmarks fail to include multi-image inputs that span a diverse range of reasoning types.} A large portion of multimodal general reasoning benchmarks focus exclusively on single-image inputs \citep{mmmu, mmmu-pro, visualpuzzle}, which contrasts with real-world conditions where we perceive visual information as a sequence of images rather than a single one. For multi-image benchmarks, existing work either incorporates non-reasoning tasks or focuses on a limited reasoning type \citep{mmrb,mmlm-compbench,mibench, mmiu}, making it difficult to support further comprehensive evaluation of MLLM reasoning performance.

To address these issues, we introduce \textbf{MMR-Life}, a comprehensive benchmark designed to evaluate the multimodal multi-image reasoning capability of MLLMs across real-life scenarios.
MMR-Life contains \textbf{2,646} carefully curated questions, covering 7 distinct reasoning types (see Figure \ref{fig:intro}, \ref{fig:example}), which broadly encompass the reasoning abilities necessary for everyday situations. 
In MMR-Life, each question is associated with a set of images, primarily taken in real-world scenarios. The answers do not require domain-specific expertise but instead ask models to extract key information from multiple real-life images and derive new conclusions. This design aligns MMR-Life more closely with the reasoning types found in everyday life.
Figure \ref{fig:intro} shows an example from MMR-Life. To address the temporal ordering problem, the model needs to detect individuals recurring across different surveillance images and track their movements, selecting the correct order.

Extensive evaluations on 37 advanced MLLMs demonstrate that the real-world reasoning scenarios in MMR-Life remain highly challenging. As illustrated in Figure \ref{fig:intro}, even the most advanced models, including GPT-5 and Gemini-2.5-Pro, reach only 58.69\% and 56.86\% accuracy on MMR-Life, falling short of human performance by 14\%. 
Besides, the evaluation results demonstrate substantial performance disparities across reasoning types. Existing MLLMs perform relatively well on analogical, deductive, and inductive reasoning, but encounter notable bottlenecks in causal, spatial, and temporal reasoning.
Based on MMR-Life, we conduct an analysis of MLLM reasoning paradigms and obtain several key findings, including that long thinking benefits only limited reasoning types, RL's weaker generalization in small models, and the clustering of reasoning types into patterns.

In summary, our contributions include: (1) We propose MMR-Life, the first comprehensive benchmark for evaluating multimodal multi-image reasoning in real-life scenarios across seven reasoning types. (2) Through an extensive evaluation of 37 state-of-the-art MLLMs on MMR-Life, we find that existing models struggle considerably in real-life reasoning, especially in causal, spatial, and temporal tasks. (3) Based on MMR-Life, we conduct an in-depth analysis of current MLLM reasoning paradigms, revealing key findings such as the limited effectiveness of long thinking to certain reasoning types, the weaker generalization of RL on small models, and the presence of pattern clustering across reasoning types.
\vspace{-5pt}
\section{The MMR-Life Benchmark}
\vspace{-3pt}
\subsection{Overview}\label{sec:2.1}
\begin{figure}[tbp]
    \centering
    \includegraphics[width=\linewidth]{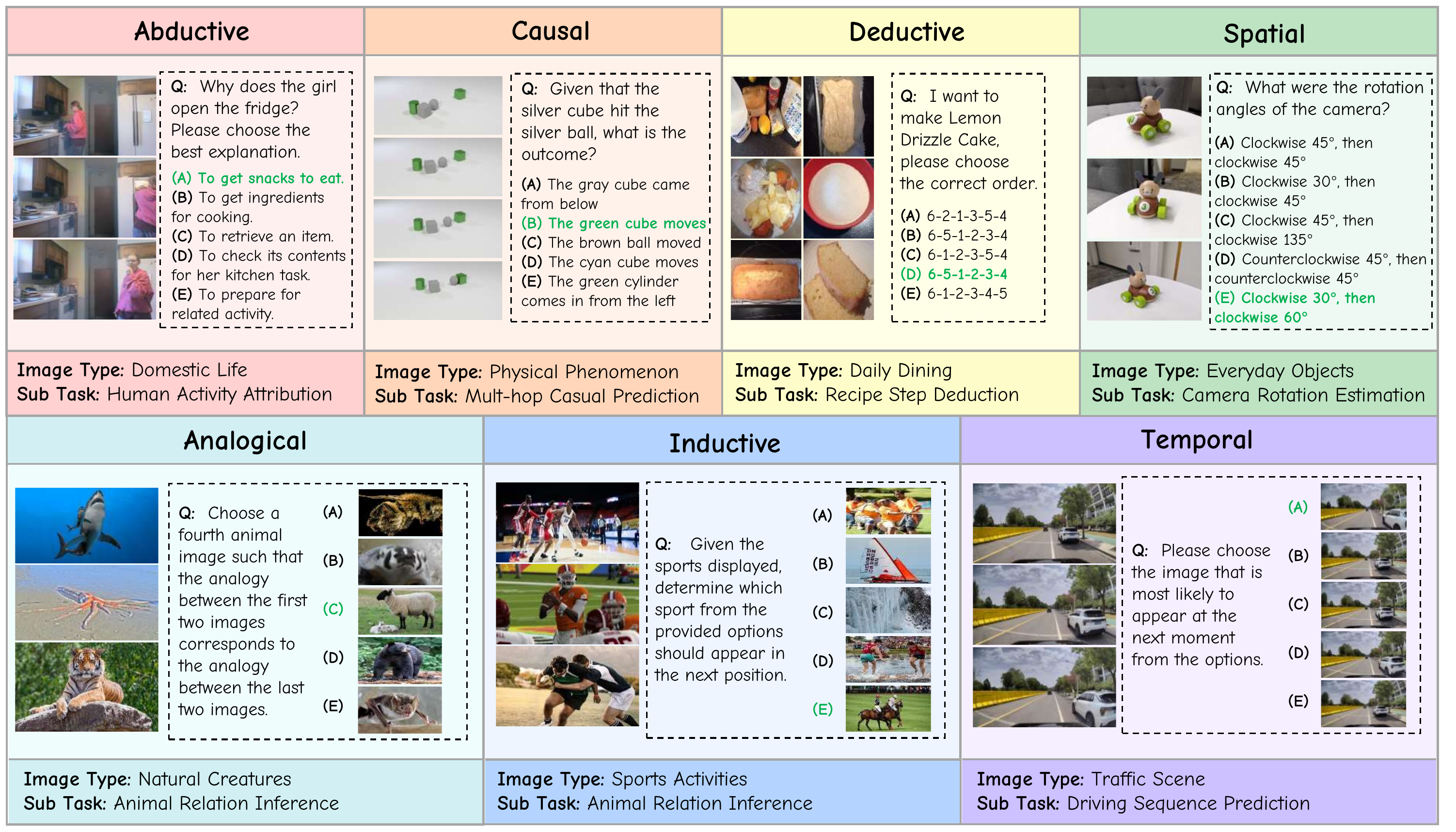}
    \vspace{-10pt}
        \caption{MMR-Life examples from each reasoning type. } \label{fig:example}
        \vspace{-10pt}
\end{figure}

We introduce the \textbf{M}ultimodal \textbf{M}ulti-image \textbf{R}easoning benchmark under real-\textbf{Life} scenarios (MMR-Life), a novel benchmark meticulously curated to evaluate the ability of MLLMs to perform diverse types of reasoning in everyday situations. MMR-Life consists of 2,646 multiple-choice questions based on 19,108 images, comprehensively covering 7 reasoning types (i.e., abductive, analogical, causal, deductive, inductive, spatial, and temporal) and 21 tasks. Each task is based on a set of multi-images, predominantly sourced from real-life contexts, such as domestic life, daily dining, and sports activities. See Figure \ref{fig:example} for examples in MMR-Life and Table \ref{tab:stats} for dataset statistics. We further discuss the key concepts (e.g., real-life scenarios) of our benchmark in Appendix \ref{append:concept}.

\begin{wraptable}{t}{0.4\linewidth}
\centering
\vspace{-20pt} 
\caption{Key statistics of MMR-Life.}\label{tab:stats}
\footnotesize
\resizebox{0.9\linewidth}{!}{
\begin{tabular}{@{}l r@{}}
\toprule
\textbf{Statistics} & \textbf{Number} \\
\midrule
Total Questions & 2,646 \\
Total Reasoning Types/Tasks & 7/21 \\
Image Types & 15 \\
\midrule
Reasoning Type \\ 
\quad - Abductive Reasoning & 307 (11.60\%) \\
\quad - Analogical Reasoning  & 568 (21.47\%) \\
\quad - Causal Reasoning & 263 (9.94\%) \\
\quad - Deductive Reasoning & 282 (10.66\%) \\
\quad - Inductive Reasoning & 429 (16.21\%) \\
\quad - Spatial Reasoning & 255 (9.64\%) \\
\quad - Temporal Reasoning & 542 (20.48\%) \\
\midrule
Text Options & 1454 (54.95\%) \\
Image Options & 1192 (45.05\%) \\
\midrule
Average Image Counts & 7.22 \\
Average Question Length & 283 \\
\bottomrule
\end{tabular}}

\end{wraptable}

\vspace{-5pt}
\subsection{Data Curation Pipeline}\label{sec:2.2}
\paragraph{Data Collection.}
We initiate our pipeline by collecting real-life images from a variety of sources, including: 
\textbf{(1) Public image datasets:} We select high-resolution real-world image datasets from Kaggle \citep{kaggle}, ensuring that the images within each dataset are related (e.g., temporal relationships), to facilitate the construction of multi-image inputs for our questions. 
\textbf{(2) Open web resources:} We take screenshots from publicly available web resources to collect real-world multi-image data. For example, we obtain bird distribution density images from the eBird website \citep{ebird}.
\textbf{(3) Public video sources:} Given the inherent correlation between frames in a video, they are ideal for multi-image data. We extract frames from publicly available video datasets to create images, while ensuring the clarity of each frame.
\textbf{(4) Other existing benchmarks:} Finally, we collect data from existing multi-image or video reasoning benchmarks, extract frames from the videos, and remove images with low quality.
The detailed collection protocol and data sources for each task are reported in Appendix \ref{append:data_source}.

\vspace{-5pt}
\paragraph{Task Design.} To make our benchmark more aligned with real-life scenarios, we aim to cover a broader range of reasoning types, reflecting diverse everyday situations. Specifically, based on the collected images, we design 7 distinct reasoning types (see Figure \ref{fig:example} for examples): 
\textbf{(1) Abductive Reasoning (Abd)}: Given the observed event, inferring the most plausible explanation for why the event occurred.  \textbf{(2) Analogical Reasoning (Ana)}: Inferring conclusions about a new situation by identifying similarities with a known case.  \textbf{(3) Causal Reasoning (Cau)}: In contrast to abductive reasoning, based on the cause, inferring the effect.  \textbf{(4) Deductive Reasoning (Ded)}: Based on general rules or premises, drawing logically certain conclusions about specific cases. \textbf{(5) Inductive Reasoning (Ind)}: Generalizing rules or patterns from specific observations.  \textbf{(6) Spatial Reasoning (Spa)}: Understanding and reasoning about the locations, movement, and spatial relations of objects. \textbf{(7) Temporal Reasoning (Tem)}: Reasoning about the order, duration, and timing of events. 

\vspace{-7pt}
\paragraph{Question-Answer Generation.} We generate question-answer pairs using either automatic synthesis or manual annotation, depending on the task type.  In some cases, the explicit information contained within the multi-image set we collect is already sufficient to fulfill the task's requirements. For example, in the temporal reasoning example of Figure \ref{fig:example}, the images themselves contain sequential information, which is sufficient for the sequence prediction task. In these cases, we can define heuristic rules and use code to automate the synthesis of question-answer pairs using the information. However, some tasks require reasoning over implicit information in images. For instance, in the abductive reasoning example of Figure \ref{fig:example}, we need to identify causal event pairs within the scene to construct the questions. In these cases, we manually design question-answer pairs according to the reasoning type to ensure the quality of the data. This process leads to the creation of a diverse set of 3.2K questions from multiple sources. See Appendix \ref{append:anno_guide} for detailed annotation guidelines and Appendix \ref{append:task} for task details.

\vspace{-7pt}
\paragraph{Negative Option Generation.}
Given that many reasoning tasks do not have a single correct answer (e.g., providing a plausible explanation in abductive reasoning), we design all questions in a multiple-choice format, where the model must choose the most appropriate answer from five options. Each option is presented as either an image or text (with the distribution provided in Table \ref{tab:stats}). For image options, we use heuristic rules to sample incorrect candidates. As an example, in the temporal reasoning example in Figure \ref{fig:example}, we construct negative options by choosing frames that either precede the input images or occur at much later time steps. For text options, we invoke GPT-5-mini \citep{gpt-5}, GPT-4o \citep{gpt-4o}, and Qwen2.5-VL-32B \citep{qwen-2.5-vl} to generate responses (see prompts in Appendix \ref{append:neg_prompt}). From all generated incorrect responses, we manually choose the four highest-quality erroneous options to serve as the final incorrect choices.
\vspace{-7pt}
\paragraph{Data Quality Control.} To further control the quality of our data, we perform three steps of data filtering. \textbf{(1) Difficulty filtering:} We employ three smaller models, Qwen2.5-VL-7B \citep{qwen-2.5-vl}, Gemma3-4B \citep{gemma3}, and InternVL3.5-8B \citep{internvl-3.5}, to generate answers for each question. If all models answer correctly, this suggests that the questions are too easy for existing MLLMs, and they are therefore filtered out. \textbf{(2) Format filtering:} The model-generated incorrect options may have significant format differences (e.g., length) compared to the human-constructed correct answers, which may result in the model relying on shortcuts. To mitigate this effect, we manually revise the options with substantial format differences.  \textbf{(3) Quality filtering:} Finally, we distribute the problems among different co-authors, filtering out questions that exhibit semantic ambiguity, have multiple correct answers, or require domain-specific expertise. 
\begin{table}[t]
\centering
\caption{The comparison between MMR-Life and other existing benchmarks. \textbf{W} (Web), \textbf{T} (Textbook), \textbf{A} (Annotated), \textbf{E} (Existing datasets), and \textbf{Avg Img.\#} (average image counts each question).} \label{tab:comp}
\vspace{-5pt}
\footnotesize
\resizebox{0.9\textwidth}{!}{
\begin{tabular}{lccccccc}
\toprule
\textbf{Dataset} & \textbf{Size} & \textbf{Images} & \textbf{Reason} & \textbf{Source} & \textbf{Knowledge} &\textbf{Avg Img.\#} \\ 
\midrule

MME-Reasoning \citep{mme-reasoning} & 1.2K & Symbolic & 3 Types & W+T+A+E & Low & 1 \\
VisualPuzzles \citep{visualpuzzle}& 1.1K & Symbolic & 5 Types & W+T+A & Low & 1\\ 
MMMU \citep{mmmu} & 11.5K & Hybrid & - & W+T+A & High & 1.05 \\ 
 
MMLU-Reason \citep{mmlu-reason}  & 1.1K  & Hybrid  & 6 Types  &   W+T+E   & Medium         & 1.85   \\ 
MEGA-Bench \citep{mega-bench} & 8K & Hybrid & - & A & Medium &  2\\
MME-COT \citep{mme-cot}& 1.1K   & Hybrid   & 6 Types  & E  & Medium  & 2.10   \\ 
MV-MATH \citep{mv-math} & 2K & Symbolic & 1 Type & W+A & High & 3.02 \\ 
MMRB \citep{mmrb} & 4.8K & Hybrid & 3 Types & E & Medium & 6.17  \\ 
MMIU \citep{mmiu} & 11.6K & Hybrid & - & A & Medium & 6.64 \\ 

\midrule
\textbf{MMR-Life (Ours)} & \textbf{2.7K}  & \textbf{Natural}  & \textbf{7 Types}   & \textbf{A}    &  \textbf{Low}    & \textbf{7.22}  \\ 
\bottomrule
\end{tabular}}
\vspace{-10pt}
\end{table}
\vspace{-5pt}
\subsection{Comparisons with Existing Benchmarks}
To further distinguish the difference between MMR-Life and other existing ones, we provide detailed comparisons in Table \ref{tab:comp}. From the image type perspective, most existing datasets include a large proportion of symbolic images such as charts and puzzles, which creates a gap from the natural images encountered in daily life. Our benchmark excludes such images, making the evaluation more closely aligned with real-life scenarios. From the source perspective, all questions in our dataset are newly annotated rather than sampled directly from existing datasets, textbooks, or the web, which reduces the risk of data contamination.

\section{Main Experiment} \label{sec:3}

\subsection{Experimental Settings}
\paragraph{Multi-modal Language Models without Thinking.} We first evaluate the performance of SOTA non-thinking MLLMs on our benchmark. These models have not undergone additional reasoning-enhancement training and lack long CoT capabilities. Open-source models include Qwen2.5-7/32/70B \citep{qwen-2.5-vl}, Gemma3-12/27B \citep{gemma3}, InternVL3.5-8B/30B-A3B \citep{internvl-3.5}. Closed-source models include GPT-4.1-mini, GPT-4.1 \citep{gpt-4.1}, GPT-4o \citep{gpt-4o}, Claude-3.7-Sonnet (without thinking) \citep{claude-3.7} and Doubao-1.5-vision \citep{doubao-1.5-vision}.

\paragraph{Multi-modal Language Models with Thinking.} To study the effect of long CoT patterns on the reasoning abilities of MLLMs, we introduce several advanced thinking models into the evaluation. Open-source models include VL-Rethinker-7/72B\citep{vl-rethinker}, MM-Eureka-Qwen-32B \citep{mm-eureka}, MiMo-VL-7B-RL \citep{mimo-vl}, Keye-VL-1.5-8B \citep{keye-vl}, QVQ-72B-Preview \citep{qvq}. Closed-source models include o4-mini \citep{o4}, Claude-Sonnet-4-Thinking \citep{claude4}, Gemini-2.5-Flash \citep{gemini-2.5}, Gemini-2.5-Pro \citep{gemini-2.5}, GPT-5-mini and GPT-5 \citep{gpt-5}. We provide complete experiments and results for a total of 37 models in Appendix \ref{append:main_result}.

\paragraph{Human Level Performance.} We employ 12 students with varying degrees and academic backgrounds. Then, we extract 10 questions from each task to form a mini test set of 210 unique questions. From this pool, we repeatedly sample 50 questions at a time and assign them to one of 12 students, yielding a total of 600 valid human answers. These students are instructed not to use external knowledge sources such as the internet or books. We report the experimental results on this tiny set in Appendix \ref{append:tiny_result}.

\paragraph{Implementation Details.}
We employ the same zero-shot CoT prompt as input for all models in the main experiments to perform reasoning. To minimize random variation, we conduct five runs for every open-source model and use the average performance as the final outcome. All experiments are performed using 8 NVIDIA A100 GPUs. The detailed experimental parameters and prompts are provided in Appendix \ref{append:main_set}.

\subsection{Main Results}
Table \ref{tab:benchmark} presents MLLMs' performance on MMR-Life, from which we draw several critical insights:

\paragraph{Our MMR-Life benchmark poses significant challenges for MLLMs.} Despite achieving nearly 90\% accuracy \citep{gpt-5} on complex multimodal reasoning tasks like GPQA \citep{gpqa} and MMMU \citep{mmmu}, GPT-5 only achieved an accuracy of \textbf{58.69\%} on MMR-Life, with a 14\% gap compared to human performance. Moreover, almost all open-source models have an accuracy rate below \textbf{40\%}, with some of the most recent models, such as Skywork-R1V-38B and InternVL3.5-8B, performing worse than random guessing (20\%). \textbf{This suggests that, although MMR-Life does not include complex knowledge requirements, our real-life reasoning scenarios still present a significant challenge for current MLLMs.} Future model training and optimization should focus more on these real-world situations.

\begin{table*}[t]
\centering
\caption{Performance comparison of SOTA MLLMs on MMR-Life. The highest and lowest scores for each model type across reasoning types are highlighted in \colorbox{mygreen}{green} and \colorbox{mypink}{red}, respectively. The highest performance achieved by the model in each type is indicated in \textbf{bold}.}
\vspace{-5pt}
\label{tab:benchmark}
\footnotesize
\resizebox{0.9\linewidth}{!}{
\begin{tabular}{lcccccccc}
\toprule
\textbf{Model} & \textbf{Abd} & \textbf{Ana} & \textbf{Cau} & \textbf{Ded} & \textbf{Ind}
& \textbf{Spa} & \textbf{Tem} & \textbf{Avg} \\
\midrule
Human & 79.76 & 57.65 & 75.00 & 70.59 & 63.41 & 79.76 & 79.76 & 72.28 \\

\midrule
\multicolumn{9}{c}{\textit{Closed-source \& Thinking}} \\
\hdashline
 GPT-5 & 53.75 & \cellcolor{mygreen}\textbf{78.87} & 41.06 & \cellcolor{mygreen}\textbf{80.14} & \cellcolor{mygreen}\textbf{78.32} & 17.25 & \cellcolor{mygreen}\textbf{41.70} & 58.69 \\
Gemini-2.5-Pro & \cellcolor{mygreen}\textbf{54.40} & 73.77 & 36.99 & 79.43 & 73.66 & \cellcolor{mygreen}\textbf{25.10} & 35.79 & 56.86 \\
Gemini-2.5-Flash & 46.25 & 75.18 & 34.22 & 71.63 & 73.66 & 23.92 & 30.81 & 53.10 \\
o4-mini & 41.37 & 73.59 & \cellcolor{mypink}27.38 & 71.28 & 68.07 & 19.22 & 32.66 & 50.49 \\
GPT-5-mini & 44.95 & 69.72 & 32.32 & 75.18 & 68.76 & \cellcolor{mypink}12.16 & 29.52 & 49.77 \\
Claude-Sonnet-4 & \cellcolor{mypink}36.96 & \cellcolor{mypink}60.92 & \cellcolor{mygreen}\textbf{44.11} & \cellcolor{mypink}67.02 & \cellcolor{mypink}56.64 & 15.69 & \cellcolor{mypink}28.23 & 45.32 \\
\midrule
\multicolumn{9}{c}{\textit{Closed-source \& No Thinking}} \\
\hdashline
GPT-4.1 & 44.30 & \cellcolor{mygreen}71.30 & \cellcolor{mypink}22.43 & \cellcolor{mygreen}67.38 & \cellcolor{mygreen}70.16 & 13.73 & 27.31 & 48.15 \\
Claude-3.7-Sonnet & 33.55 & 66.55 & 35.36 & 59.93 & 59.67 & \cellcolor{mygreen}20.78 & 26.01 & 45.09 \\
GPT-4o & \cellcolor{mygreen}46.91 & 65.67 & 25.86 & \cellcolor{mypink}51.42 & 66.20 & \cellcolor{mypink}11.37 & 26.01 & 44.75 \\
GPT-4.1-mini & \cellcolor{mypink}32.90 & 61.62 & \cellcolor{mygreen}30.80 & 52.13 & 65.27 & 16.47 & \cellcolor{mygreen}30.63 & 44.10 \\
Doubao-1.5-vision & 37.13 & \cellcolor{mypink}53.70 & 31.18 & 59.57 & \cellcolor{mypink}54.31 & 12.16 & \cellcolor{mypink}23.06 & 39.98 \\
\midrule
\multicolumn{9}{c}{\textit{Open-source \& Thinking}} \\
\hdashline

VL-Rethinker-72B & 36.48 & \cellcolor{mygreen}50.88 & 33.08 & 56.03 & \cellcolor{mygreen}57.58 & 15.69 & 21.59 & 39.68 \\
QVQ-72B-Preview & 31.27 & 41.20 & \cellcolor{mygreen}38.02 & 47.87 & 31.24 & 14.12 & 16.42 & 31.14 \\
MM-Eureka-Qwen-32B & 26.06 & 41.02 & 25.10 & 47.52 & 27.97 & \cellcolor{mygreen}16.08 & 17.34 & 29.02 \\
MiMo-VL-7B-RL & \cellcolor{mygreen}38.76 & 25.88 & 28.14 & \cellcolor{mygreen}60.99 & 24.94 & 14.12 & 19.19 & 28.68 \\
Keye-VL-1.5-8B & \cellcolor{mypink}19.87 & 21.30 & 23.95 & \cellcolor{mypink}14.18 & 20.28 & 13.73 & \cellcolor{mygreen}23.62 & 20.22 \\
Skywork-R1V-38B & 22.15 & \cellcolor{mypink}10.39 & \cellcolor{mypink}16.73 & 23.76 & \cellcolor{mypink}11.89 & \cellcolor{mypink}9.80 & \cellcolor{mypink}11.07 & 14.13 \\
\midrule
\multicolumn{9}{c}{\textit{Open-source \& No Thinking}} \\
\hdashline
Qwen2.5-VL-72B & 35.50 & 55.46 & 35.36 & \cellcolor{mygreen}52.13 & 55.48 & 12.94 & \cellcolor{mygreen}23.80 & 40.21 \\
Gemma3-27B & 35.18 & \cellcolor{mygreen}57.92 & \cellcolor{mygreen}36.88 & 31.21 & \cellcolor{mygreen}60.61 & 12.94 & 18.27 & 38.32 \\
Gemma3-12B & 25.08 & 50.70 & \cellcolor{mypink}17.11 & 27.30 & 42.42 & 10.20 & 15.87 & 29.52 \\
Qwen2.5-VL-32B & \cellcolor{mypink}23.45 & 42.78 & 21.29 & 50.00 & 27.27 & \cellcolor{mygreen}15.69 & 16.24 & 28.61 \\
Qwen2.5-VL-7B & 26.06 & 35.74 & 20.53 & \cellcolor{mypink}20.92 & 38.93 & \cellcolor{mypink}9.41 & \cellcolor{mypink}12.18 & 24.68 \\
InternVL3.5-30B-A3B & \cellcolor{mygreen}45.60 & 19.19 & 33.46 & 36.52 & 14.45 & 12.16 & 14.39 & 23.09 \\
InternVL3.5-8B & 35.18 & \cellcolor{mypink}11.44 & 18.63 & 34.04 & \cellcolor{mypink}11.19 & 14.90 & 16.61 & 18.67 \\
\bottomrule
\end{tabular}}
\vspace{-10pt}
\end{table*}
\paragraph{MLLMs exhibit large disparities across different types of reasoning.} While current models perform well in analogical, deductive, and inductive reasoning tasks, they still have substantial room for improvement in causal, spatial, and temporal reasoning tasks. We observe that all models perform poorly in spatial reasoning, with the highest accuracy being only \textbf{25.10\%}, compared to the human accuracy of \textbf{79.76\%}. In contrast, for tasks like analogical reasoning, most closed-source models outperform human performance. Current models can easily acquire abilities such as analogy and deductive reasoning through feature associations or by memorizing explicit reasoning paths. However, \textbf{they struggle to learn more abstract world representations, such as spatial and temporal reasoning}. This bias is one that future model training should seek to correct.


\paragraph{Current open-source thinking models bring limited improvement.} When evaluating the effect of adding a thinking mode to MLLMs, we find that closed-source thinking models generally outperform closed-source no-thinking models. 
However,\textbf{ for open-source models, the thinking mode does not show improved reasoning capabilities.} In Table \ref{tab:benchmark}, the open-source no-thinking model achieves an average accuracy of 29.01\%, whereas the thinking model achieves only \textbf{27.15\%} on average.
This implies that there is substantial potential for improving the reasoning abilities of current open-source thinking models, particularly in their ability to generalize to real-world contexts.

\vspace{-8pt}
\section{Thinking Pattern Analysis}
\vspace{-5pt}
\subsection{Is Longer Thinking Always Better?}
\begin{figure}[tbp]
  \centering
  \begin{minipage}[b]{0.44\textwidth}
    \centering
    \includegraphics[width=0.9\linewidth]{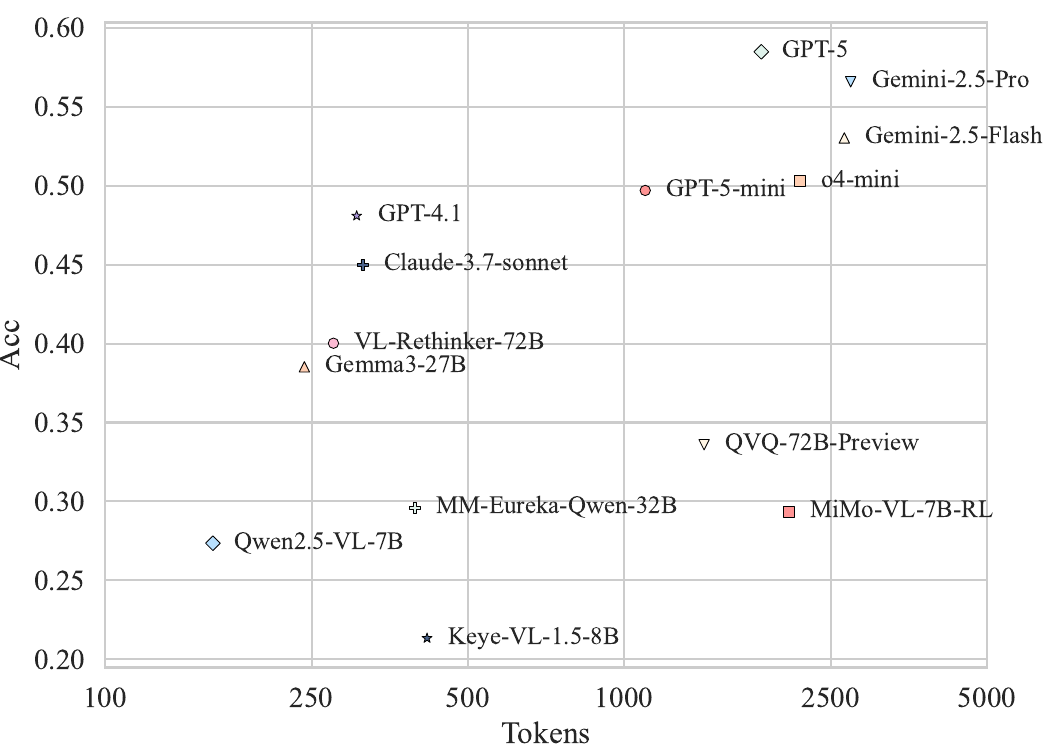}
    \caption{Response tokens vs. Accuracy.} \label{fig:token_acc_comp}
  \end{minipage}
  \begin{minipage}[b]{0.55\textwidth}
    \centering
    \begin{subfigure}[t]{0.49\linewidth}
        \centering
        \includegraphics[width=\linewidth]{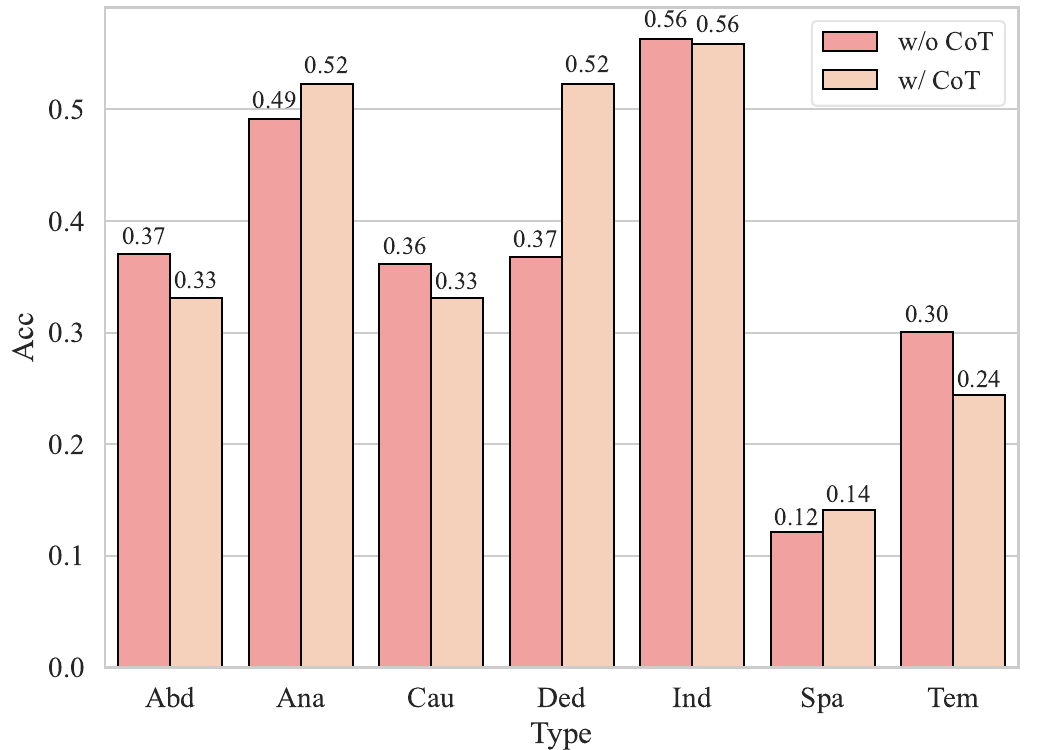}
        \caption{Qwen2.5-VL-72B}\label{fig:nothink_qwen2.5-72B}
    \end{subfigure}
    \begin{subfigure}[t]{0.49\linewidth}
        \centering
        \includegraphics[width=\linewidth]{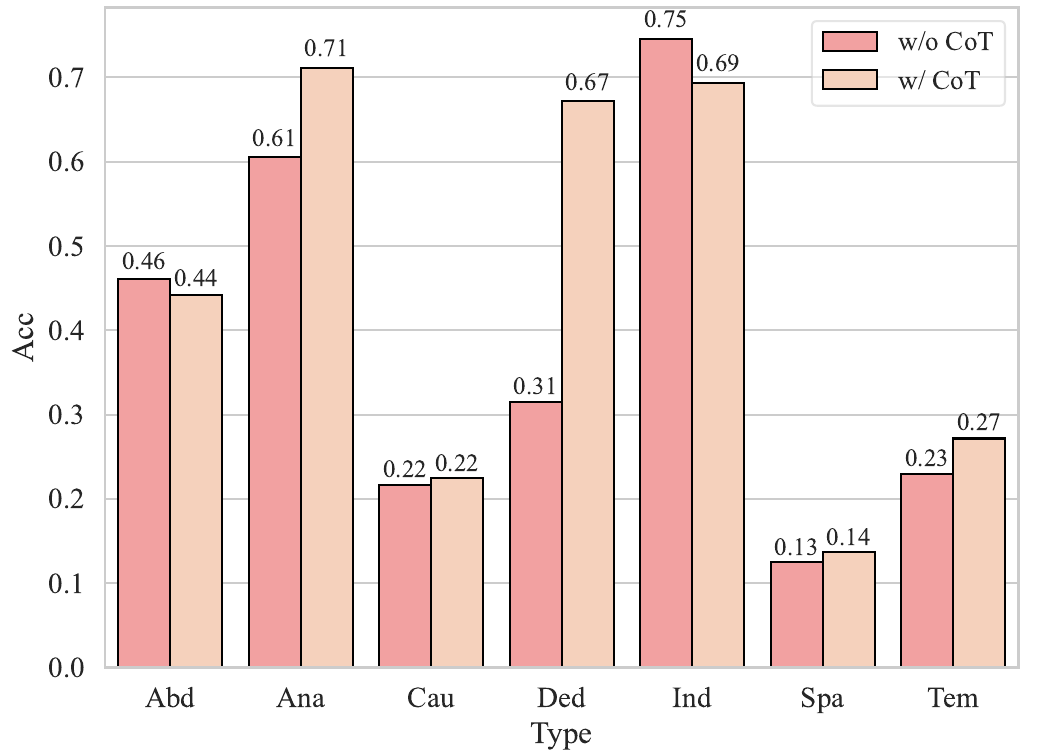}
        \caption{GPT-4.1} \label{fig:nothink_gpt-4.1}
    \end{subfigure}
    \caption{Performance: without CoT vs. with CoT.} \label{fig:nothink}
  \end{minipage}

\end{figure}

From Table \ref{tab:benchmark}, we find that closed-source thinking models perform best on MMR-Life. An important question then arises: Is this superior performance associated with the longer reasoning processes?
\vspace{-7pt}
\paragraph{Reasoning Performance Scales Logarithmically With Thinking Length.} To investigate the question, we first present the semi-log plot of average response token count versus average accuracy over 14 models (see Figure \ref{fig:token_acc_comp}).
The overall trend shows that models with longer outputs tend to achieve higher scores, indicating that reasoning capabilities scale roughly in proportion to the logarithm of the reasoning length. However, there are notable exceptions. Certain open-source thinking models, including MiMo-VL-7B-RL and QVQ-72B-Preview, are located in the lower-right region of Figure \ref{fig:token_acc_comp}, demonstrating that balancing reasoning efficiency and model effectiveness remains a major challenge for future open-source MLLMs.

\begin{figure}[tbp]
    \centering
        \begin{subfigure}[b]{0.329\textwidth}
            \centering
            \includegraphics[width=\textwidth]{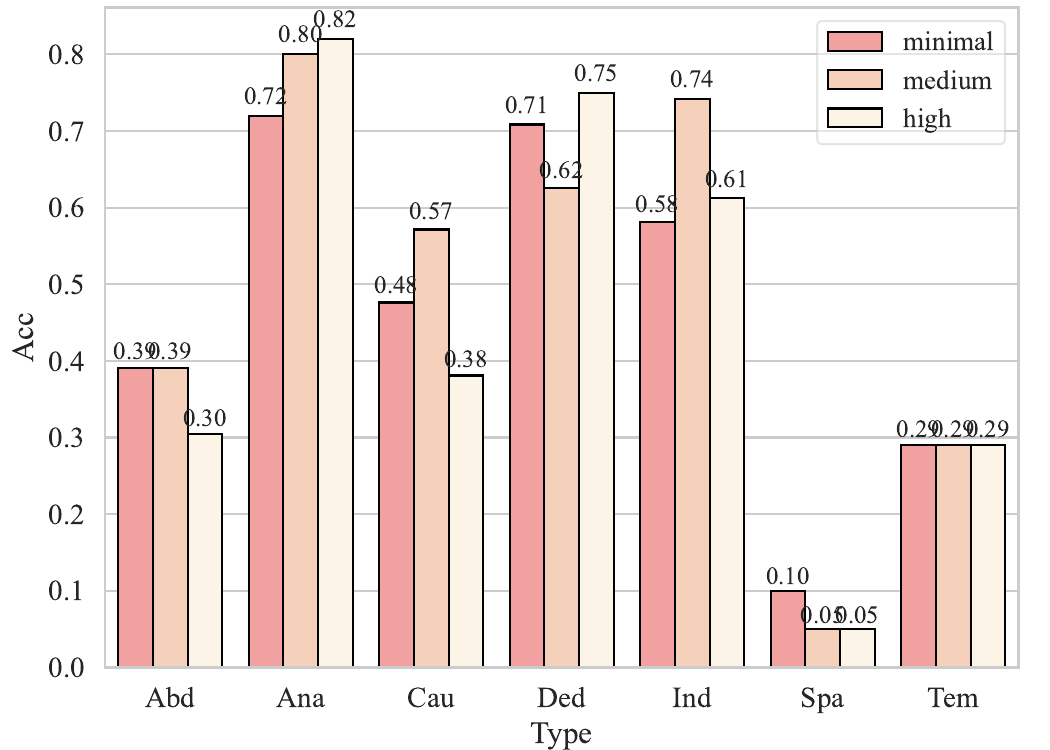}
            \caption{Gemini-2.5-Flash}\label{fig:think_1}
        \end{subfigure}
        \begin{subfigure}[b]{0.329\textwidth}
            \centering
            \includegraphics[width=\textwidth]{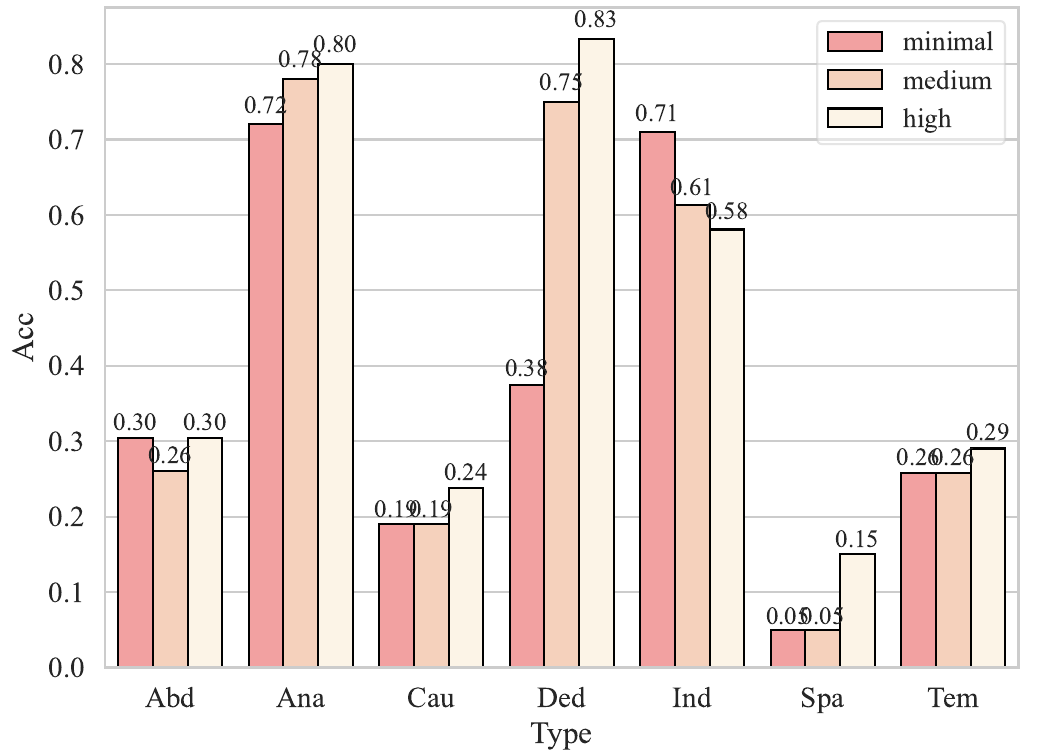}
            \caption{GPT-5-mini}\label{fig:think_2}
        \end{subfigure}
        \begin{subfigure}[b]{0.329\textwidth}
            \centering
            \includegraphics[width=\textwidth]{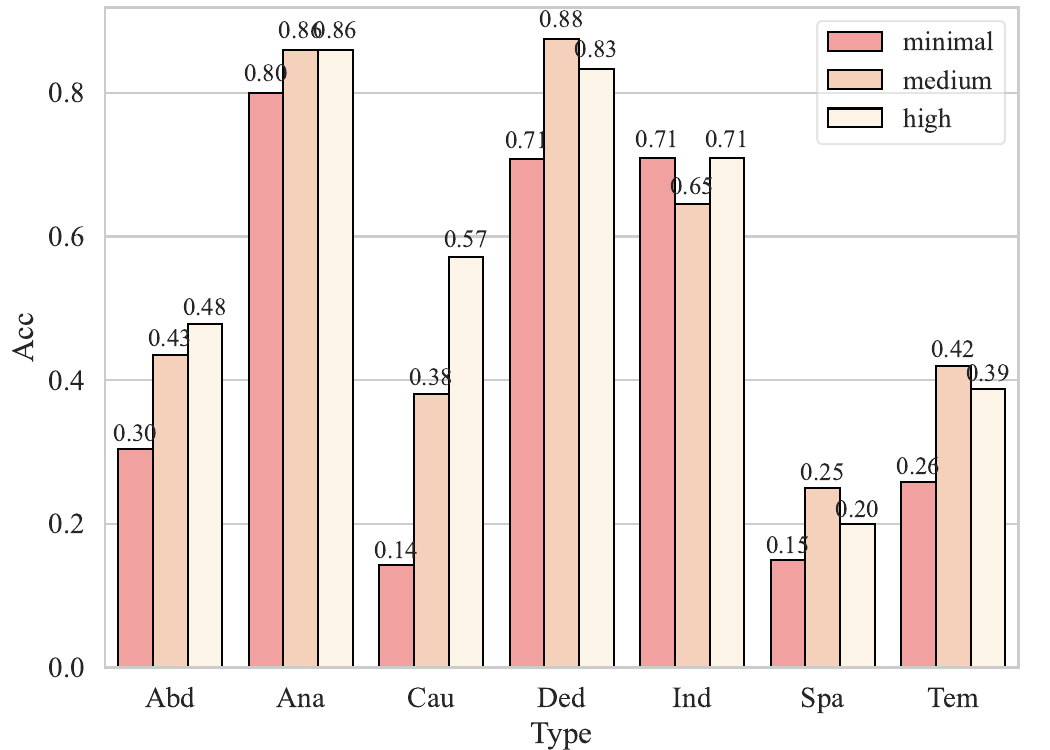}
            \caption{GPT-5}\label{fig:think_3}
        \end{subfigure}
        \caption{Performance comparison under different thinking budgets.} \label{fig:think}
\vspace{-4mm}
\end{figure}

\paragraph{Longer Thinking Is Not All You Need.}
We conduct a more fine-grained analysis to investigate the relationship between model performance and thinking length across distinct reasoning types. Specifically, for no-thinking models, we follow prior work by comparing their reasoning performance with and without CoT \citep{ljc1,cot_or_not} (see Figure \ref{fig:nothink}). For thinking models, we select those with a controllable reasoning budget and vary the budget (minimal, medium, and high) to gradually increase CoT length, thereby comparing performance across different thinking lengths (see Figure \ref{fig:think}).
From both figures, it is evident that longer thoughts do not lead to better performance for all reasoning types. For reasoning types like inductive reasoning, the performance with CoT is worse in no-thinking models (see Figure \ref{fig:nothink}) and using more reasoning budget does not lead to better performance in thinking models (see Figure \ref{fig:think}). Conversely, for reasoning types such as analogical reasoning, the incorporation of CoT or longer CoT results in a noticeable performance improvement. We hypothesize that this is because longer CoT may only be suitable for tasks requiring step-by-step reasoning, while types like inductive reasoning may benefit more from faster thinking \citep{ljc1}. 

\vspace{-1.5mm}
\subsection{Do Generalizable Reasoning Enhancement Methods Exist?}
From the inception of CoT \citep{cot} to the broad application of GRPO \citep{deepseek-r1}, the reasoning-enhancement techniques have undergone substantial evolution. In this section, we analyze and compare the generalizability of these approaches.
\vspace{-5pt}
\paragraph{Failure of Enhancement Methods in Larger Models.} We select four representative reasoning-enhancement methods for comparison: CoT, Self-Consistency (SC), Best-of-N (BoN), and GRPO. To evaluate the generalizability of these methods, we directly use previously trained models for inference without any training on MMR-Life. Specifically, we adopt the Skywork-VL Reward \citep{reward_skywork} as the reward model for BoN and the VL-Rethinker series \citep{vl-rethinker} as the GRPO-trained models. As shown in Table \ref{tab:method}, the results demonstrate that: Across model scales from 7B to 72B, the average performance difference between other methods and CoT consistently decreases, while an increasing number of subtypes transition from performance gains to performance drops (from green to red). Strikingly, on Qwen-2.5-VL-72B, the performance of BoN and GRPO falls short of simply applying CoT.  According to previous works \citep{rl_learn}, we hypothesize that this is because these methods primarily improve sampling efficiency towards correct reasoning paths. For larger models, the likelihood of sampling correct paths is naturally higher, which diminishes the gains from reasoning-enhancement methods.
\begin{table*}[tbp]
\centering
\caption{Performance across different methods. Scores higher and lower than the base model's CoT performance are marked in \colorbox{mygreen}{green} and \colorbox{mypink}{red}. The highest score in each column is in \textbf{bold}.}
\label{tab:method}
\footnotesize
\resizebox{0.9\linewidth}{!}{
\begin{tabular}{llcccccccc}
\toprule
\textbf{Model} & \textbf{Method} & \textbf{Abd} & \textbf{Ana} & \textbf{Cau} & \textbf{Ded} & \textbf{Ind}
& \textbf{Spa} & \textbf{Tem} & \textbf{Avg ($\Delta$)} \\
\midrule

\multirow{4}{*}{\textbf{Qwen2.5-VL-7B}} 
& CoT & 26.06 & 35.74 & 20.53 & 20.92 & 38.93 & 9.41 & 12.18 & 24.68\\
& SC@8 & \cellcolor{mygreen}28.01 & \cellcolor{mygreen}39.44 & \cellcolor{mygreen}\textbf{23.57} & \cellcolor{mygreen}25.18 & \cellcolor{mygreen}45.45 & \cellcolor{mygreen}10.98 & \cellcolor{mygreen}\textbf{13.10} & 27.85 \textcolor{green}{(+3.17)}\\
& BoN@8 & \cellcolor{mygreen}27.64 & \cellcolor{mygreen}\textbf{44.72} & \cellcolor{mygreen}22.81 & \cellcolor{mygreen}25.53 & \cellcolor{mygreen}\textbf{48.02} & \cellcolor{mygreen}13.33 & \cellcolor{mygreen}\textbf{13.10} & \textbf{29.54} \textcolor{green}{(+4.86)}\\
& GRPO & \cellcolor{mygreen}\textbf{30.62} & \cellcolor{mygreen}40.49 & \cellcolor{mygreen}21.29 & \cellcolor{mygreen}\textbf{28.72} & \cellcolor{mygreen}43.59 & \cellcolor{mygreen}\textbf{13.73} & \cellcolor{mypink}11.81 & 28.23 \textcolor{green}{(+3.55)} \\

\midrule
\multirow{4}{*}{\textbf{Qwen2.5-VL-32B}} 
& CoT & 23.45 & 42.78 & 21.29 & 50.00 & 27.27 & 15.69 & 16.24 & 28.61 \\
& SC@8 & \cellcolor{mygreen}\textbf{26.06} & \cellcolor{mygreen}\textbf{45.42} & \cellcolor{mygreen}23.95 & \cellcolor{mygreen}51.77 & \cellcolor{mygreen}28.67 & \cellcolor{mygreen}\textbf{16.47} & \cellcolor{mygreen}17.90 & 30.57 \textcolor{green}{(+1.96)} \\
& BoN@8 & \cellcolor{mygreen}25.78 & \cellcolor{mygreen}44.89 & \cellcolor{mypink}19.39 & \cellcolor{mygreen}\textbf{55.32} & \cellcolor{mygreen}30.54 & \cellcolor{mygreen}\textbf{16.47} & \cellcolor{mygreen}\textbf{19.56} & \textbf{30.97} \textcolor{green}{(+2.36)} \\
& GRPO & \cellcolor{mypink}22.98 & \cellcolor{mygreen}42.96 & \cellcolor{mygreen}\textbf{28.14} & \cellcolor{mypink}49.65 & \cellcolor{mygreen}\textbf{30.77} & \cellcolor{mypink}14.90 & \cellcolor{mygreen}19.19 & 30.29 \textcolor{green}{(+1.68)} \\

\midrule

\multirow{4}{*}{\textbf{Qwen2.5-VL-72B}} 
& CoT & 35.50 & 55.46 & \textbf{35.36} & 52.13 & 55.48 & 12.94 & 23.80 & 40.21 \\
& SC@8 & \cellcolor{mypink}35.18 & \cellcolor{mygreen}\textbf{56.16} & \textbf{35.36} & 52.13 & \cellcolor{mypink}54.78 & 12.94 & \cellcolor{mygreen}24.35 & \textbf{40.33} \textcolor{green}{(+0.12)}\\
& BoN@8 & \cellcolor{mypink}34.20 & \cellcolor{mypink}53.35 & \cellcolor{mypink}32.70 & \cellcolor{mypink}51.77 & \cellcolor{mygreen}56.88 & \cellcolor{mygreen}13.73 & \cellcolor{mygreen}\textbf{24.72} & 39.80 \textcolor{red}{(-0.41)}\\
& GRPO & \cellcolor{mygreen}\textbf{36.48} & \cellcolor{mypink}50.88 & \cellcolor{mypink}33.08 & \cellcolor{mygreen}\textbf{56.03} & \cellcolor{mygreen}\textbf{57.58} & \cellcolor{mygreen}\textbf{15.69} & \cellcolor{mypink}21.59 & 39.68 \textcolor{red}{(-0.53)}\\

\bottomrule
\end{tabular}}

\end{table*}

\begin{figure}[tbp]
  \centering
  \begin{minipage}[b]{0.37\textwidth}
    \centering
    \includegraphics[width=\linewidth]{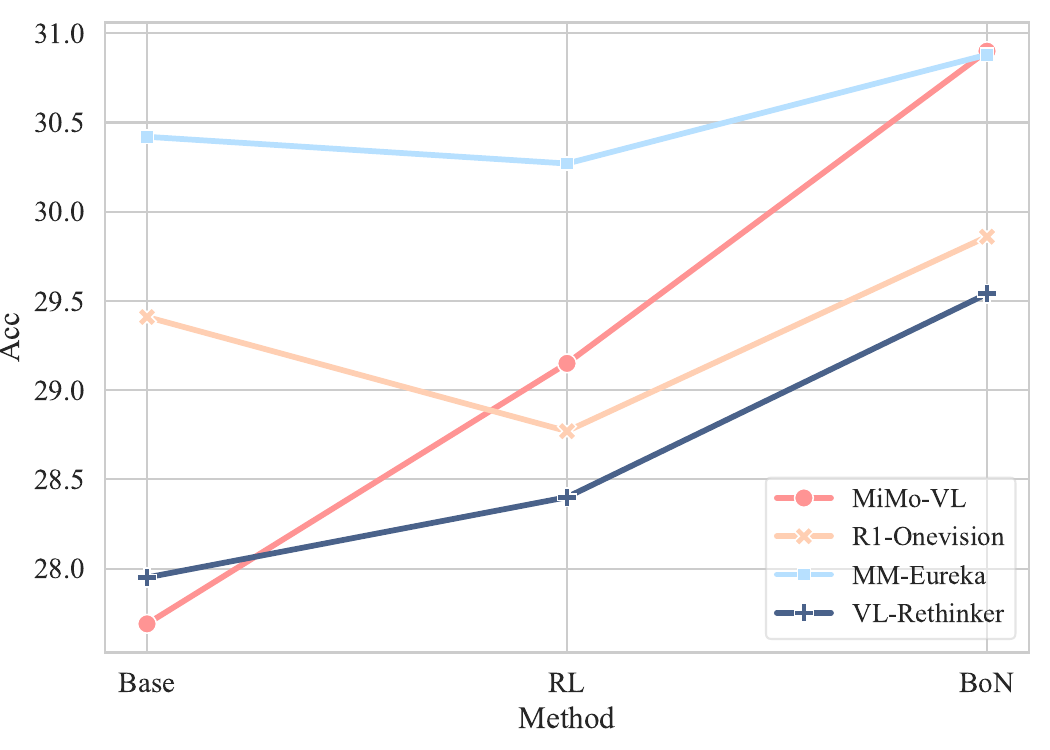}
    \caption{Comparison of BoN and RL performance on different models.} \label{fig:rl_bon_comp}
  \end{minipage}
  \hfill
  \begin{minipage}[b]{0.61\textwidth}
    \centering
    \begin{subfigure}[b]{0.495\linewidth}
            \centering
            \includegraphics[width=\linewidth]{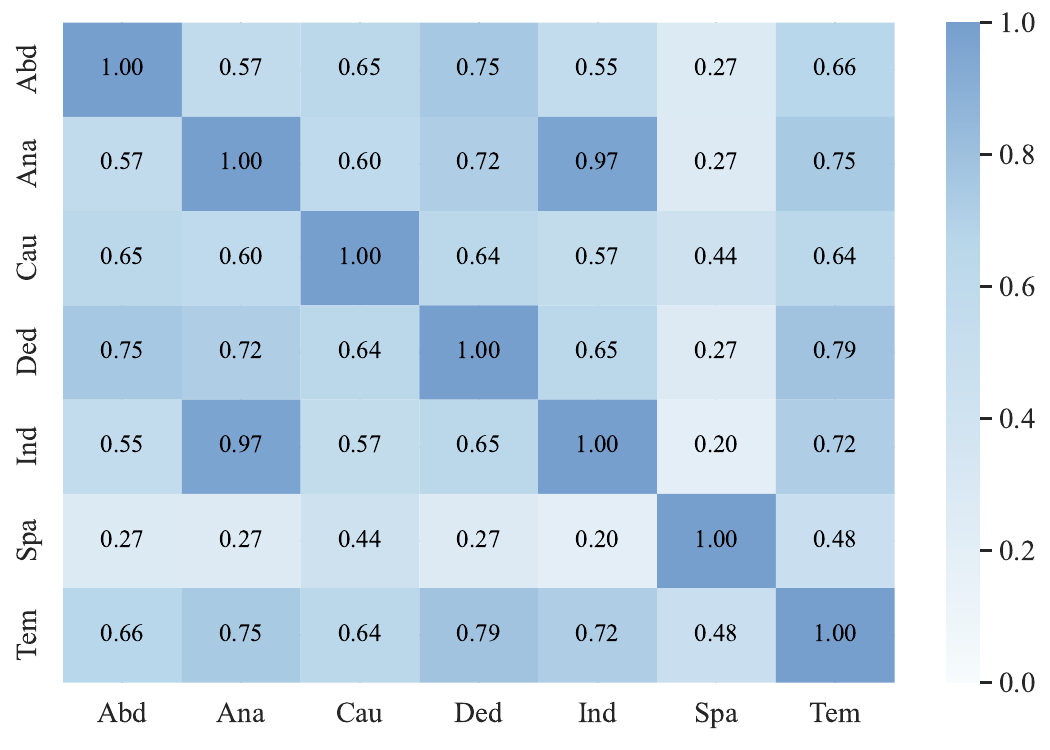}
            \caption{Correlation heatmap}\label{fig:type_corr}
        \end{subfigure}
        \begin{subfigure}[b]{0.495\linewidth}
            \centering
            \includegraphics[width=\linewidth]{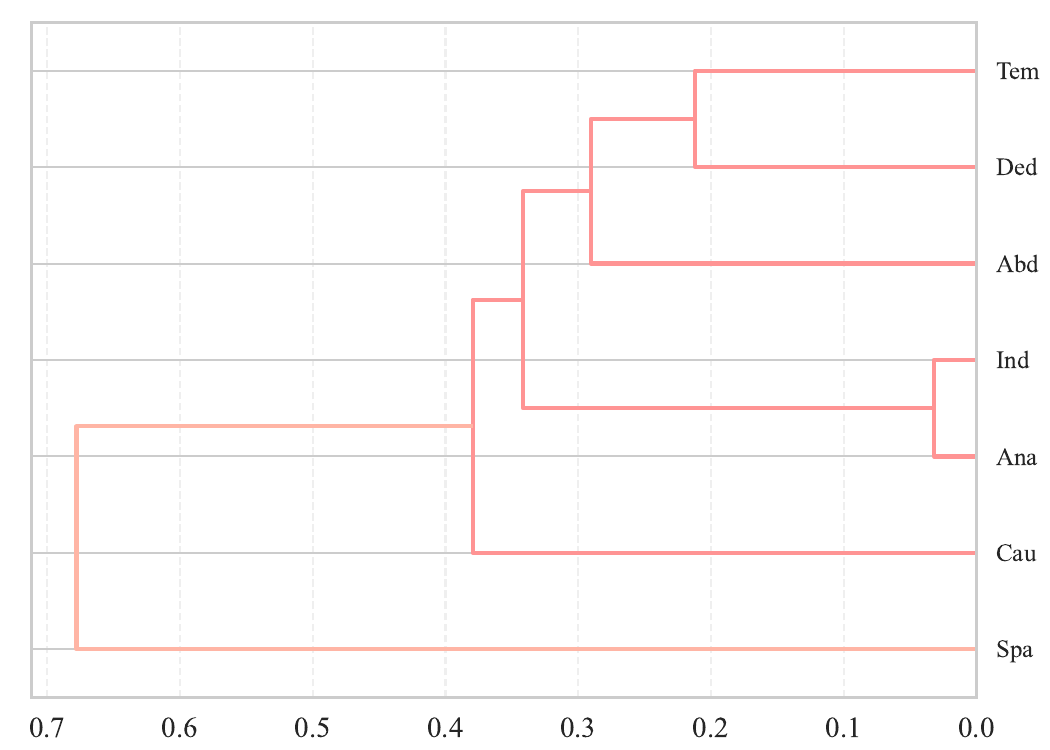}
            \caption{Hierarchical clustering} \label{fig:type_cluster}
        \end{subfigure}
    \caption{Analysis of correlations across different reasoning types (averaged across all models we evaluate).} \label{fig:corr}
  \end{minipage}
\vspace{-10pt}
\end{figure}

\vspace{-5pt}
\paragraph{RL Generalizes Worse than BoN on Small Models.} Reinforcement learning methods, exemplified by GRPO, have gained wide adoption for their strong reasoning generalization \citep{deepseek-r1}. Nevertheless, our results in Table \ref{tab:method} reveal that on all of the three models, GRPO exhibits weaker generalization compared to BoN. To further validate this finding, we conduct experiments on additional small MLLMs (see Appendix \ref{append:thinking} for details), comparing the performance of BoN@8 applied to base models with that of RL-trained models. The results in Figure \ref{fig:rl_bon_comp} show that across different model architectures and training datasets, RL-trained models consistently underperform compared to BoN inference on the corresponding base models. In some cases, RL models even perform worse than the base models using CoT.
This calls for a reconsideration of RL techniques: Do RL methods on small models merely lead to overfitting on specific datasets? We leave this question open for further exploration in future work.

\subsection{Do Different Reasoning Types Correlate?}
Former findings demonstrate significant differences in model performance across types. In this section, we aim to capture the underlying relationships among these categories.
\vspace{-2pt}
\paragraph{Correlations Between Reasoning Types.} We compute the accuracy of all models across reasoning types, calculate the Pearson correlation coefficients between them, and present the results in Figure \ref{fig:type_corr}. It demonstrates substantial differences in correlations across these types. Some categories, such as inductive and analogical reasoning, exhibit very high correlations (0.97), while some others, such as spatial and inductive reasoning, show low correlations (0.40). 
\vspace{-2pt}
\paragraph{Uncovering Pattern Clusters in Reasoning.} 
Furthermore, we normalize the negative of the correlation coefficient as the distance between categories and perform hierarchical clustering. In Figure \ref{fig:type_cluster}, we observe clusters formed by similar reasoning types (e.g., Ana–Ind), suggesting the existence of higher-order reasoning patterns in MLLMs. For example, both analogical and inductive reasoning rely on a shared pattern of abstracting general rules from concrete features. Conversely, reasoning types with greater distances suggest that they involve relatively disjoint patterns. As an example, spatial reasoning is distant from all other categories, suggesting that the capabilities it requires (e.g., location, distance estimation) are difficult to learn from non-spatial tasks.
In conclusion, MMR-Life enables us to uncover a higher-level hierarchy of reasoning patterns, facilitating a deeper understanding of reasoning generalization across diverse tasks.

  



\begin{wrapfigure}{r}{0.5\textwidth} 
\vspace{-10mm}
  \centering
  \begin{subfigure}[b]{0.245\textwidth}
    \centering
    \includegraphics[width=\linewidth]{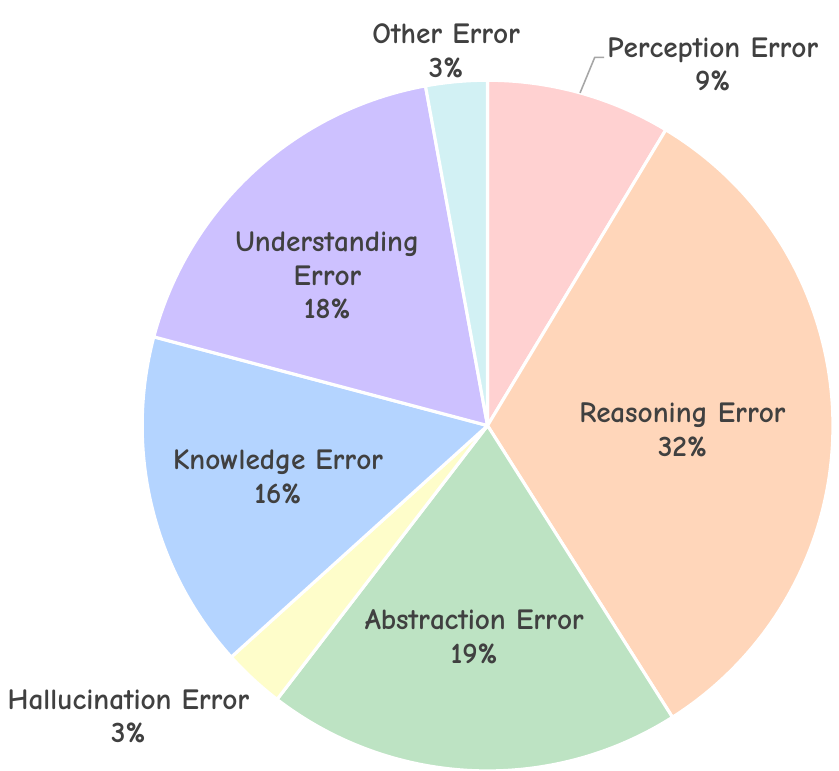}
    \caption{GPT-5}
    \end{subfigure}
    \begin{subfigure}[b]{0.245\textwidth}
    \centering
    \includegraphics[width=\linewidth]{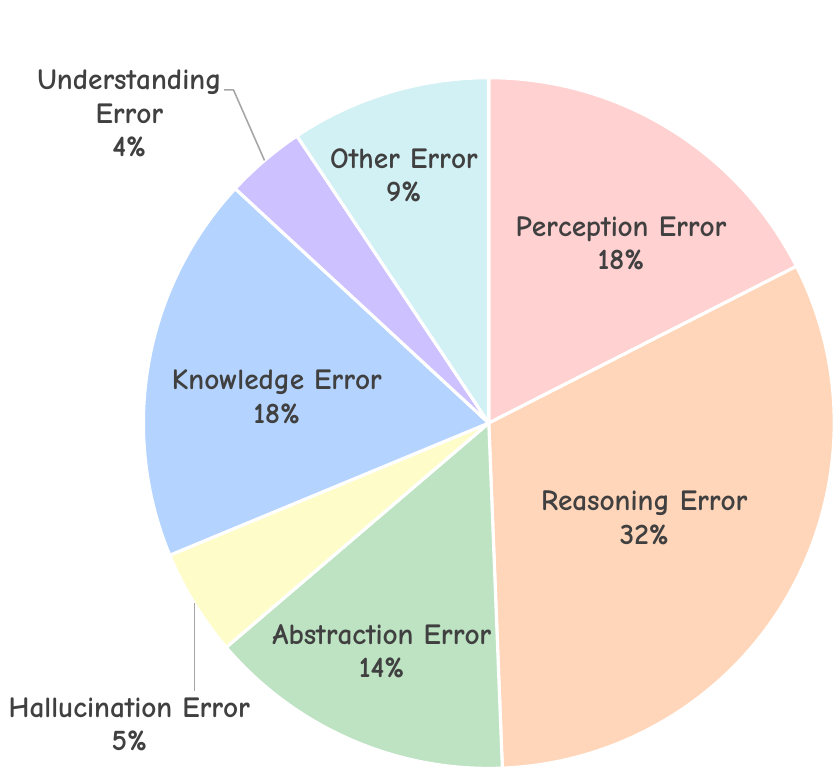}
    \caption{Geimini-2.5-Pro}
    \end{subfigure}
\caption{Error distribution over 140 errors for each model on MMR-Life.}\label{fig:error}
\vspace{-4mm}
\end{wrapfigure}

\section{Error Analysis}
\vspace{-3pt}
This section focuses on the errors made by GPT-5 and Gemini-2.5-Pro, the two strongest models on MMR-Life. For each model, we randomly select 20 incorrect examples from each reasoning type and identify the root causes of the model's erroneous responses. The distribution of these errors is shown in Figure \ref{fig:error}, with a selection of notable 42 cases and detailed analyses provided in the Appendix \ref{append:case}. The results reveal that reasoning errors dominate at 32\%, with the model frequently making basic logical mistakes such as causal inversion (24\%), temporal confusion (42\%), and missing key steps (24\%) during reasoning. 
In addition, abstraction errors (17\%), which reflect the model's short-term thinking capabilities, such as the ability to make associations, are also notable.
Knowledge errors (17\%) and  perception errors (12\%) constitute substantial portions of failures, indicating challenges in recalling the correct knowledge for reasoning, as well as difficulties in identifying static attributes of objects (e.g., color, shape) and dynamic changes (e.g., movement). By systematically examining these failures, we not only expose critical shortcomings in current MLLMs but also derive actionable insights that can inform the next generation of MLLMs.
\vspace{-5pt}
\section{Related Work}
\vspace{-2pt}
\paragraph{Multimodal Reasoning Enhancement Methods.} The development of methods in multimodal reasoning closely follows the approaches established in pure language processing. Inspired by its success in text-only settings, CoT has recently been extended to MLLMs, leading to the development of prompt-guided multimodal reasoning. Studies such as IPVR \citep{prompt_ipvr}, CCoT\citep{prompt_ccot}, and VisualSketchpad \citep{prompt_Visual_Sketchpad} combine reasoning with perception, enhancing the reliability of the reasoning process. After that, the search-based inference method brings reward models into the multimodal reasoning process, training a scoring model to evaluate and select the best reasoning path \citep{reward_visualPRM, reward_ixc2_5, reward_skywork}. Recently, following the success of Deepseek-R1 GRPO \citep{deepseek-r1}, a group of thinking MLLMs like VL-Rethinker \citep{vl-rethinker}, MM-Eureka \citep{mm-eureka}, and MiMo-VL \citep{mimo-vl} have emerged. Our benchmark comprehensively evaluates different methods and MLLMs, aiming to guide their further optimization. 

\vspace{-5pt}
\paragraph{Multimodal Reasoning Benchmarks.} There exists a number of multimodal benchmarks testing MLLMs' reasoning abilities. Several studies combine world knowledge with reasoning and assess the reasoning capabilities of MLLMs across various STEM fields, such as GPQA \citep{gpqa}, OlympiadBench \citep{olympiadbench}, MME-CoT \citep{mme-cot}, MMR-V \citep{mmrv} and MMLU-Reason \citep{mmlu-reason}. Other works argue that reasoning should be decoupled from knowledge, using symbolic patterns to evaluate the model's logical reasoning abilities, such as PuzzleVQA \citep{puzzlevqa}, VisualPuzzles \citep{visualpuzzle}, and MME-Reasoning \citep{mme-reasoning}. However, both types of benchmarks exhibit deviations from real-life reasoning scenarios due to the expert-level knowledge and symbolic images. Although recent work on spatial reasoning meets real-life requirements \citep{spatial1, spatial2, spatial3}, it covers only a limited set of reasoning types. Our MMR-Life benchmark covers seven different reasoning types and introduces real-life multi-image input, addressing former gaps.

\vspace{-5pt}
\section{Conclusion}
\vspace{-5pt}
We present MMR-Life, a novel benchmark designed to evaluate the multimodal reasoning abilities of current MLLMs across seven distinct reasoning types using multiple real-life images as inputs. Through careful and diverse data curation, our dataset provides a comprehensive evaluation of MLLMs' reasoning performance across various real-life scenarios, which shows that existing MLLMs still face significant challenges and exhibit notable performance imbalances across different reasoning types. We conduct a further analysis of the reasoning paradigms of these models, uncovering the relationship between the thinking length, enhancement methods, and reasoning abilities of MLLMs, which lays the foundation for the development of more generalizable AI systems.
\section*{Ethics Statement}
In constructing our benchmark, we ensure strict adherence to copyright and licensing regulations, explicitly avoiding data from sources that prohibit copying or redistribution. Besides, we avoid the images that contain any private information or harmful content. The data in our MMR-Life are not intended to replace, nor are they capable of replacing, the original data source. Therefore, we assert that their inclusion does not affect the market value or utility of the original materials.
We did not employ external crowdsourcing or paid annotation platforms. All participants volunteered, with a complete understanding of the research goals, procedures, and the intended use of the data.
\section*{Reproducibility Statement}
We have taken several steps to improve the reproducibility of our research. Regarding the data, we provide a thorough description of the data sources for each task, along with links, in Appendix \ref{append:data_source}. A subset of 210 items, including the questions and their corresponding images, is also uploaded in the supplementary materials. Additionally, we describe the dataset construction process and the prompts used in both \S\ref{sec:2.2} and Appendix \ref{append:annotation}. On the experimental side, we offer a detailed account of the model versions, parameter settings, and prompts used in the experiments, which are outlined in Appendix \ref{append:main_set}. The full experimental code is also uploaded in the supplementary materials. We commit to making all data and code open source if the paper is accepted.

\section*{Acknowledgement}
This work is supported by the National Natural Science Foundation of China (No. U24A20335, No. 62406321). This work is also supported by Beijing Natural Science Foundation (L243006).
 
\bibliography{iclr2026_conference}
\bibliographystyle{iclr2026_conference}

\appendix
\newtheorem{definition}{Definition}
\newpage
\section{The Use of Large Language Models}
In this study, a large language model (LLM) was employed as a tool to assist in the refinement and enhancement of the manuscript's language. The specific usages of the LLM include:
\begin{itemize}
    \item \textbf{Grammar and Syntax Improvement:} The LLM helped to correct grammatical errors and improve sentence structures, contributing to greater clarity and fluency in the writing.
    \item \textbf{Conciseness and Precision:} It provided suggestions for more concise and precise wording, aiding in the refinement of certain sections without altering their meaning.
\end{itemize}
It is important to note that while the LLM contributed to the refinement of the manuscript's language, the research ideas, data analysis, and conclusions were independently conceived and developed by the authors. The LLM's contributions were exclusively related to text refinement and did not extend to the conceptual aspects of the study.

\section{Key Concepts in MMR-Life}\label{append:concept}
We begin by discussing key concepts in the benchmark to clearly define the core problems and design principles that our work addresses. 
\subsection{Reasoning in Real-life Scenario} As real-life reasoning is a fundamental design principle of our benchmark, we provide a brief definition of it:

\begin{definition}[\textbf{Reasoning in Real-life Scenarios}]
Reasoning in real-life scenarios refers to the process of applying diverse reasoning capabilities to solve problems from everyday situations, which are defined by a set of images and textual descriptions that satisfy the following conditions:
\begin{enumerate}[label=(\roman*)]
  \item \textbf{Multiple natural images:} The input must contain \emph{multiple} images, each depicting 
  objects or events that either objectively exist in real life 
  or are realistically simulated to resemble real-world conditions. Purely abstract diagrams or symbolic renderings are \emph{excluded}.
  \item \textbf{Commonsense solvability:} The answer to the problem must \emph{not rely} on complex domain-specific knowledge. Instead, it should be solvable using only basic human commonsense reasoning and general logic.
\end{enumerate}
\end{definition}
As mentioned in \S\ref{sec:1}, the two existing benchmark types do not fully adhere to the above definition, as they often incorporate unnatural images, such as charts and synthetic puzzles, and may require specialized domain knowledge. In contrast, MMR-Life is constructed in strict accordance with the above definition, emphasizing the evaluation of reasoning in real-life scenarios from the outset. It should be noted that this definition is not intended to be broadly applicable but serves as the guiding principle for the design of this study.

\subsection{Multi-Image vs. Video} In \S\ref{sec:1}, we noted that real-life images are continuous, which led us to adopt multi-image input. However, a natural question arises: why not use continuous videos instead? In the following, we compare and discuss this choice. Overall, we opt not to use video as our input format for the following reasons:
\begin{itemize}
    \item \textbf{\textit{Low Reasoning Types Coverage:}} The relationship between multiple images in a video is typically limited to temporal sequencing. In this context, it is difficult to design reasoning tasks, such as analogy or inductive reasoning, since these tasks often require a parallel relationship between the images, which cannot be fully captured by a video input.
    \item \textbf{\textit{Low Data Diversity:}} From a data perspective, as discussed in \S\ref{sec:2.2}, real-world videos are only a subset of our image sources. If all inputs were required to be videos, we would lose a significant variety of data sources, such as natural photographs, thereby reducing data diversity.
    \item \textbf{\textit{High Noise in Input:}} In video-based benchmarks, frames are typically sampled from videos and input into the model, which can introduce many irrelevant frames that interfere with reasoning. While this setup is closer to real-world scenarios, our benchmark aims to directly assess the model's reasoning abilities, minimizing interference from other factors.
\end{itemize}

\section{Details of Annotation Protocols}\label{append:annotation}
This section presents additional details of our task annotation pipeline and protocols, providing complete details for \S\ref{sec:2.2} of the main paper.

\subsection{Data Sources of Different Tasks} \label{append:data_source}
Table \ref{tab:data_source} presents the data sources for all the tasks included in MMR-Life. During the data collection phase, all annotators strictly adhere to copyright and licensing regulations on the source sites or datasets. Moreover, following Definition 1, we limit the dataset strictly to natural images, explicitly excluding symbolic diagrams and other non-photographic forms.
\begin{table}[tbp]
\centering
\caption{Data sources and image types of different tasks in MMR-Life}\label{tab:data_source}
\resizebox{\linewidth}{!}{
\begin{tabular}{llll}
\toprule
\textbf{Reasoning Type} & \textbf{Task Type} &\textbf{Image Type}  &\textbf{Data Source} \\ \midrule
\multirow{3}{*}{Abductive}  & Human Activity Attribution   & Domestic Life & TVbench \citep{TVbench}    \\ 
  & Character Interaction Attribution & Human Animation & Tom \& Jerry Cartoon \citep{CausalChaos} \\ 
 & Multi-Hop Collision Attribution & Physical Phenomenon & CLEVRER-Humans \citep{Clevrer-humans}  \\ \midrule
\multirow{3}{*}{Analogical} & Animal Relation Inference & Natural Creatures    & Kaggle \citep{animal}   \\ 
 & Product Similarity Inference   & Product Shots & Kaggle \citep{nikolasgegenava_sneakers_classification_2025}    \\ 
& Artwork Style Inference   & Human Artwork   & Kaggle \citep{art}   \\ \midrule
\multirow{3}{*}{Causal}     & Character Interaction Prediction & Human Animation & Tom \& Jerry Cartoon \citep{CausalChaos} \\ 
   & Multi-Hop Collision Prediction & Physical Phenomenon & CLEVRER-Humans \citep{Clevrer-humans} \\ 
  & Counterfactual Fluid Prediction  & Physical Phenomenon  & ContPhy \citep{contphy}\\ \midrule
\multirow{3}{*}{Deductive}  & Material Composition Deduction & Everyday Objects & MathVisa \citep{mathvisa} \\ 
  & Card Winner Deduction    & Game Symbols   & Kaggle \citep{card_game_2025}     \\ 
 & Recipe Step Deduction      &   Daily Dining  & RecipeQA \citep{recipeqa} \\ \midrule
\multirow{3}{*}{Inductive}  & Bird Migration Induction  & Migration Map     & eBird \citep{ebird}  \\ 
  & Plant Disease Induction  & Pathology Photos & Kaggle \citep{plant_diseases_dataset_2024}   \\ 
 & Sport Feature Induction     & Sports Activities    & Kaggle \citep{sport} \\ \midrule
\multirow{3}{*}{Spatial}    & Relative Position Estimation & Interior Views  & ViewSpatial-Bench \citep{ViewSpatial-Bench}    \\ 
   & Camera Rotation Estimation  & Everyday Objects & NAVI \citep{navi}    \\ 
 & Navigation Route Planning   & Interior Views   & ViewSpatial-Bench \citep{ViewSpatial-Bench} \\ \midrule
\multirow{3}{*}{Temporal}   & Crowd Timeline Reconstruction & Crowd Surveillance  & Kaggle \citep{crowd} \\ 
 & Driving Sequence Prediction   & Traffic Scene & Drivingdojo \citep{drivingdojo}  \\ 
  & Human Activity Localization & Domestic Life   & TVBench \citep{TVbench}  \\ \bottomrule
\end{tabular}}

\end{table}

\subsection{Annotation Guidelines} \label{append:anno_guide}
During the annotation of questions and golden answers, all annotators were given the following guidelines:
\begin{itemize}
    \item All questions must contain multiple images (at least two images).
    \item All questions should be written in English.
    \item All questions should be solvable without complex domain-specific knowledge.
    \item The question should not be ambiguous and can be answered with one option.
    \item The questions should adhere to the definitions of the respective reasoning types (see \S\ref{sec:2.2}), ensuring clear differentiation between tasks of different reasoning types.
\end{itemize}

\subsection{Prompts for Negative Option Generation} \label{append:neg_prompt}
\begin{figure}[htbp]
  \centering
    \includegraphics[width=0.9\linewidth]{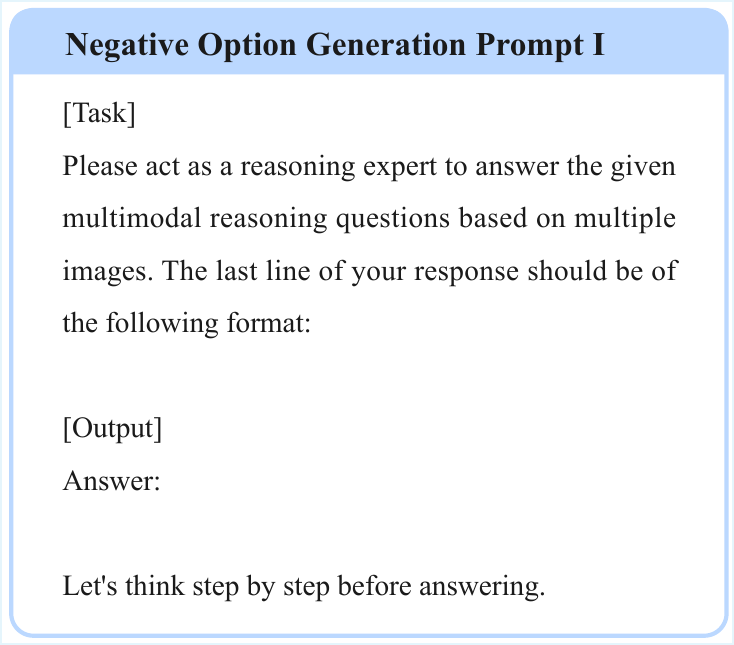}
    \caption{Negative option generation prompt.} \label{fig:nega_prompt1}
\end{figure}

\begin{figure}[htbp]
  \centering
    \includegraphics[width=0.9\linewidth]{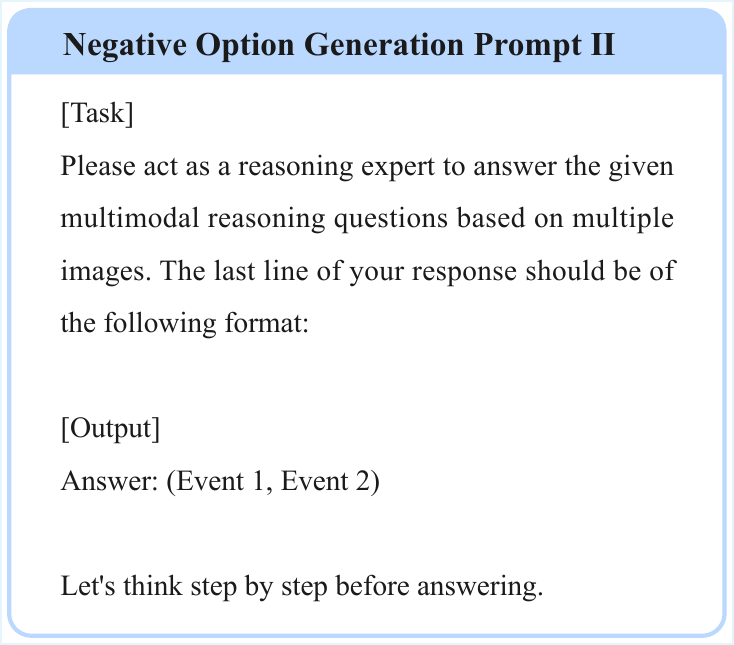}
    \caption{Negative option generation prompt.} \label{fig:nega_prompt2}
\end{figure}

\begin{figure}[htbp]
  \centering
    \includegraphics[width=0.9\linewidth]{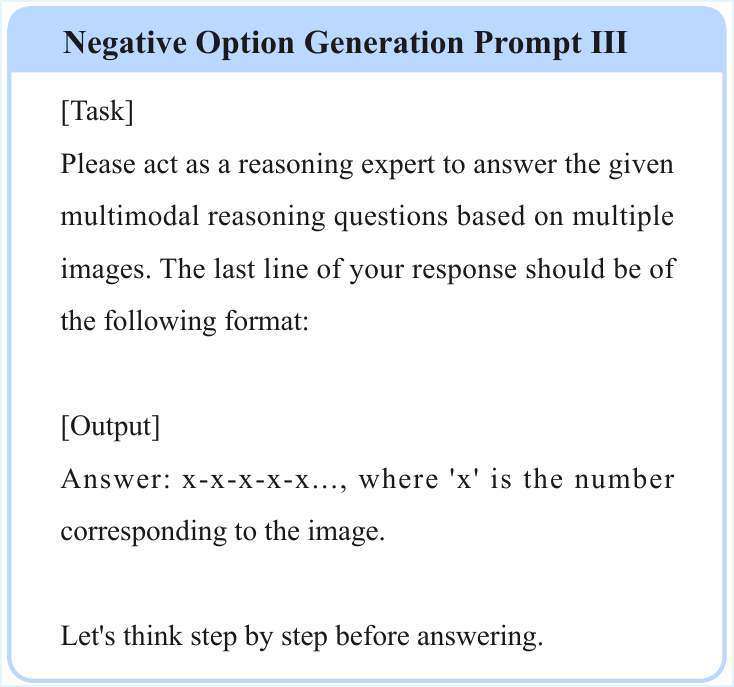}
    \caption{Negative option generation prompt.} \label{fig:nega_prompt3}
\end{figure}

\begin{figure}[htbp]
  \centering
    \includegraphics[width=0.9\linewidth]{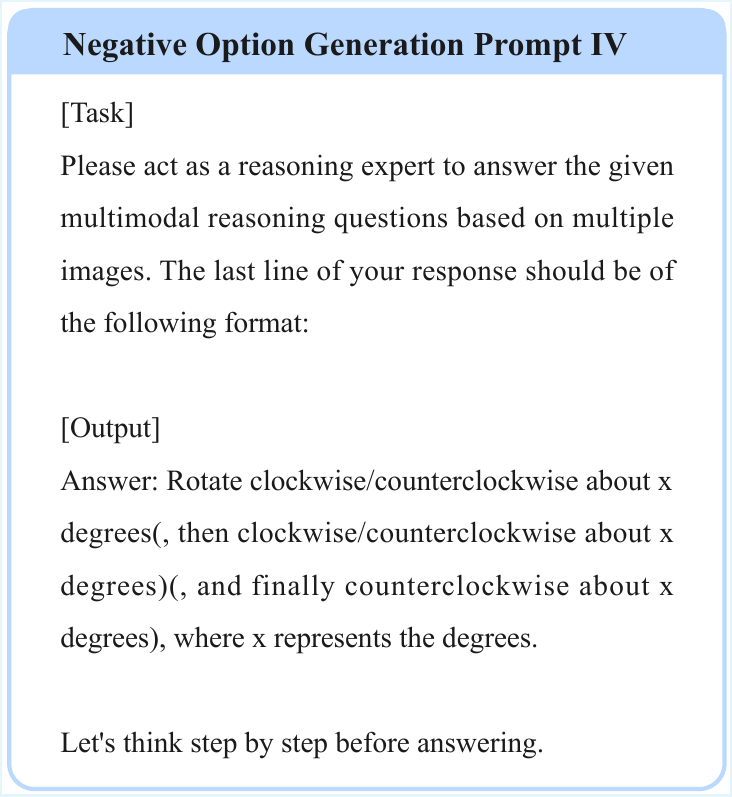}
    \caption{Negative option generation prompt.} \label{fig:nega_prompt4}
\end{figure}

\begin{figure}[htbp]
  \centering
    \includegraphics[width=0.9\linewidth]{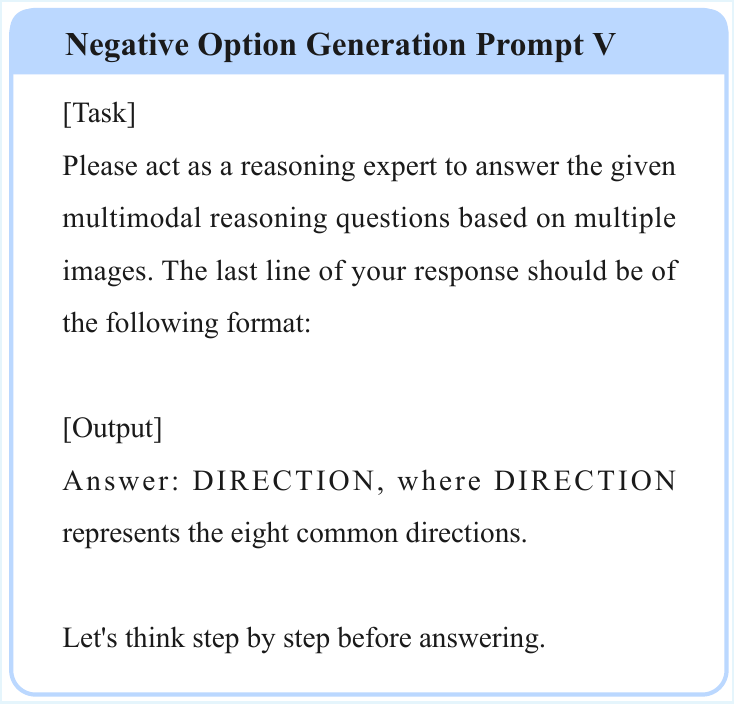}
    \caption{Negative option generation prompt.} \label{fig:nega_prompt5}
\end{figure}

\begin{figure}[htbp]
  \centering
    \includegraphics[width=0.9\linewidth]{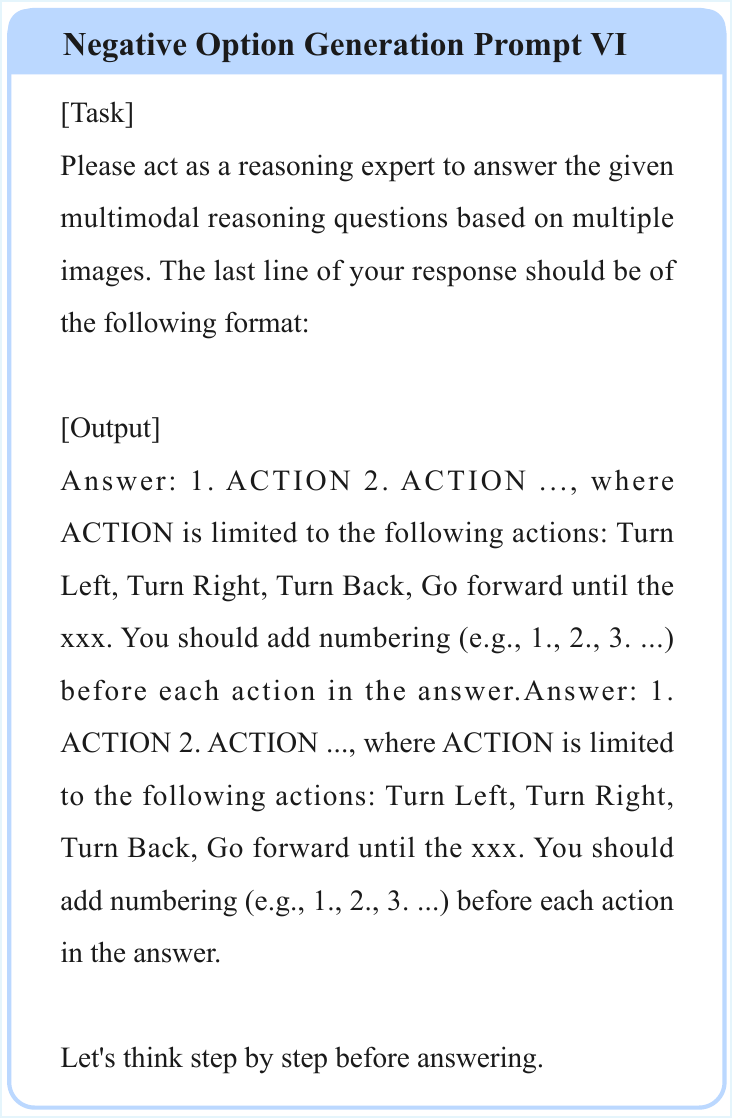}
    \caption{Negative option generation prompt.} \label{fig:nega_prompt6}
\end{figure}

\begin{figure}[htbp]
  \centering
    \includegraphics[width=0.9\linewidth]{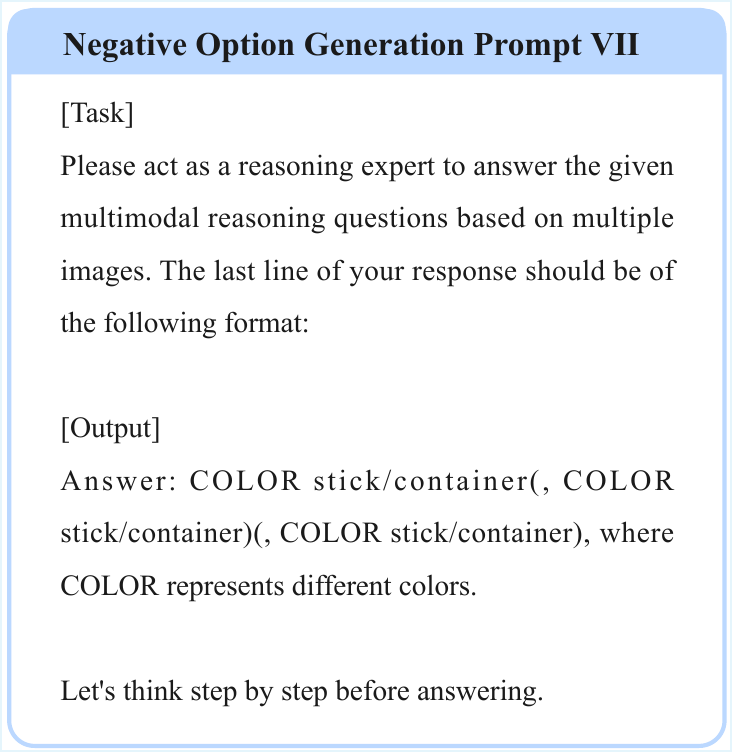}
    \caption{Negative option generation prompt.} \label{fig:nega_prompt7}
\end{figure}

We list our negative option generation prompts from Figure \ref{fig:nega_prompt1} to Figure \ref{fig:nega_prompt7}.

\section{Data Diversity of MMR-Life}
\begin{figure}[htbp]
  \centering
    \includegraphics[width=\linewidth]{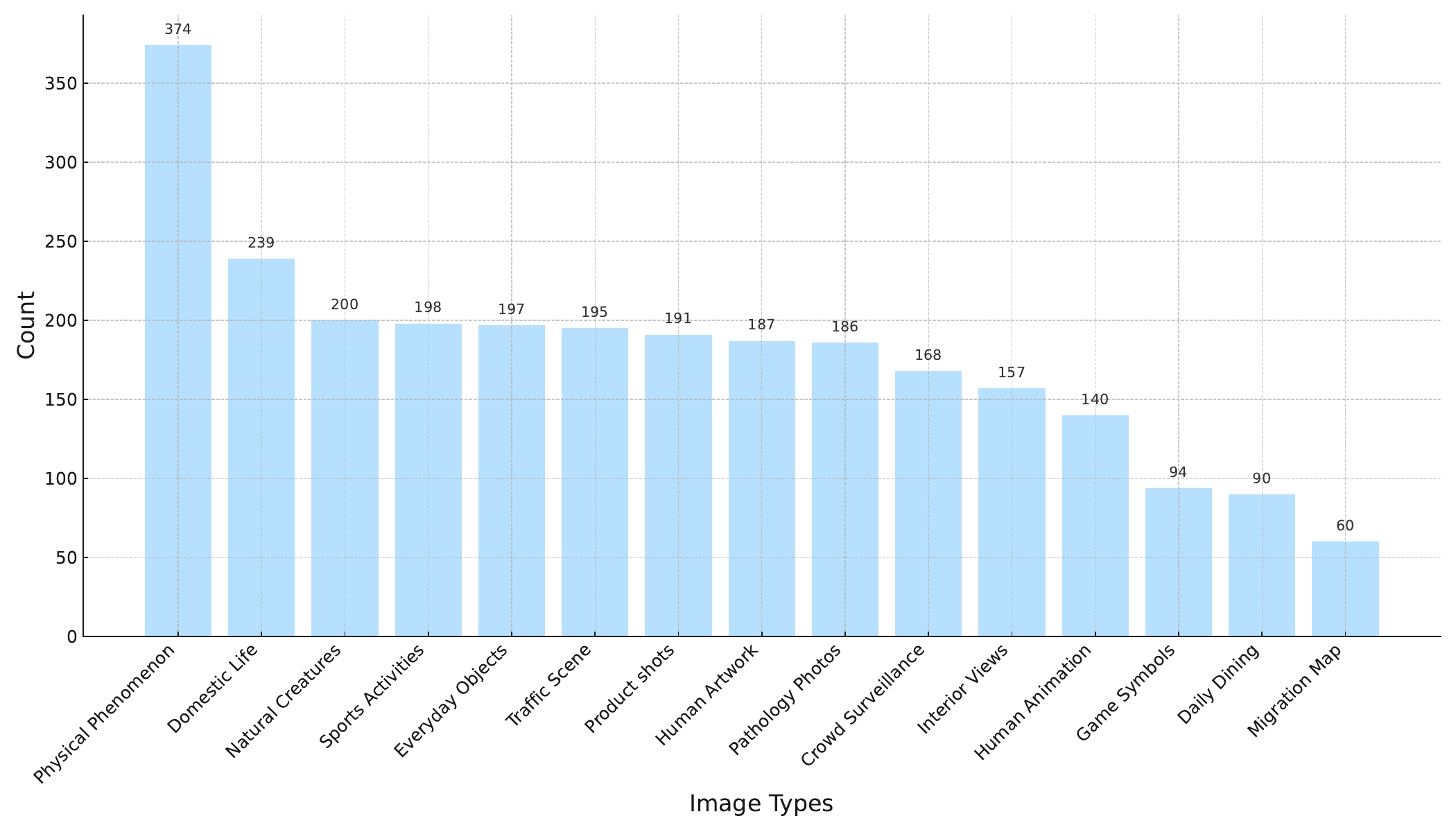}
    \caption{Image type distribution in MMR-Life.} \label{fig:img_type}
\end{figure}

\begin{figure}[htbp]
  \centering
    \includegraphics[width=\linewidth]{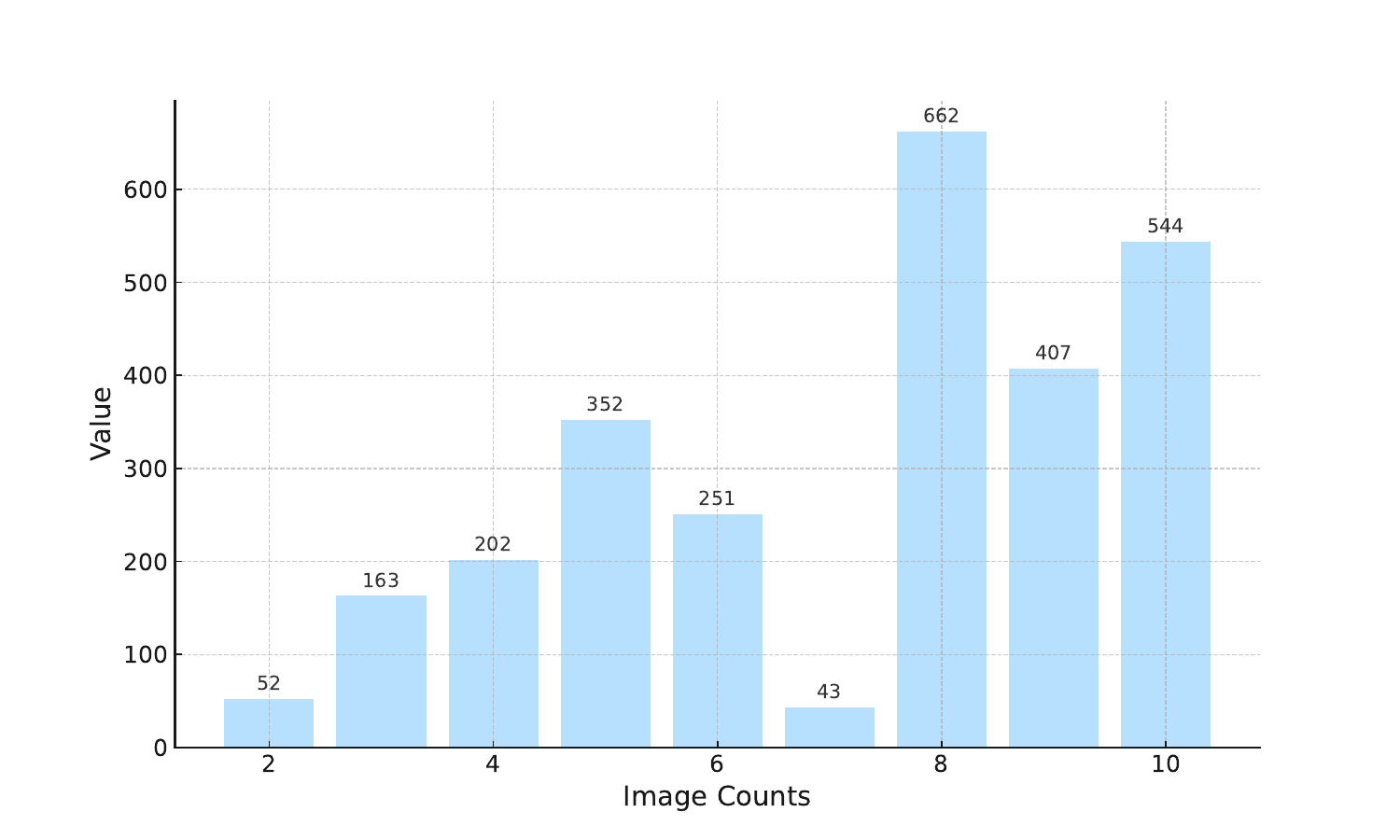}
    \caption{Image counts distributions in MMR-Life.} \label{fig:img_count}
\end{figure}

We demonstrate the diversity of data in MMR-Life in this section, where Figure \ref{fig:img_type} visualizes the variety of image types and Figure \ref{fig:img_count} presents the distribution of input image counts. 
The various types of tasks included in our study are illustrated in Appendix \ref{append:task}.

\section{Task Details} \label{append:task}
In this section, we give a detailed description of each task presented in MMR-Life.
\subsection{Abductive Reasoning}
\subsubsection{Human Activity Attribution}
\paragraph{Task Description.} 
This task tests a model's reasoning about human behavior motivations. By observing people's behavior in a given context, the model must analyze environmental clues and behavior cues to select the most plausible motivation among candidate explanations.
\paragraph{Examples.}
See Figure \ref{haa_correct}, \ref{haa_error}.
\subsubsection{Character Interaction Attribution}
\paragraph{Task Description.} 
This task requires the model to understand causal relationships between characters (e.g., in Tom \& Jerry). Given a scene of interaction, the model must analyze character behaviors and situational factors to infer the most reasonable cause for a specific event or outcome.
\paragraph{Examples.}
See Figure \ref{cia_correct}, \ref{cia_error}.
\subsubsection{Multi-Hop Collision Attribution}
\paragraph{Task Description.} 
This task assesses a model's causal reasoning in complex physical collision chains. In scenes involving multiple objects colliding, the model must trace the collision chain and identify the root cause or triggering event for a given outcome.
\paragraph{Examples.}
See Figure \ref{mca_correct}, \ref{mca_error}.
\subsection{Analogical Reasoning}
\subsubsection{Animal Relation Inference}
\paragraph{Task Description.} 
This task requires models to understand visual analogical relationships between animals. Given three animal images, the model must recognize the relational pattern between the first two animals and then select a fourth animal from the options so that the relation between the third and fourth animals matches the original pattern.
\paragraph{Examples.}
See Figure \ref{ari_correct}, \ref{ari_error}.

\subsubsection{Product Similarity Inference}
\paragraph{Task Description.} 
This task assesses a model's reasoning about product style preference. Based on a person's owned or disliked product samples, the model must analyze design features and style preferences to recommend, from candidate options, a product that best suits their intentions or tastes.
\paragraph{Examples.}
See Figure \ref{psi_correct}, \ref{psi_error}.

\subsubsection{Artwork Style Inference}
\paragraph{Task Description.} 
This task evaluates a model's understanding and recognition of artistic style. Given multiple sample works from the same artist, the model must learn the distinctive stylistic features and then identify which candidate option is most likely also created by that artist.
\paragraph{Examples.}
See Figure \ref{asi_correct}, \ref{asi_error}.

\subsection{Causal Reasoning}
\subsubsection{Character Interaction Prediction}
\paragraph{Task Description.} 
This task tests a model's ability to predict outcomes of interactions between animated characters. Given a specific behavior or event by a character, the model must use contextual understanding and character relations to predict the most likely follow-up reaction or result.
\paragraph{Examples.}
See Figure \ref{cip_correct}, \ref{cip_error}.

\subsubsection{Multi-Hop Collision Prediction}
\paragraph{Task Description.} 
Given a sequence of consecutive images capturing object motion from initial to current time, the model must reason about the underlying physics and simulate possible multi-stage collision propagation, ultimately predicting the most likely next collision event or chain reaction.
\paragraph{Examples.}
See Figure \ref{mcp_correct}, \ref{mcp_error}.
\subsubsection{Counterfactual Fluid Prediction}
\paragraph{Task Description.} 
This task examines a model's counterfactual reasoning ability in fluid dynamics. The model must analyze how a fluid flows and, if a barrier is removed, predict how the flow would change (i.e., determine the altered flow paths) and final positions under the new condition.
\paragraph{Examples.}
See Figure \ref{cfp_correct}, \ref{cfp_error}.

\subsection{Deductive Reasoning}
\subsubsection{Material Composition Deduction}
\paragraph{Task Description.} 
This task requires complex combinatorial reasoning about material composition. Given different types and quantities of material components and the material requirements for certain products, the model must calculate how many products can be produced under the current material constraints.
\paragraph{Examples.}
See Figure \ref{mcd_correct}, \ref{mcd_error}.

\subsubsection{Card Winner Deduction}
\paragraph{Task Description.} 
This task examines a model's understanding of Texas Hold 'em poker rules and logical reasoning. In a multiplayer poker game, each player has hole cards and there are community cards on the board; based on these, the model must analyze the best possible hand for each player and determine the winner.
\paragraph{Examples.}
See Figure \ref{cwd_correct}, \ref{cwd_error}.

\subsubsection{Recipe Step Deduction}
\paragraph{Task Description.} 
This task requires understanding the logical order of cooking processes. Given a dish name and a set of unordered images depicting stages of preparation, the model must deduce the correct cooking sequence based on ingredient states, tool usage, and causal relationships.
\paragraph{Examples.}
See Figure \ref{rsd_correct}, \ref{rsd_error}.

\subsection{Inductive Reasoning}
\subsubsection{Bird Migration Induction}
\paragraph{Task Description.} 
This task requires the model to analyze temporal distribution changes of birds. By observing how bird distributions change over past years, the model must infer migration patterns and predict the likely distribution in the upcoming year.
\paragraph{Examples.}
See Figure \ref{bmi_correct}, \ref{bmi_error}.
\subsubsection{Plant Disease Induction}
\paragraph{Task Description.} 
This task evaluates a model's ability to learn disease patterns in plants. Given samples of leaves afflicted with a particular disease, the model must learn the visual features and then identify which candidate leaves also suffer from the same disease.
\paragraph{Examples.}
See Figure \ref{pdi_correct}, \ref{pdi_error}.
\subsubsection{Sport Feature Induction}
\paragraph{Task Description.} 
This task tests the model's ability to induce patterns in sports characteristics. Given a series of images depicting sports with certain patterns or rules, the model must understand the characteristic relationships and choose the next sport that best matches the pattern.
\paragraph{Examples.}
See Figure \ref{sfi_correct}, \ref{sfi_error}.

\subsection{Spatial Reasoning}

\subsubsection{Relative Position Estimation}
\paragraph{Task Description.} 
This task tests a model's spatial relationship reasoning. Given the relative positions of some objects in an indoor scene, the model must infer the relative positions of others and judge directional relationships (e.g. east, west, north, south).
\paragraph{Examples.}
See Figure \ref{rpe_correct}, \ref{rpe_error}.
\subsubsection{Camera Rotation Estimation}
\paragraph{Task Description.} 
This task requires the model to analyze viewpoint changes between consecutive images. By comparing the same scene from different angles in the image sequence, the model must accurately estimate the camera's rotation angles and directions at each step.

\paragraph{Examples.}
See Figure \ref{cre_correct}, \ref{cre_error}.

\subsubsection{Navigation Route Planning}
\paragraph{Task Description.} 
This task tests a model's spatial reasoning and path planning capability. A robot must navigate in a given indoor environment from a start point to a goal point. Only 90° or 180° turns and forward moves are allowed, and obstacles must be avoided. The model must plan the correct sequence of moves.
\paragraph{Examples.}
See Figure \ref{nrp_correct}, \ref{nrp_error}.

\subsection{Temporal Reasoning}
\subsubsection{Crowd Timeline Reconstruction}
\paragraph{Task Description.} 
This task assesses a model's understanding of temporal sequences in complex scenes. Given a set of unordered images of crowd activities, the model must use cues from people's positions, actions, and environmental changes to infer the correct chronological order.
\paragraph{Examples.}
See Figure \ref{ctr_correct}, \ref{ctr_error}.
\subsubsection{Driving Sequence Prediction}
\paragraph{Task Description.} 
This task evaluates a model's ability to predict time-varying driving scenes. Given a sequence of images from a front-facing cockpit (driver's perspective) view, the model must integrate road geometry, vehicle motions, traffic participants, and environmental cues to predict the most likely next frame.
\paragraph{Examples.}
See Figure \ref{dsp_correct}, \ref{dsp_error}.
\subsubsection{Human Activity Localization}
\paragraph{Task Description.} 
This task asks the model to locate when in a video sequence a particular human activity occurs. Given a video and a description of an activity, the model must precisely predict which time segment (start, middle, end, or throughout) the activity takes place.
\paragraph{Examples.}
See Figure \ref{hal_correct}, \ref{hal_error}.


\section{Details of Main Experiment}
\subsection{Detailed Experimental Setup} \label{append:main_set}
\paragraph{Multimodal Language Models.} Here, we list all the models used in our experiment and provide the corresponding version (if available):  \textit{gpt-5-2025-08-07} \citep{gpt-5}, \textit{gpt-5-mini-2025-08-07} \citep{gpt-5}, \textit{gpt-4.1-2025-04-14} \citep{gpt-4.1}, \textit{gpt-4.1-mini-2025-04-14} \citep{gpt-4.1},  \textit{gpt-4o-2024-11-20} \citep{gpt-4o}, \textit{gpt-4o-mini-2024-07-18} \citep{gpt-4o}, \textit{o4-mini-2025-04-16} \citep{o4}, \textit{claude-sonnet-4-20250514} \citep{claude4}, \textit{claude-3-7-sonnet-20250219} \citep{claude-3.7}, \textit{gemini-2.5-flash} \citep{gemini-2.5}, \textit{gemini-2.5-pro} \citep{gemini-2.5}, \textit{doubao-1-5-vision-pro-32k} \citep{doubao-1.5-vision}, \textit{Kimi-VL-A3B-Thinking-2506} \citep{kimi-vl}, \textit{Keye-VL-1.5-8B} \citep{keye-vl}, \textit{MiMo-VL-7B-RL-2508} \citep{mimo-vl}, \textit{MiMo-VL-7B-SFT-2508} \citep{mimo-vl}, \textit{MM-Eureka-Qwen-7B} \citep{mm-eureka}, \textit{MM-Eureka-Qwen-32B} \citep{mm-eureka}, \textit{OpenVLThinker-7B-v1.2} \citep{openvl}, \textit{OpenVLThinker-7B-v1.2-sft-iter3} \citep{openvl}, \textit{Qwen2.5-VL-7B-Instruct} \citep{qwen-2.5-vl}, \textit{Qwen2.5-VL-32B-Instruct} \citep{qwen-2.5-vl}, \textit{Qwen2.5-VL-72B-Instruct} \citep{qwen-2.5-vl},\textit{R1-Onevision-7B} \citep{r1-onevision}, \textit{R1-Onevision-7B-RL} \citep{r1-onevision}, \textit{Skywork-R1V-38B} \citep{skyworkr1v}, \textit{VL-Rethinker-7B} \citep{vl-rethinker}, \textit{VL-Rethinker-32B} \citep{vl-rethinker}, \textit{VL-Rethinker-72B} \citep{vl-rethinker}, \textit{InternVL3.5-8B} \citep{internvl-3.5}, \textit{InternVL3.5-30B-A3B} \citep{internvl-3.5}, \textit{InternVL3.5-38B} \citep{internvl-3.5}, \textit{gemma-3-4b-it} \citep{gemma3}, \textit{gemma-3-12b-it} \citep{gemma3}, \textit{gemma-3-27b-it} \citep{gemma3}, \textit{QVQ-72B-Preview} \citep{qvq}. 

\paragraph{Parameters.} For parameters during the model's inference. We set the temperature to 0.5, top p to 0.5, and seed to 17.

\begin{figure}[htbp]
  \centering
    \includegraphics[width=0.9\linewidth]{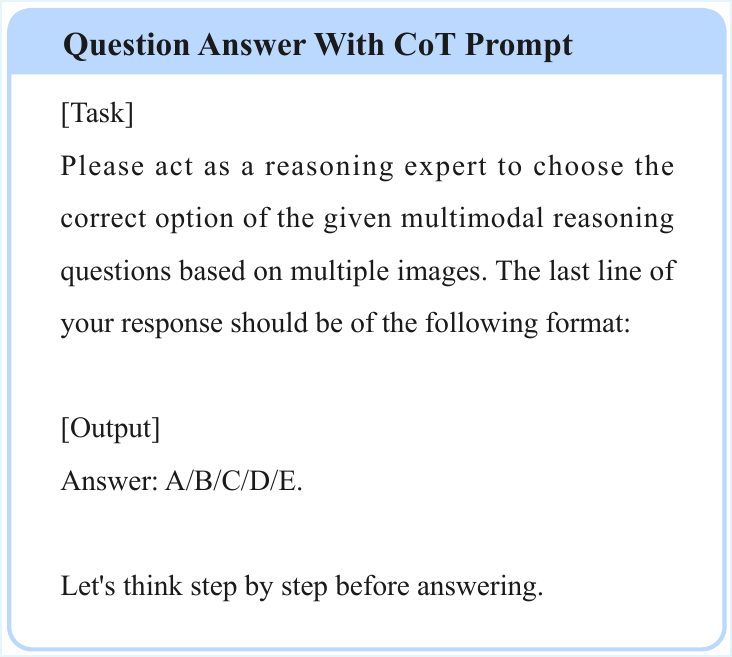}
    \caption{Prompt used in the main experiment.} \label{fig:cot_prompt}
\end{figure}

\paragraph{Prompts.} The prompt used in the main experiments are illustrated in Figure \ref{fig:cot_prompt}.

\subsection{Full Experimental Results} \label{append:main_result}
We demonstrate full evaluation results on 37 MLLMs in Table \ref{tab:full}.

\subsection{Experimental Results on Tiny Set} \label{append:tiny_result}
We present the model performance comparison on the mini test set in Table \ref{tab:mini}.

\section{Details of Thinking Pattern Analysis} \label{append:thinking}
For the base setting, we use MiMo-VL-7B-SFT, RL-Onevision, Qwen-2.5-VL-32B and Qwen-2.5-VL-7B (with CoT prompting). For the RL setup, we use the model corresponding to the RL training version for CoT:  MiMo-VL-7B-RL, RL-Onevision-RL, MM-Eureka-32B, and VL-Rethinker-7B. 
These models are trained on various datasets to illustrate the generalizability of our conclusions.

\section{Case Study}\label{append:case}
\begin{figure}[htbp]
  \centering
    \includegraphics[width=0.95\linewidth]{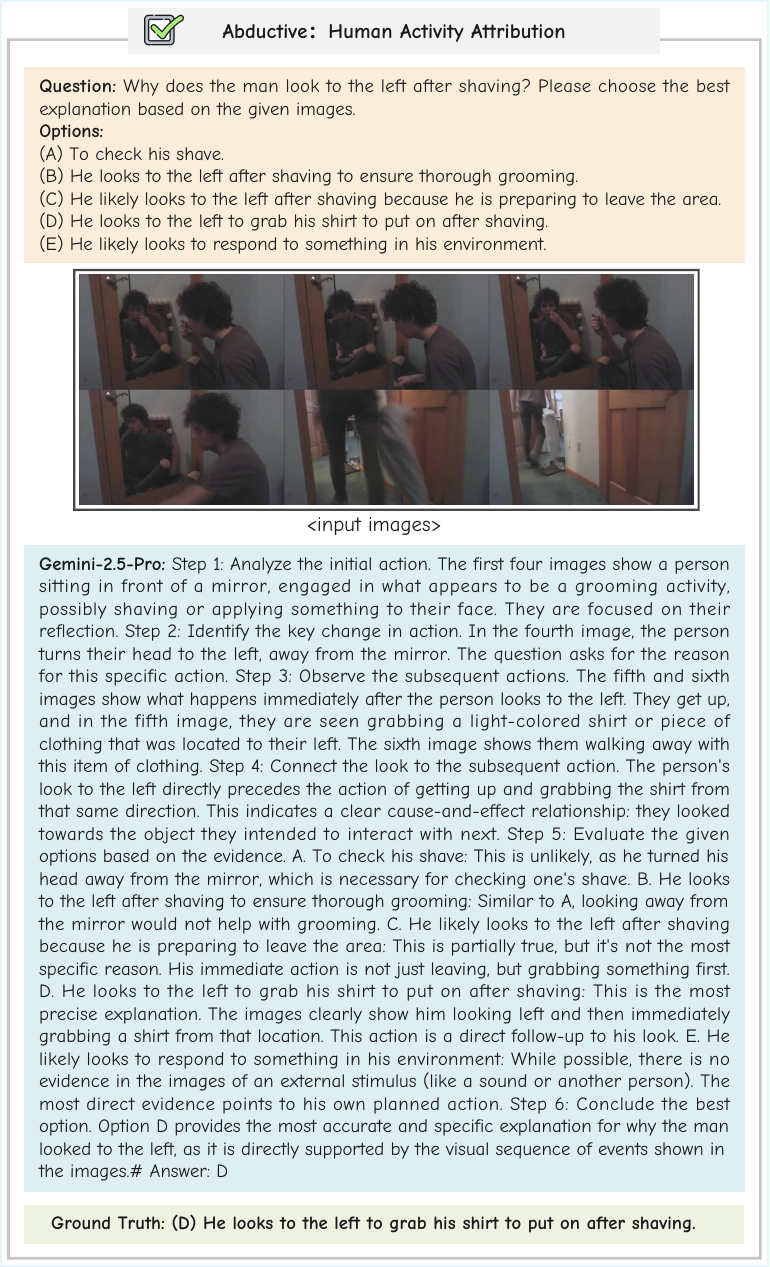}
    \caption{A correct example of Human Activity Attribution task.} 
    \label{haa_correct}
\end{figure}

\begin{figure}[htbp]
  \centering
    \includegraphics[width=\linewidth]{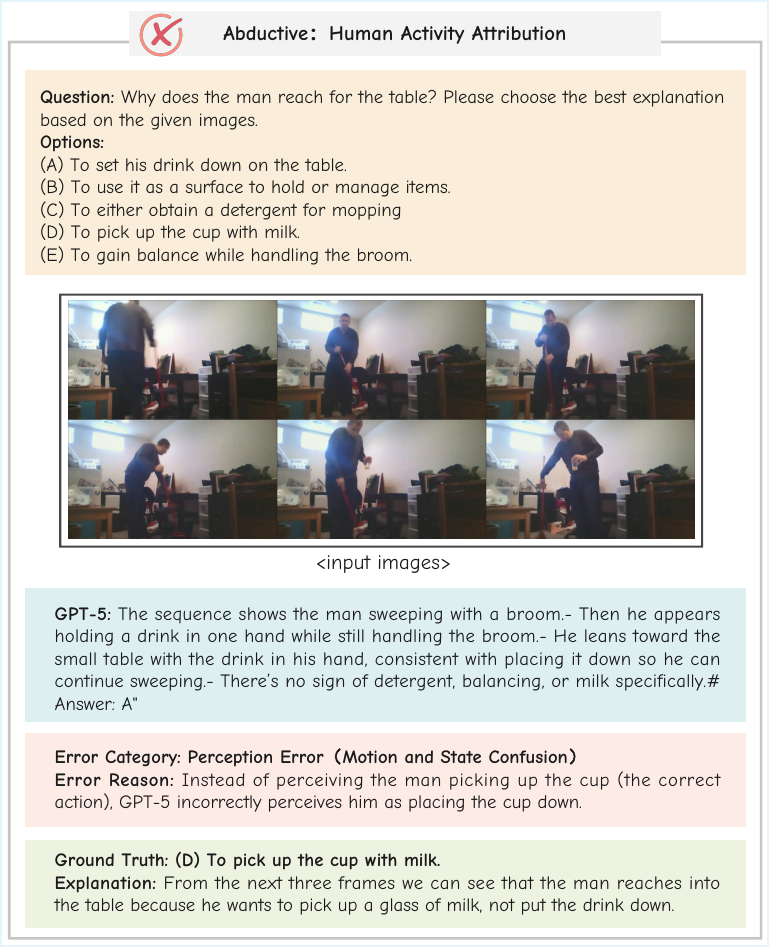}
    \caption{An error example of Human Activity Attribution task.}
    \label{haa_error}
\end{figure}

\begin{figure}[htbp]
  \centering
    \includegraphics[width=0.8\linewidth]{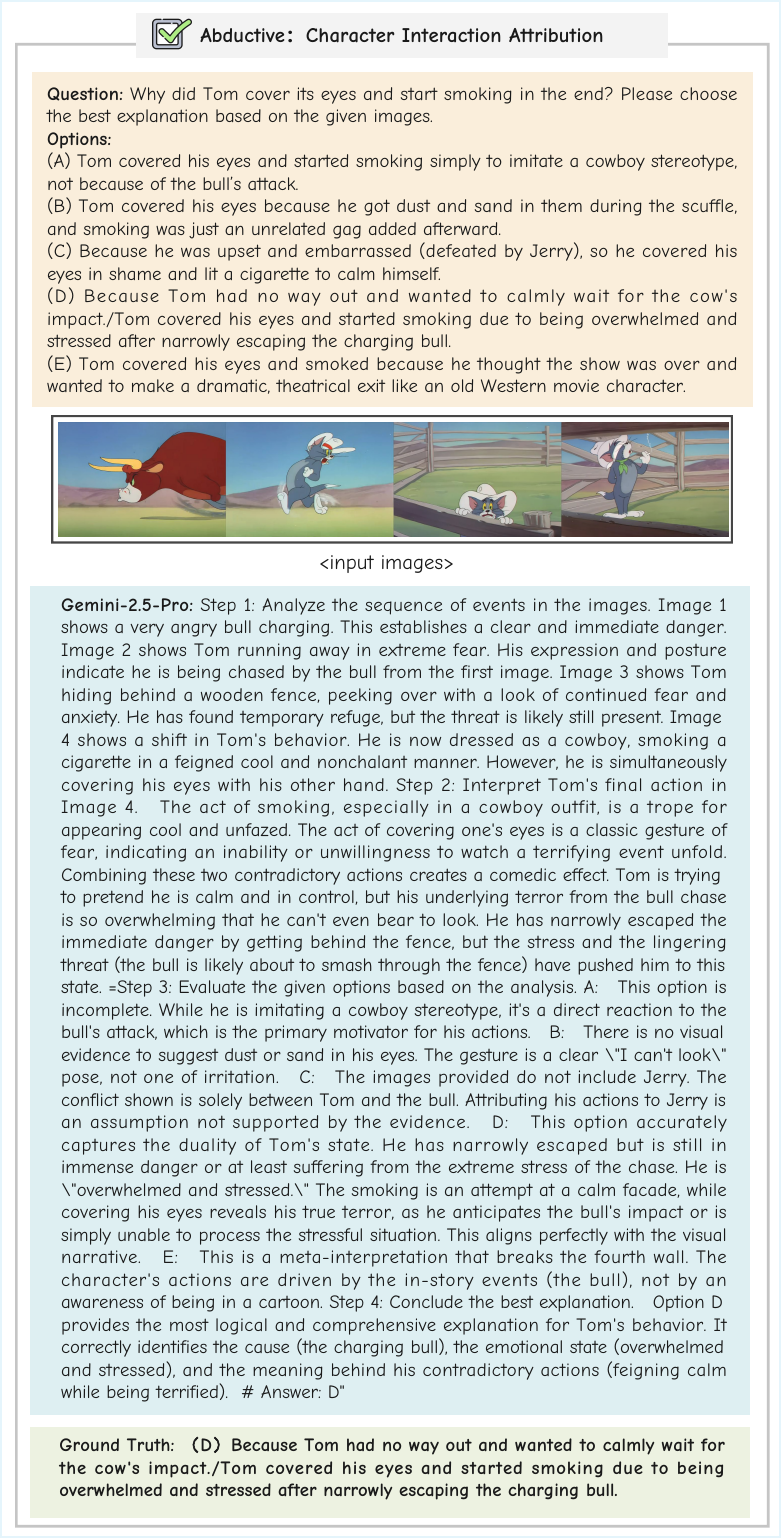}
    \caption{A correct example of Character Interaction Attribution task.}
    \label{cia_correct}
\end{figure}

\begin{figure}[htbp]
  \centering
\includegraphics[width=0.8\linewidth]{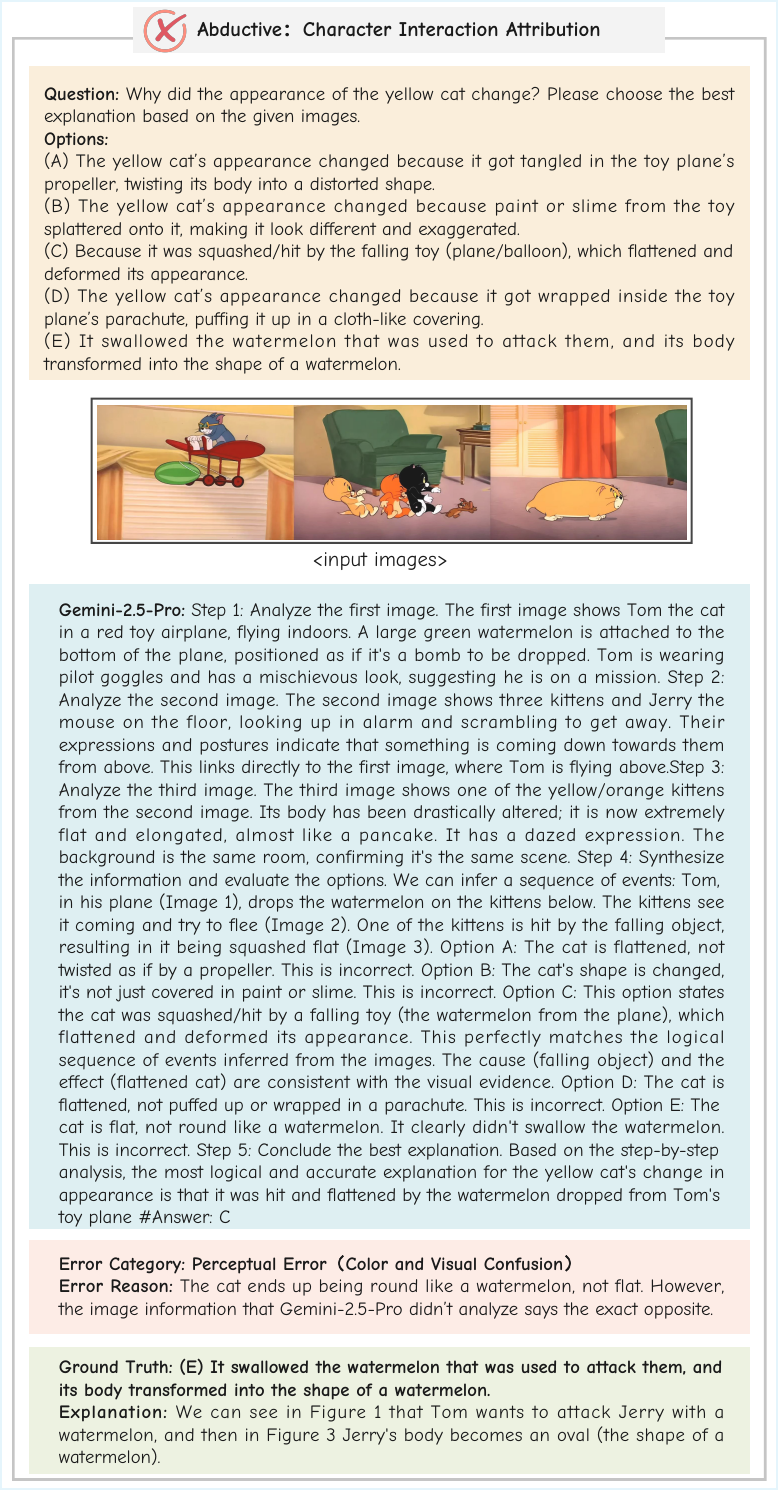}
    \caption{An error example of Character Interaction Attribution task.}
    \label{cia_error}
\end{figure}

\begin{figure}[htbp]
  \centering
    \includegraphics[width=\linewidth]{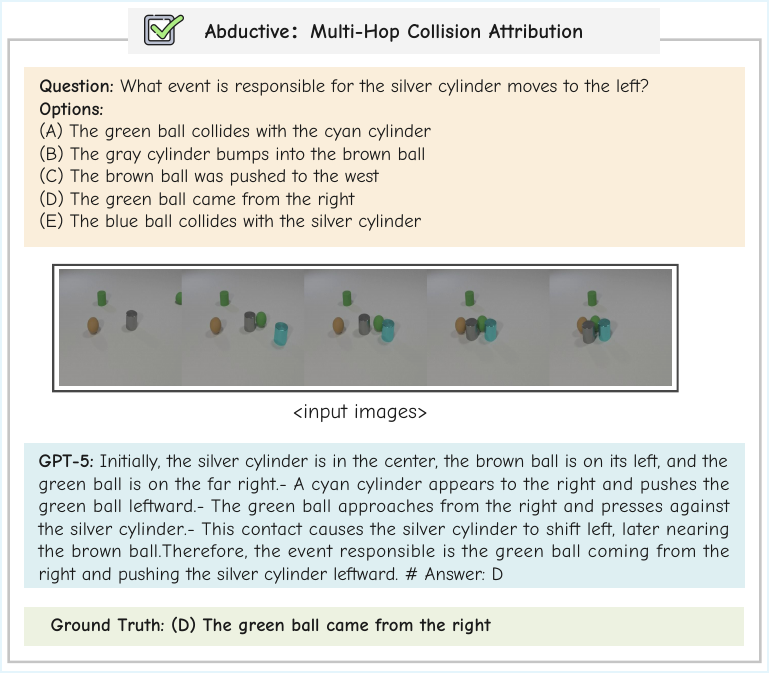}
    \caption{A correct example of Multi-Hop Collision Attribution task.} \label{mca_correct}
\end{figure}

\begin{figure}[htbp]
  \centering
    \includegraphics[width=\linewidth]{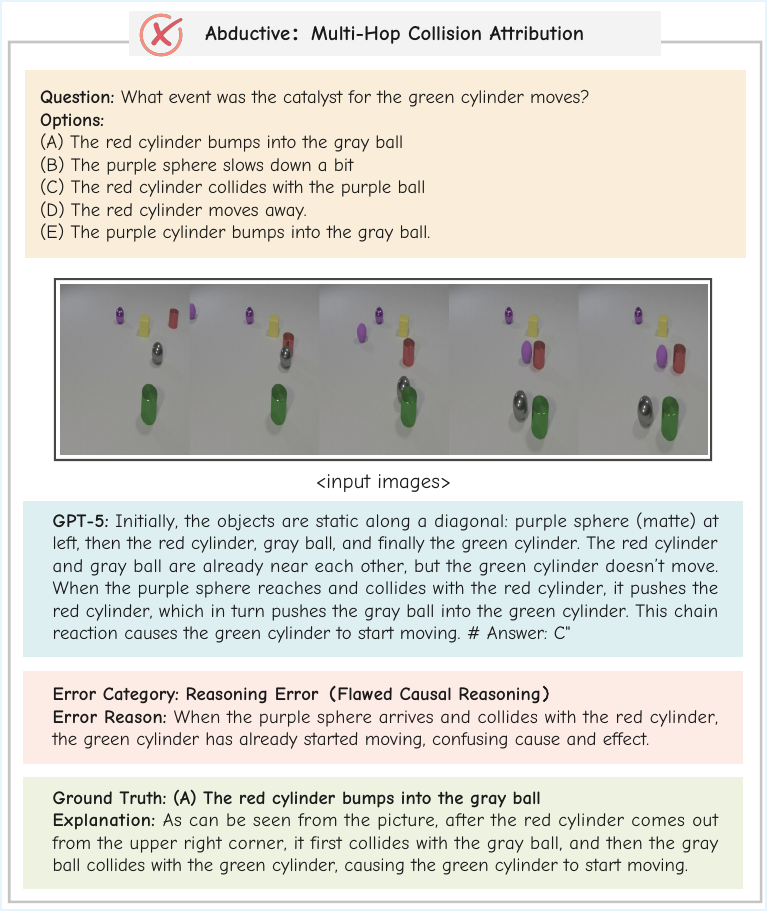}
    \caption{An error example of Multi-Hop Collision Attribution task.} \label{mca_error}
\end{figure}

\begin{figure}[htbp]
  \centering
    \includegraphics[width=0.9\linewidth]{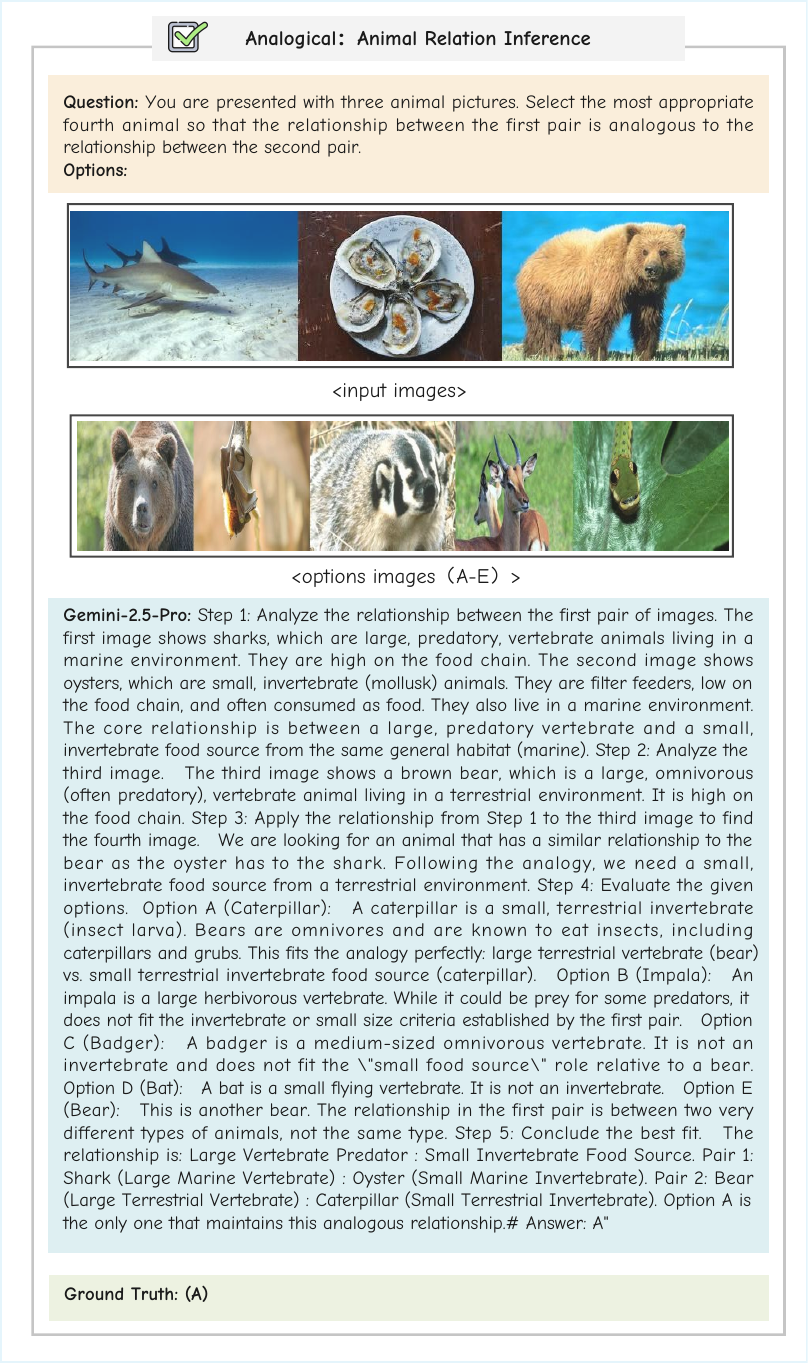}
    \caption{A correct example of Animal Relation Inference task.} \label{ari_correct}
\end{figure}

\begin{figure}[htbp]
  \centering
    \includegraphics[width=0.9\linewidth]{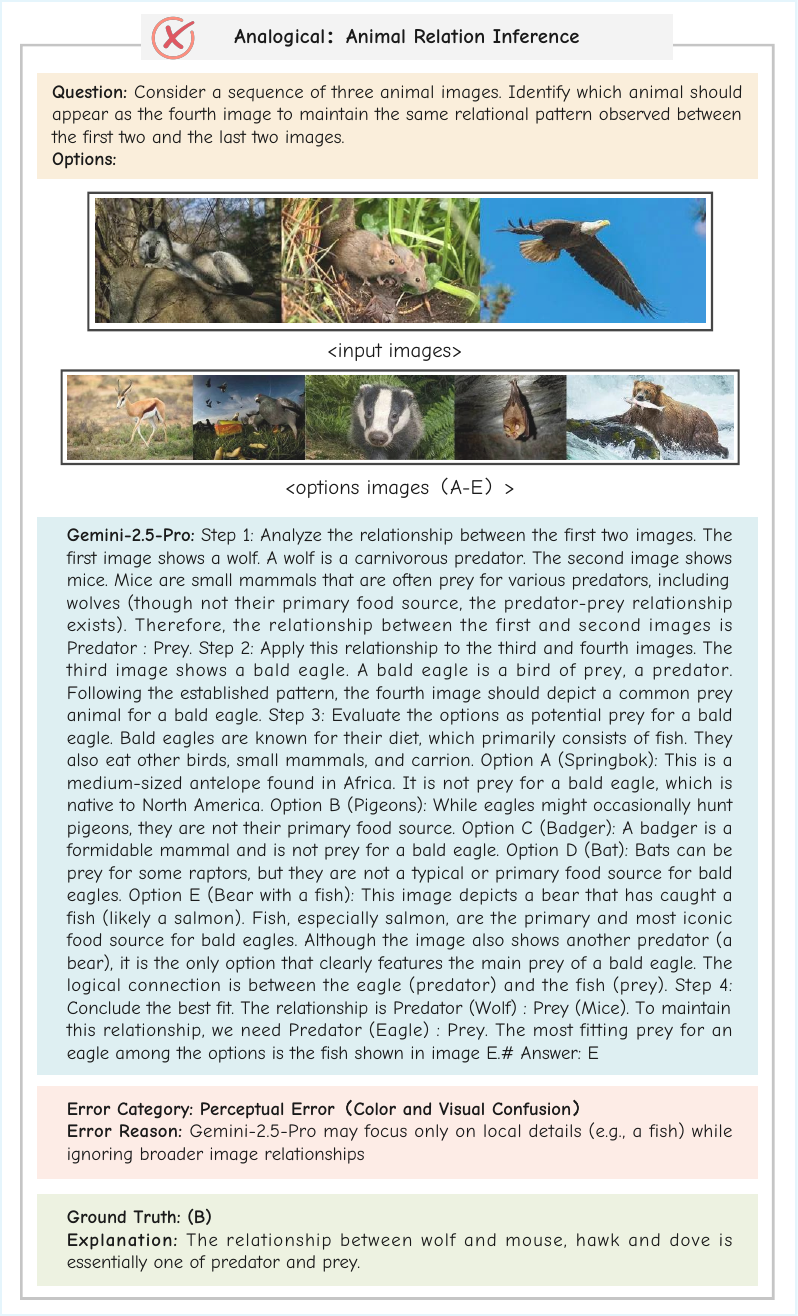}
    \caption{An error example of Animal Relation Inference task.} \label{ari_error}
\end{figure}

\begin{figure}[htbp]
  \centering
    \includegraphics[width=\linewidth]{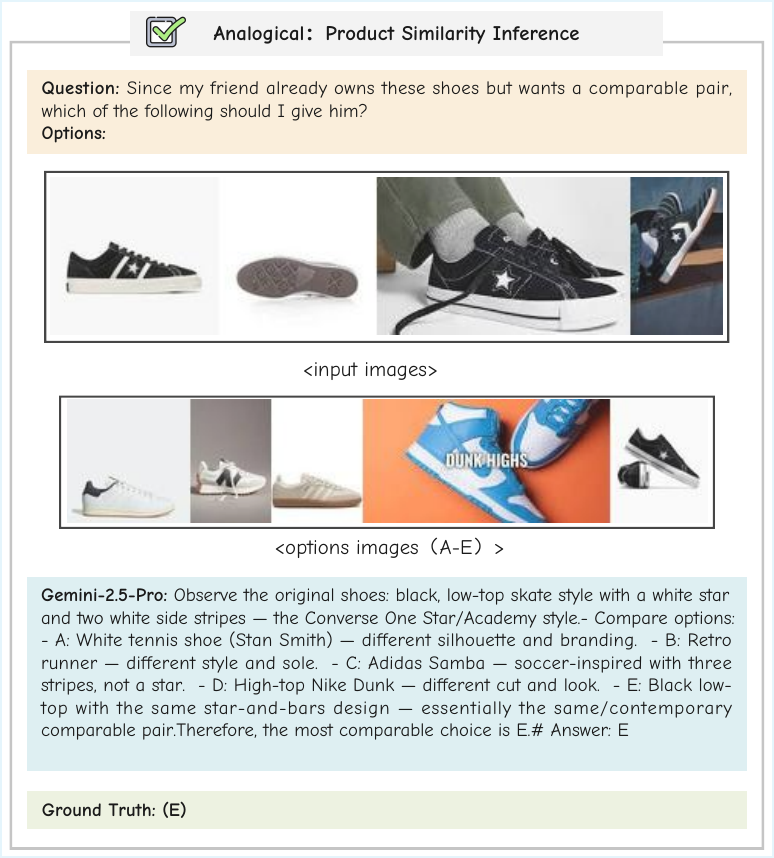}
     \caption{A correct example of Product Similarity Inference task.} \label{psi_correct}
\end{figure}

\begin{figure}[htbp]
  \centering
    \includegraphics[width=\linewidth]{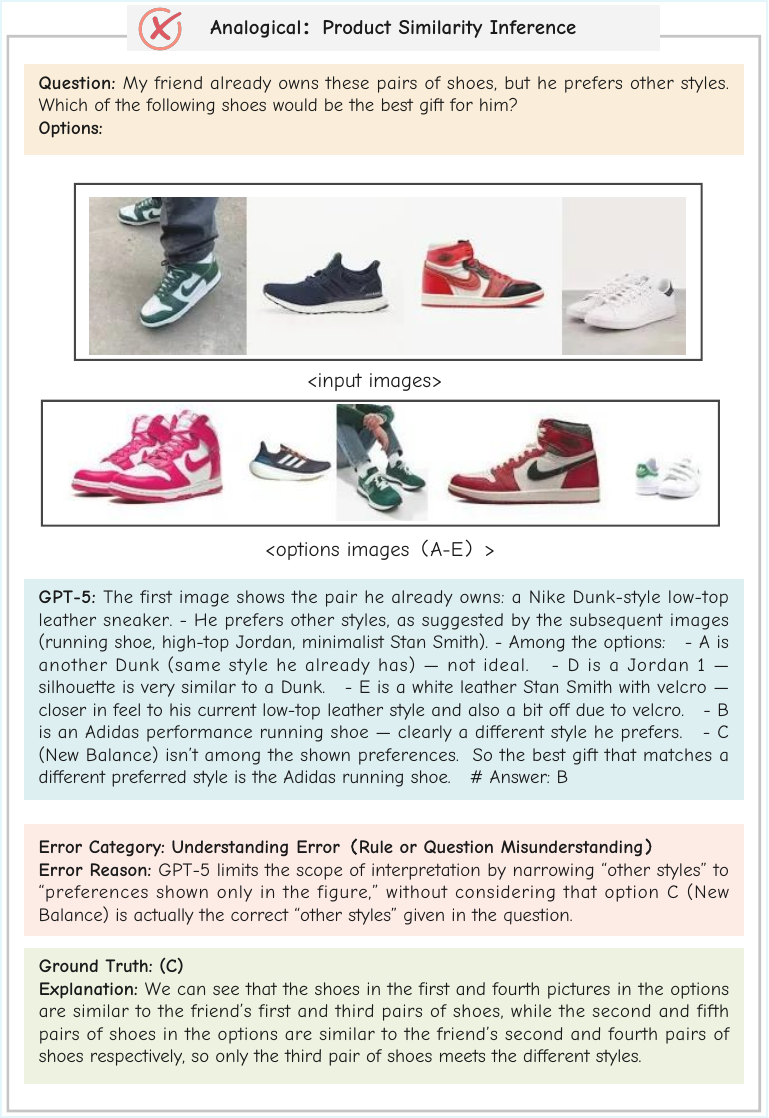}
    \caption{An error example of Product Similarity Inference task.} \label{psi_error}
\end{figure}

\begin{figure}[htbp]
  \centering
    \includegraphics[width=\linewidth]{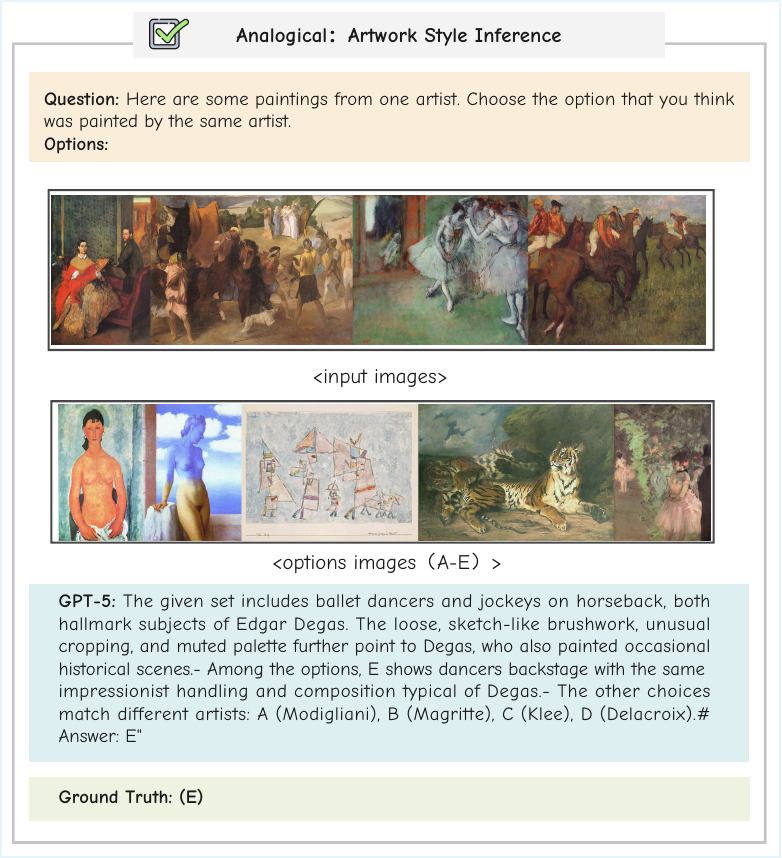}
    \caption{A correct example of Artwork Style Inference task.} \label{asi_correct}
\end{figure}

\begin{figure}[htbp]
  \centering
    \includegraphics[width=\linewidth]{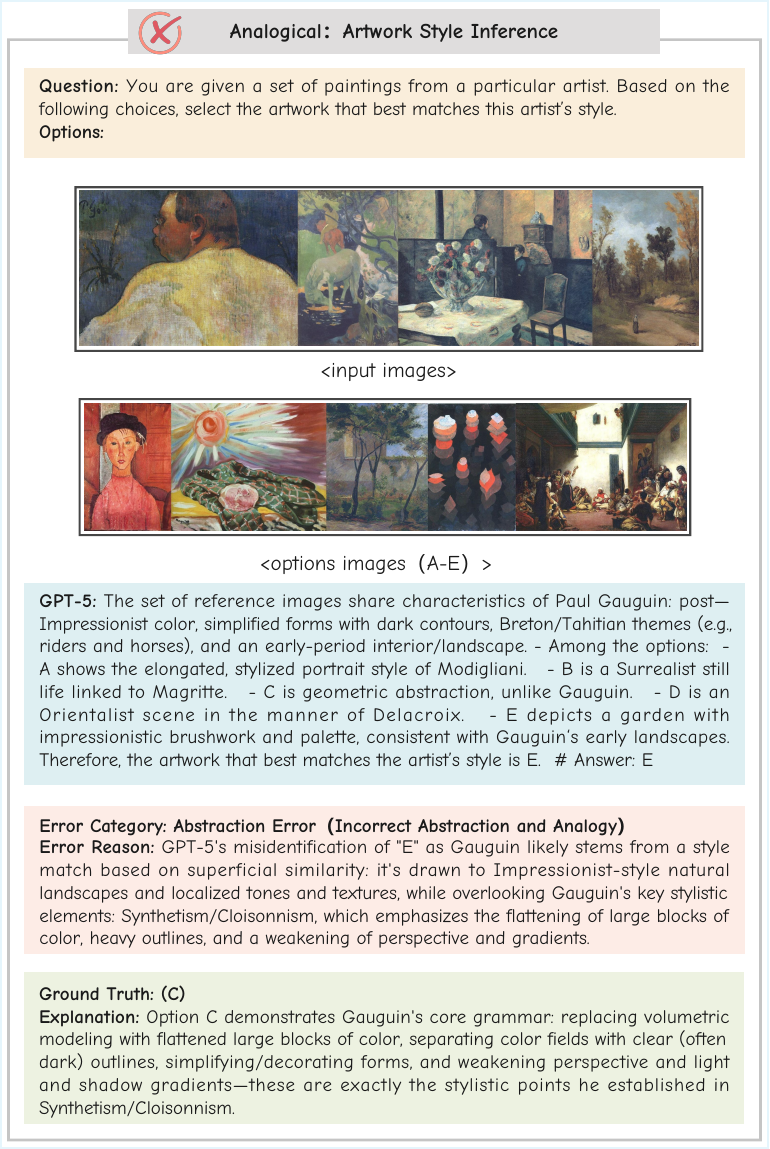}
    \caption{An error example of Artwork Style Inference task.} \label{asi_error}
\end{figure}

\begin{figure}[htbp]
  \centering
    \includegraphics[width=\linewidth]{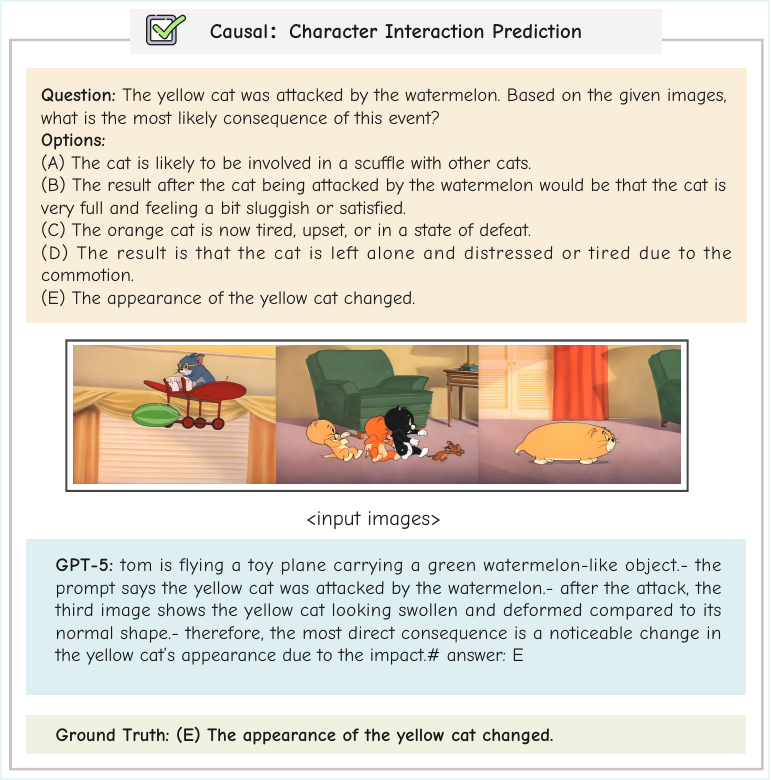}
    \caption{A correct example of Character Interaction Prediction task.} \label{cip_correct}
\end{figure}

\begin{figure}[htbp]
  \centering
    \includegraphics[width=\linewidth]{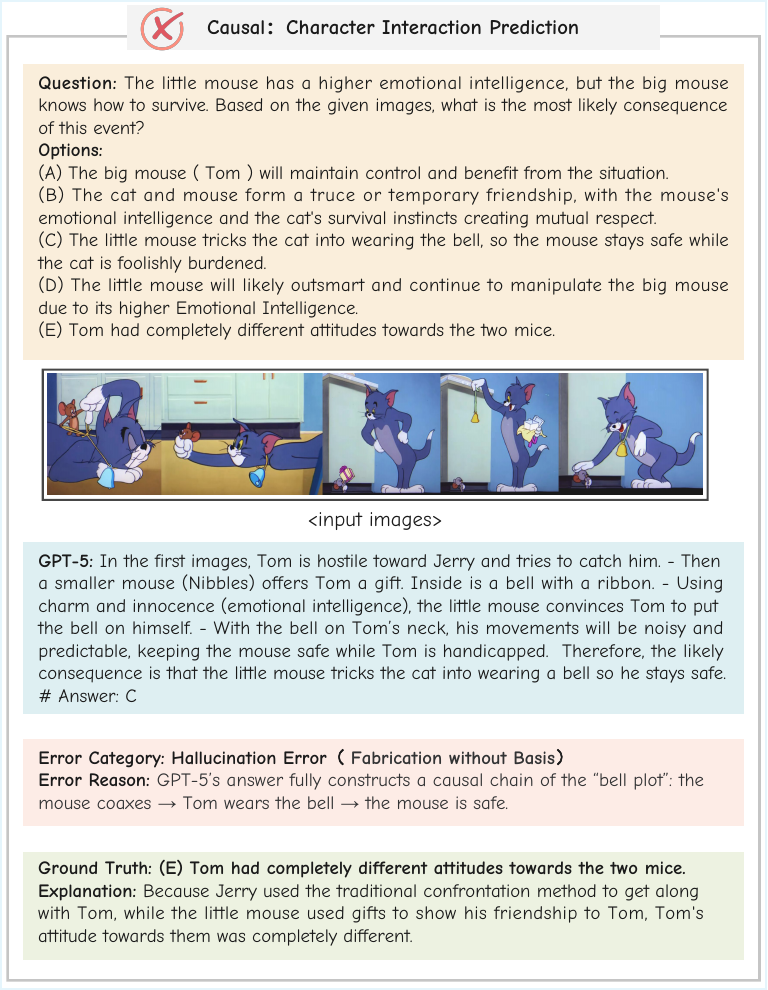}
    \caption{An error example of Character Interaction Prediction task.} \label{cip_error}
\end{figure}

\begin{figure}[htbp]
  \centering
    \includegraphics[width=0.95\linewidth]{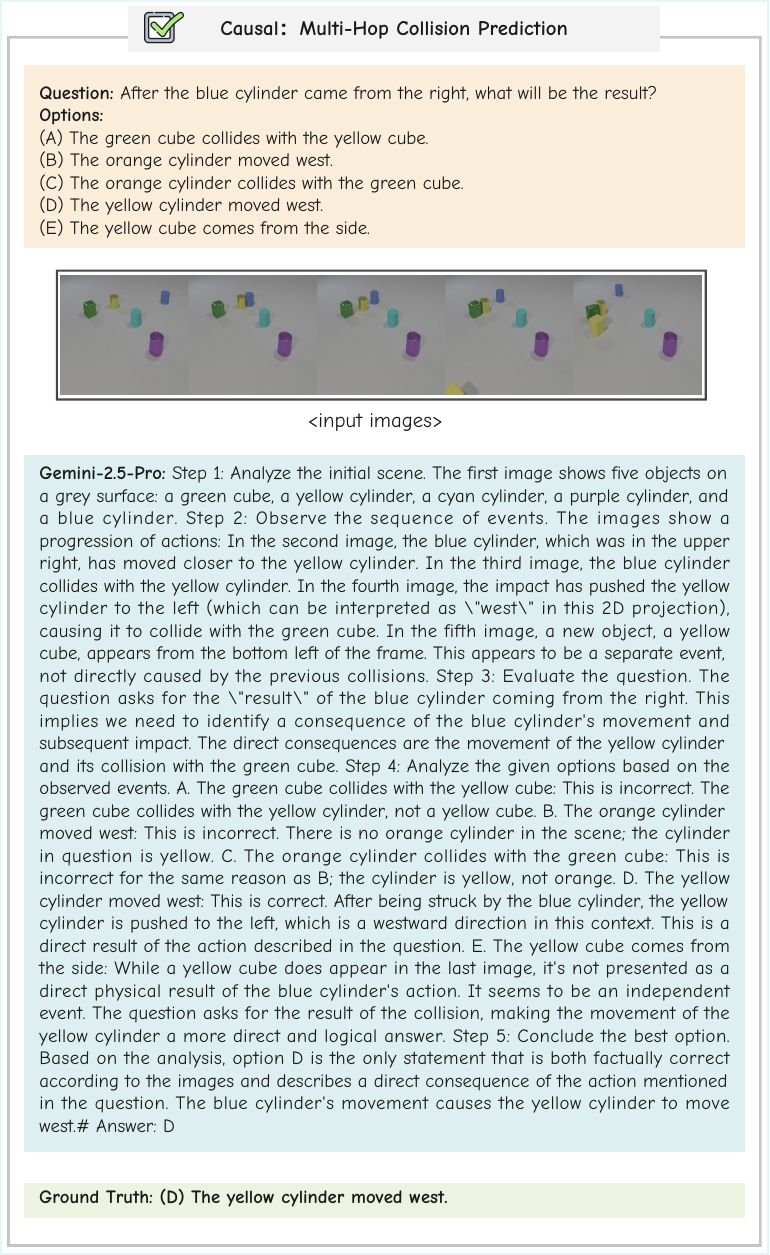}
    \caption{A correct example of Multi-Hop Collision Prediction task.} \label{mcp_correct}
\end{figure}

\begin{figure}[htbp]
  \centering
    \includegraphics[width=\linewidth]{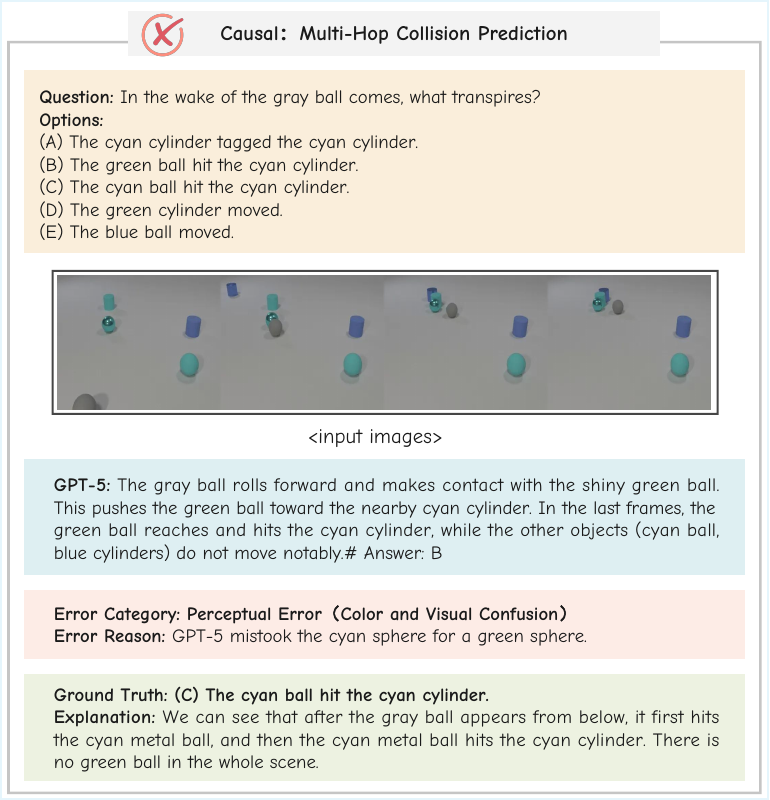}
    \caption{An error example of Multi-Hop Collision Prediction task.} \label{mcp_error}
\end{figure}

\begin{figure}[htbp]
  \centering
    \includegraphics[width=\linewidth]{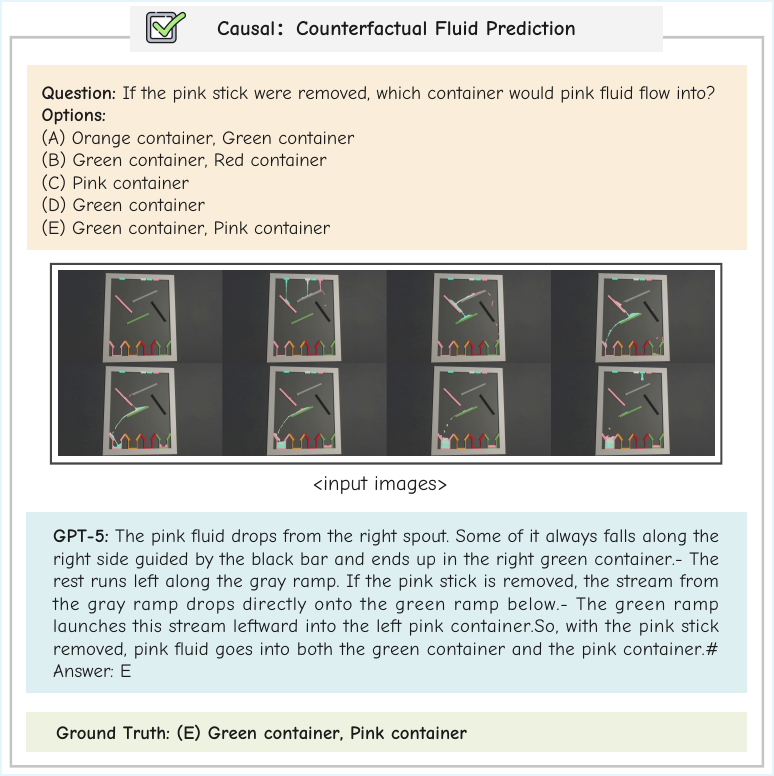}
   \caption{A correct example of Counterfactual Fluid Prediction task.} \label{cfp_correct}
\end{figure}

\begin{figure}[htbp]
  \centering
    \includegraphics[width=0.8\linewidth]{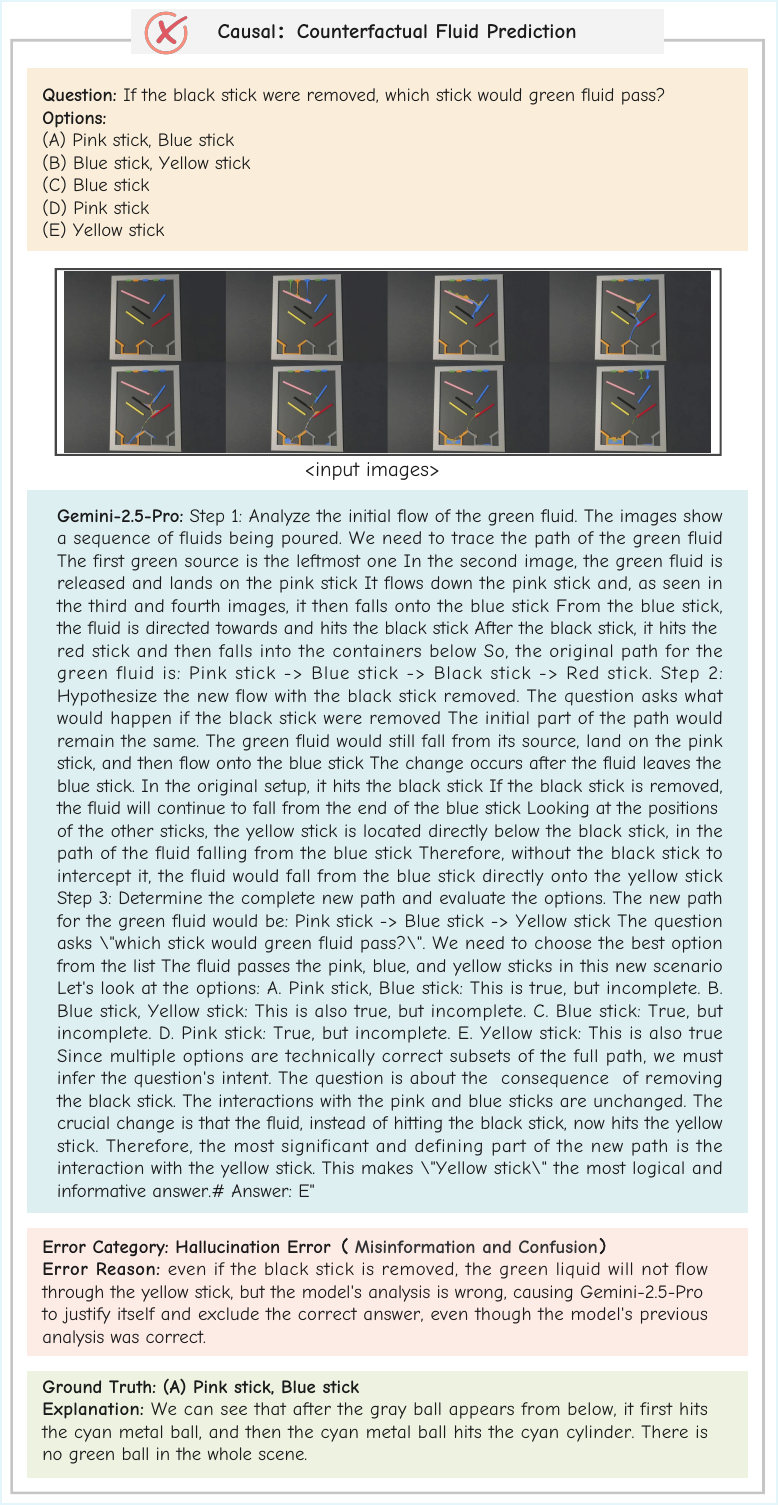}
    \caption{An error example of Counterfactual Fluid Prediction task.} \label{cfp_error}
\end{figure}

\begin{figure}[htbp]
  \centering
    \includegraphics[width=\linewidth]{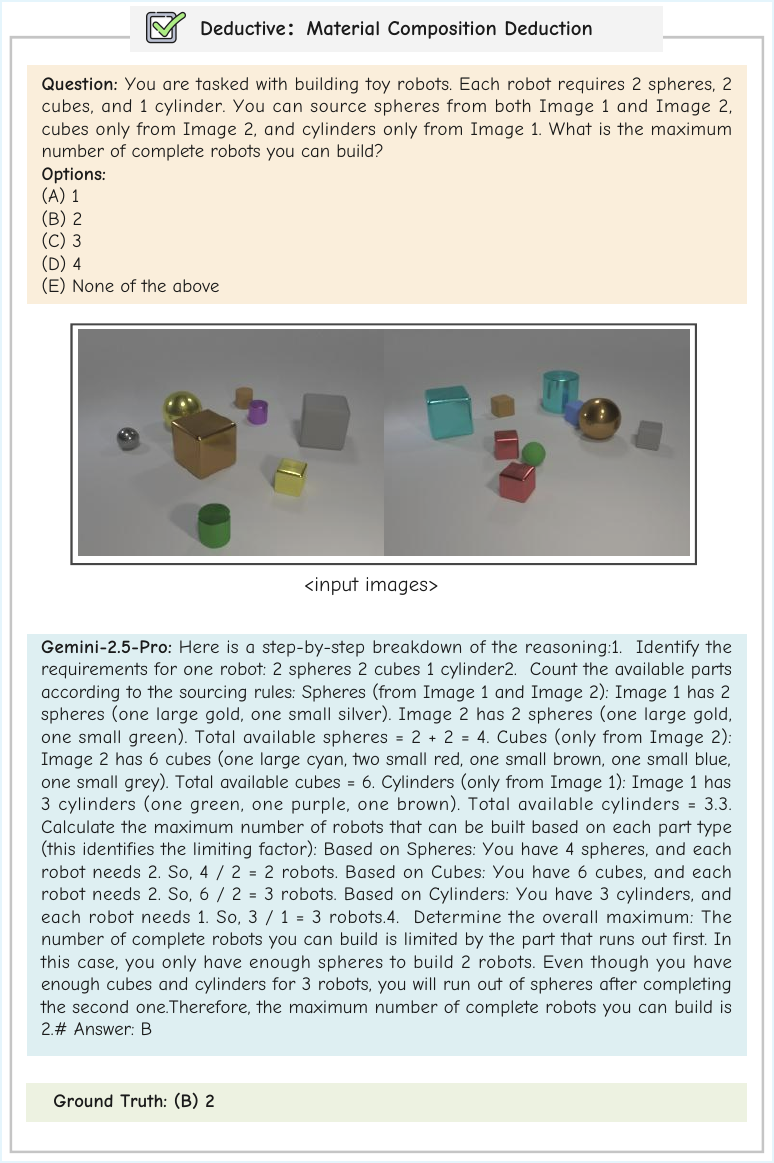}
    \caption{A correct example of Material Composition Deduction task.} \label{mcd_correct}
\end{figure}

\begin{figure}[htbp]
  \centering
    \includegraphics[width=\linewidth]{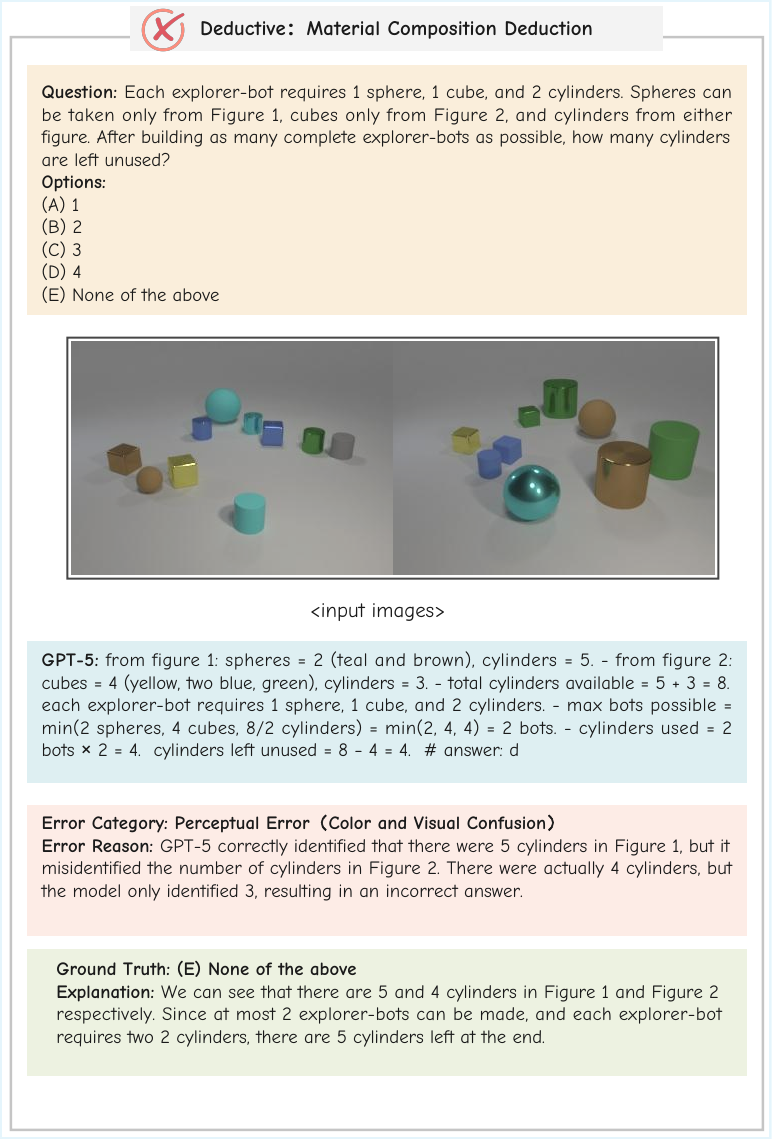}
    \caption{An error example of Material Composition Deduction task.} \label{mcd_error}
\end{figure}

\begin{figure}[htbp]
  \centering
    \includegraphics[width=\linewidth]{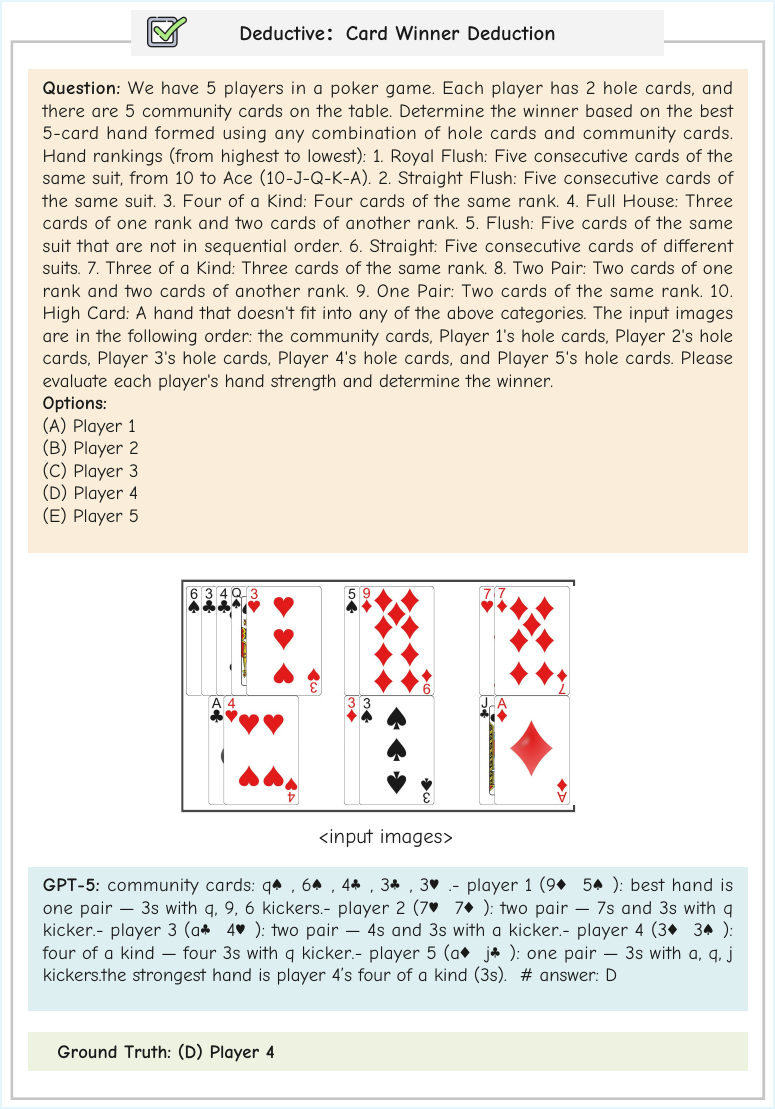}
    \caption{A correct example of Card Winner Deduction task.} \label{cwd_correct}
\end{figure}

\begin{figure}[htbp]
  \centering
    \includegraphics[width=0.9\linewidth]{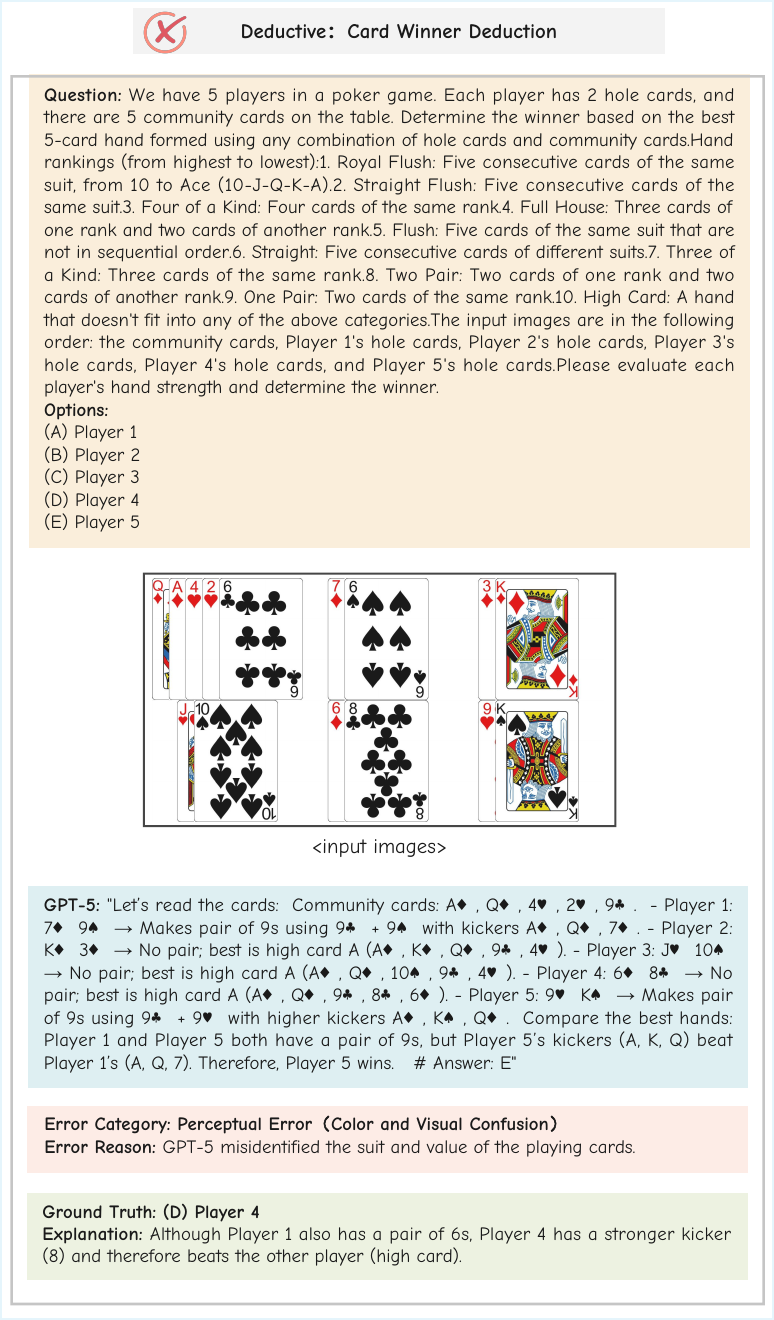}
    \caption{An error example of Card Winner Deduction task.} \label{cwd_error}
\end{figure}

\begin{figure}[htbp]
  \centering
    \includegraphics[width=0.8\linewidth]{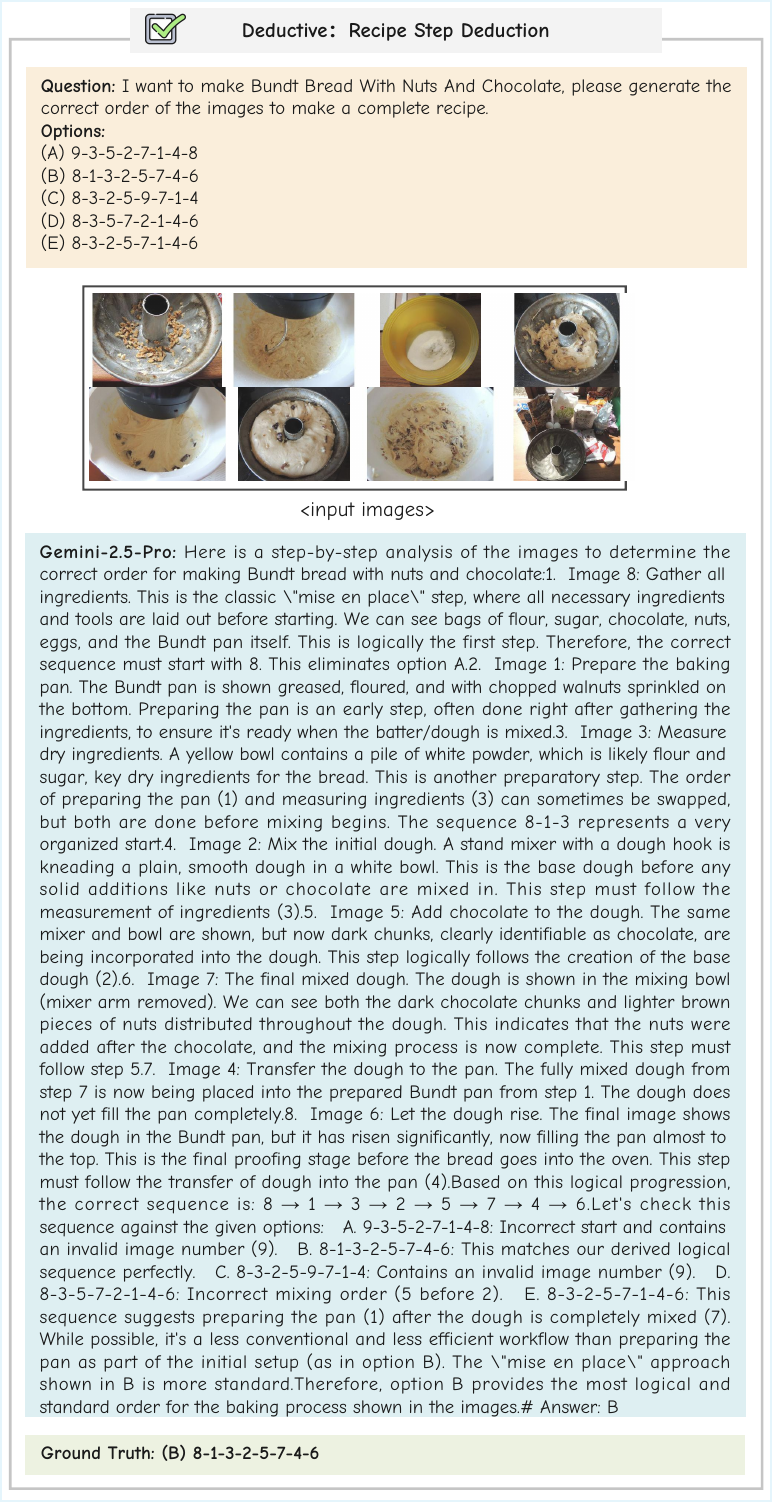}
    \caption{A correct example of Recipe Step Deduction task.} \label{rsd_correct}
\end{figure}

\begin{figure}[htbp]
  \centering
    \includegraphics[width=0.8\linewidth]{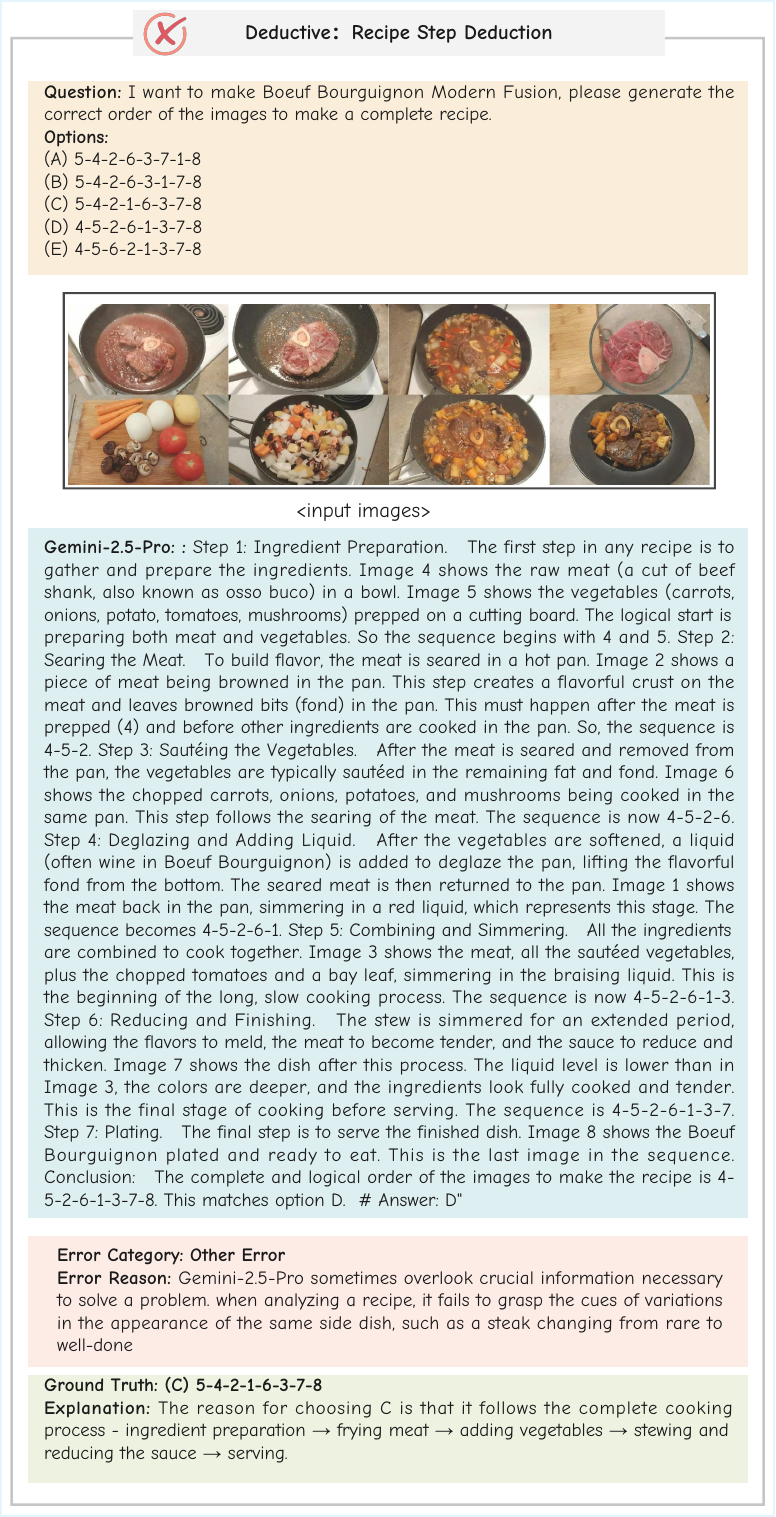}
    \caption{An error example of Recipe Step Deduction task.} \label{rsd_error}
\end{figure}

\begin{figure}[htbp]
  \centering
    \includegraphics[width=0.8\linewidth]{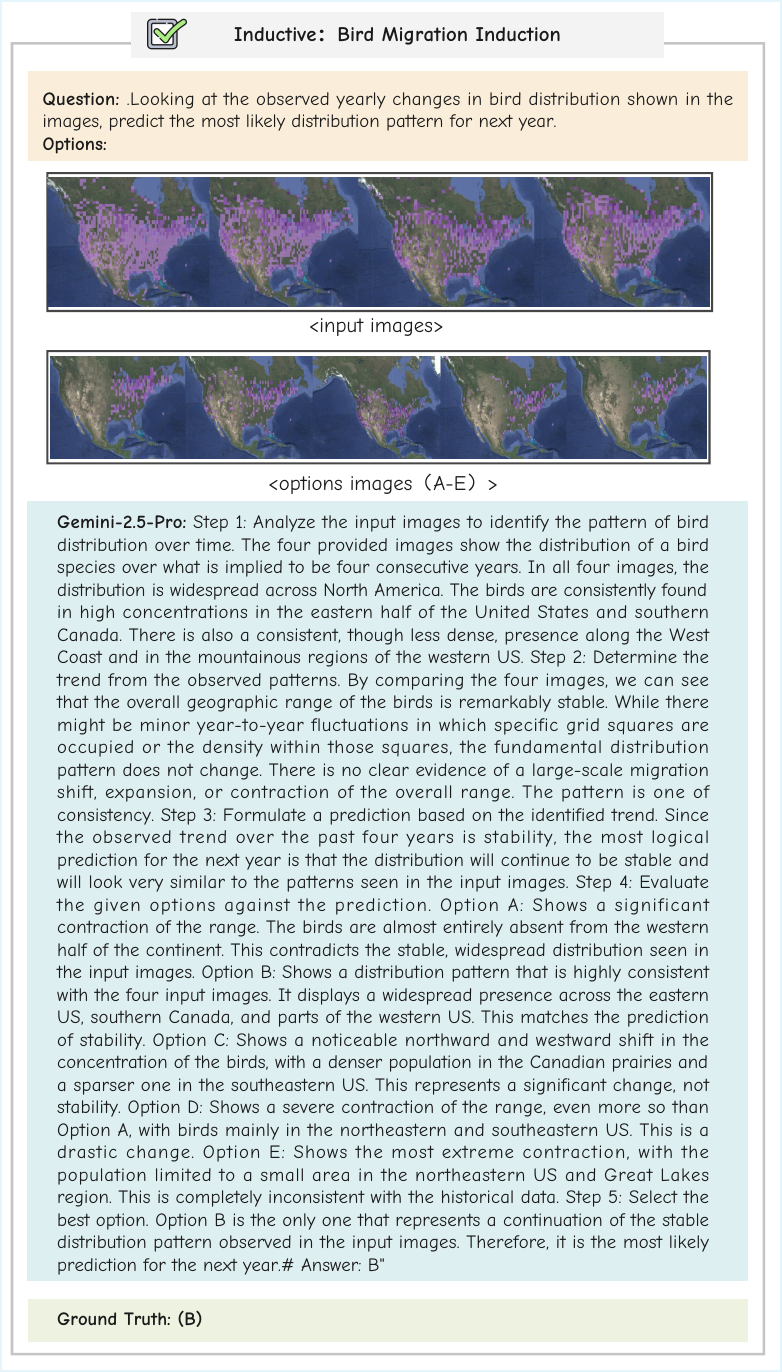}
    \caption{A correct example of Bird Migration Induction task.} \label{bmi_correct}
\end{figure}

\begin{figure}[htbp]
  \centering
    \includegraphics[width=\linewidth]{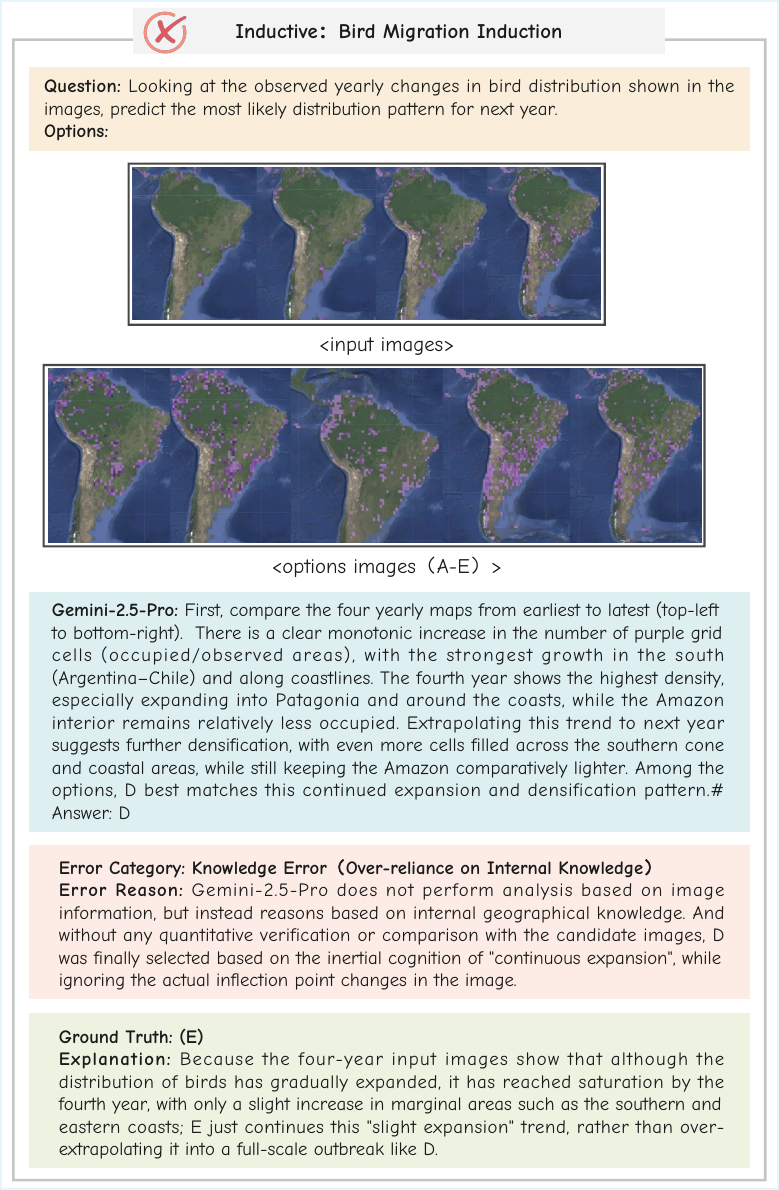}
    \caption{An error example of Bird Migration Induction task.} \label{bmi_error}
\end{figure}

\begin{figure}[htbp]
  \centering
    \includegraphics[width=0.9\linewidth]{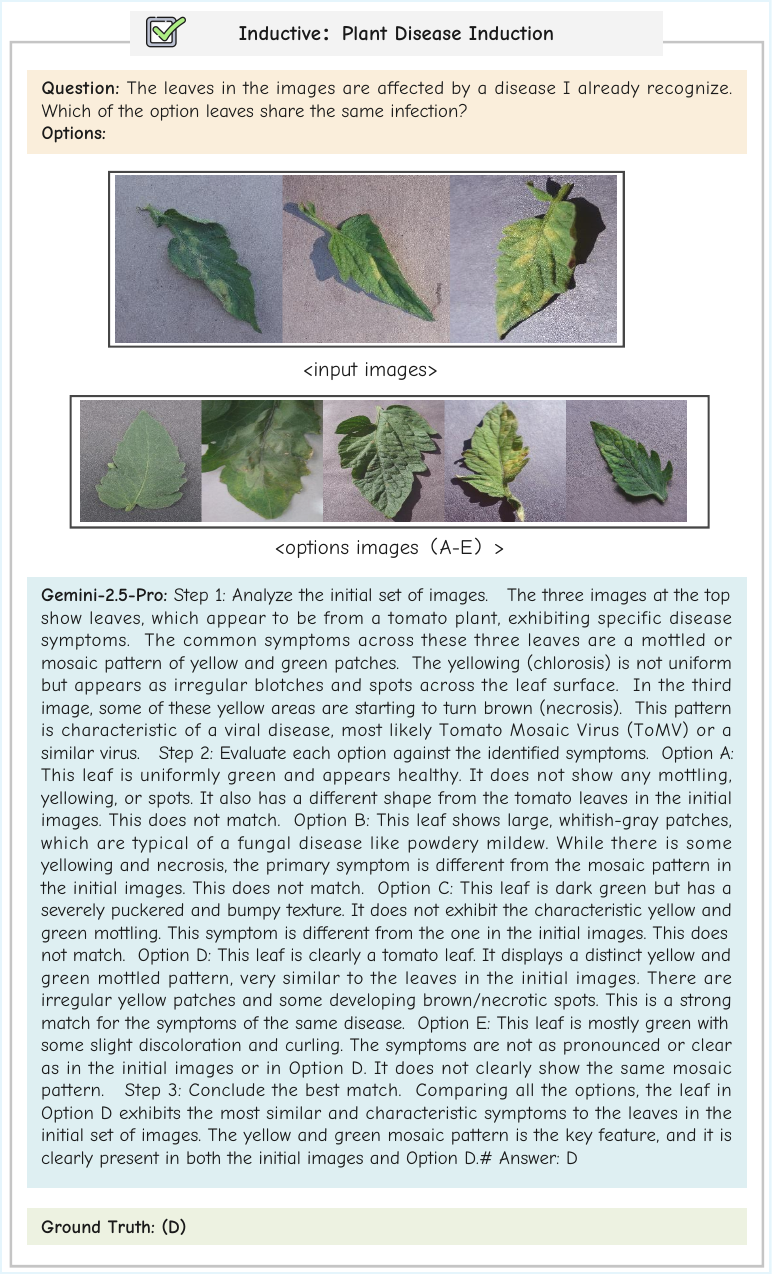}
    \caption{A correct example of Plant Disease Induction task.} \label{pdi_correct}
\end{figure}

\begin{figure}[htbp]
  \centering
    \includegraphics[width=0.9\linewidth]{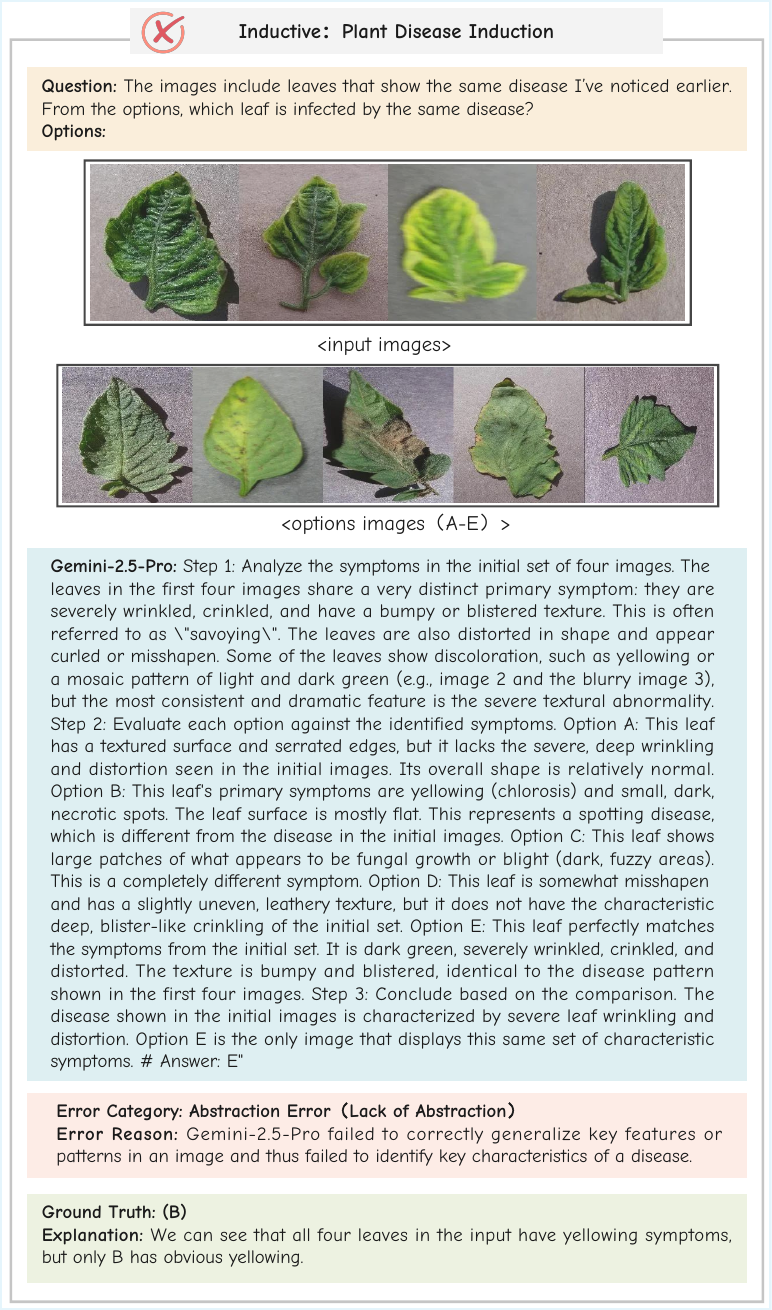}
    \caption{An error example of Plant Disease Induction task.} \label{pdi_error}
\end{figure}

\begin{figure}[htbp]
  \centering
    \includegraphics[width=\linewidth]{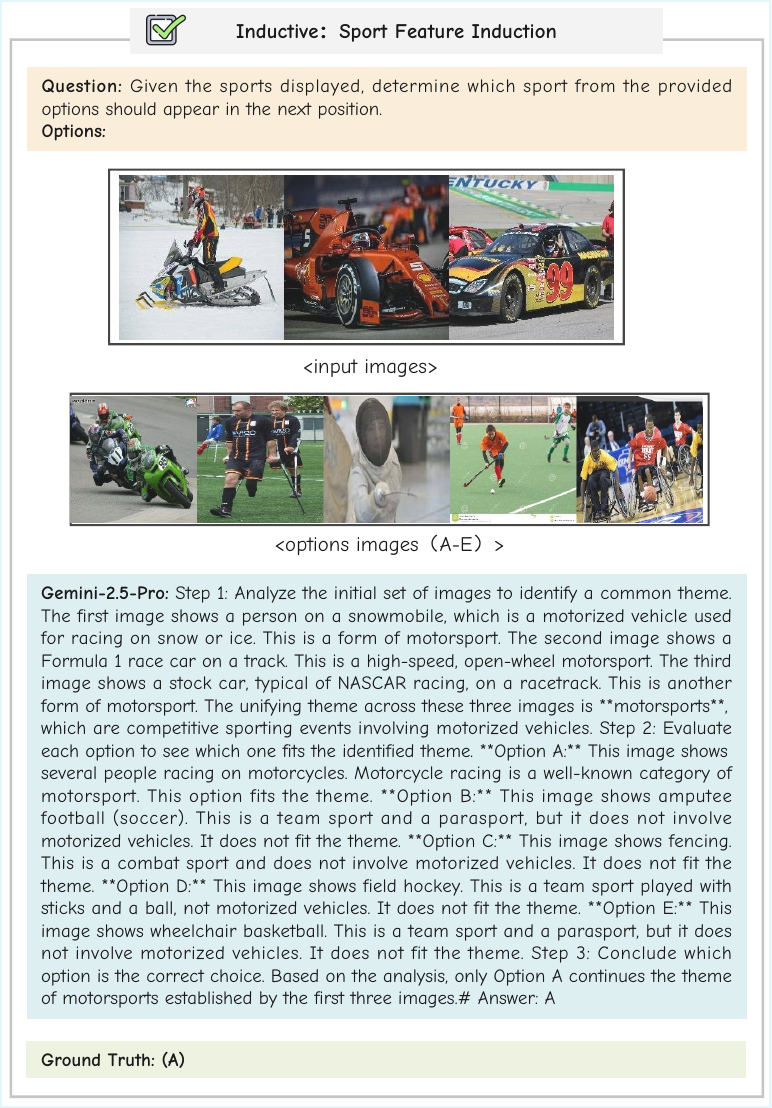}
    \caption{A correct example of Sport Feature Induction task.} \label{sfi_correct}
\end{figure}

\begin{figure}[htbp]
  \centering
    \includegraphics[width=\linewidth]{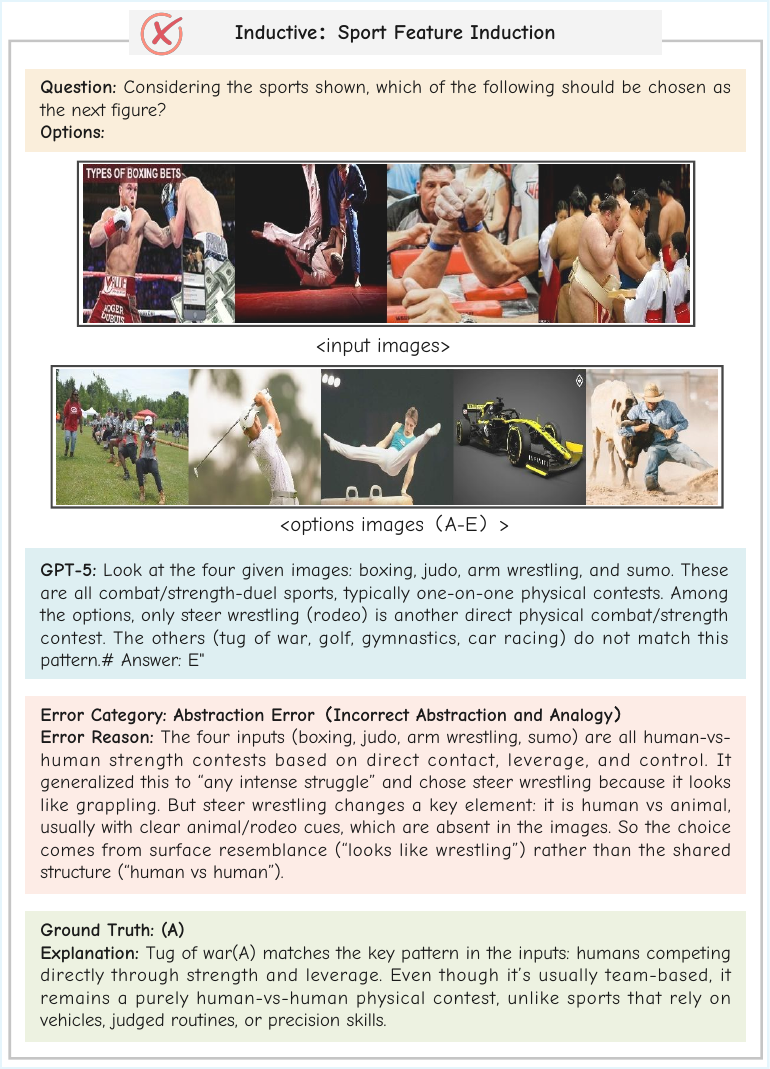}
    \caption{An error example of Sport Feature Induction task.} \label{sfi_error}
\end{figure}

\begin{figure}[htbp]
  \centering
    \includegraphics[width=\linewidth]{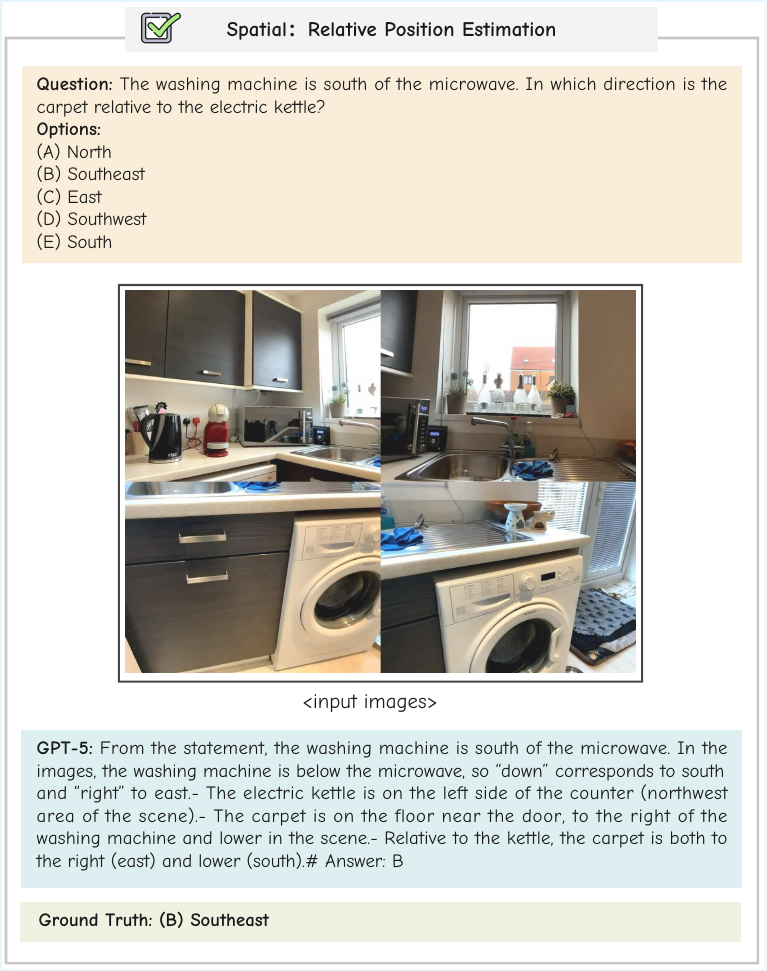}
    \caption{A correct example of Relative Position Estimation task.} \label{rpe_correct}
\end{figure}

\begin{figure}[htbp]
  \centering
    \includegraphics[width=0.95\linewidth]{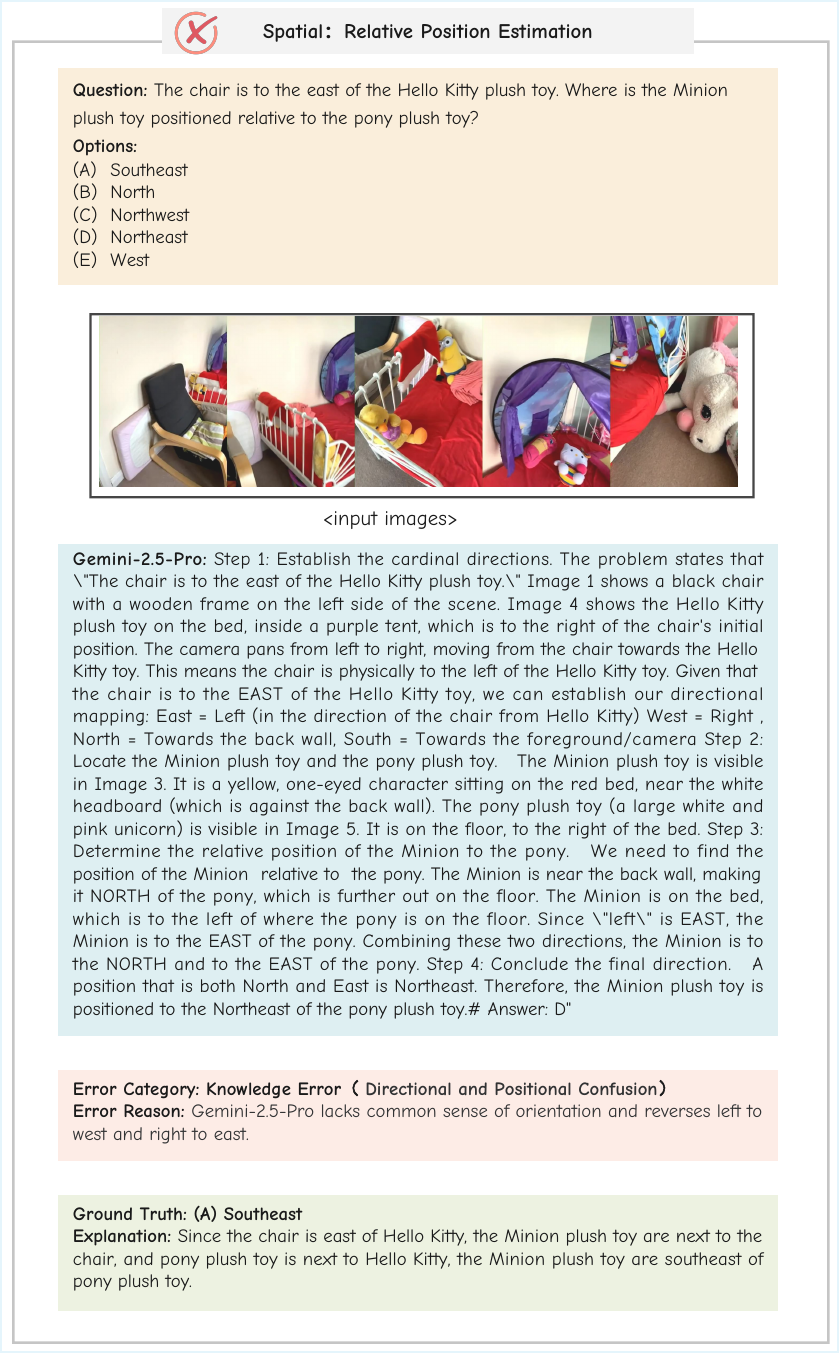}
    \caption{An error example of Relative Position Estimation task.} \label{rpe_error}
\end{figure}

\begin{figure}[htbp]
  \centering
    \includegraphics[width=\linewidth]{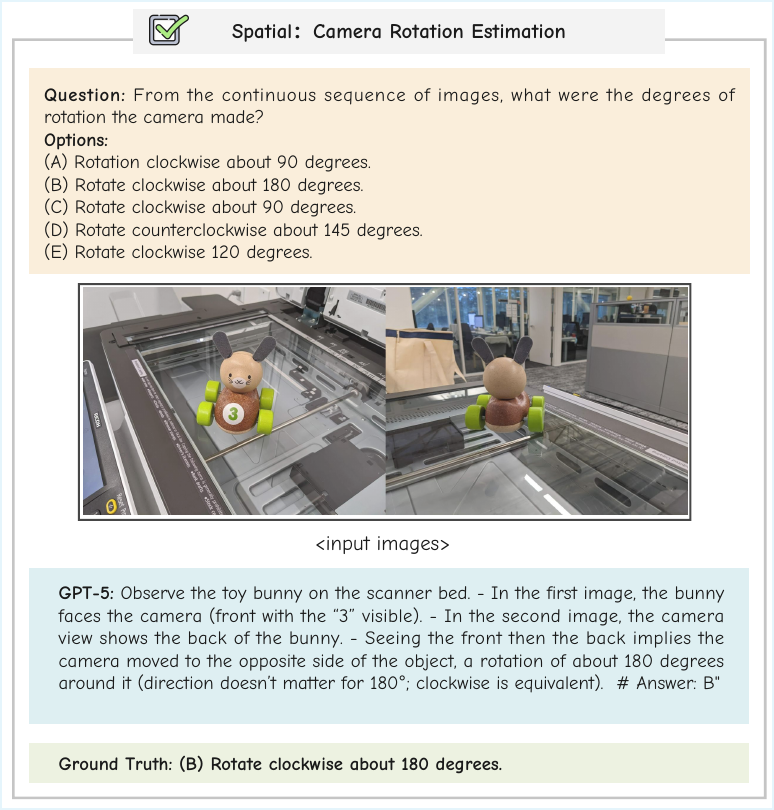}
    \caption{A correct example of Camera Rotation Estimation task.} \label{cre_correct}
\end{figure}

\begin{figure}[htbp]
  \centering
    \includegraphics[width=\linewidth]{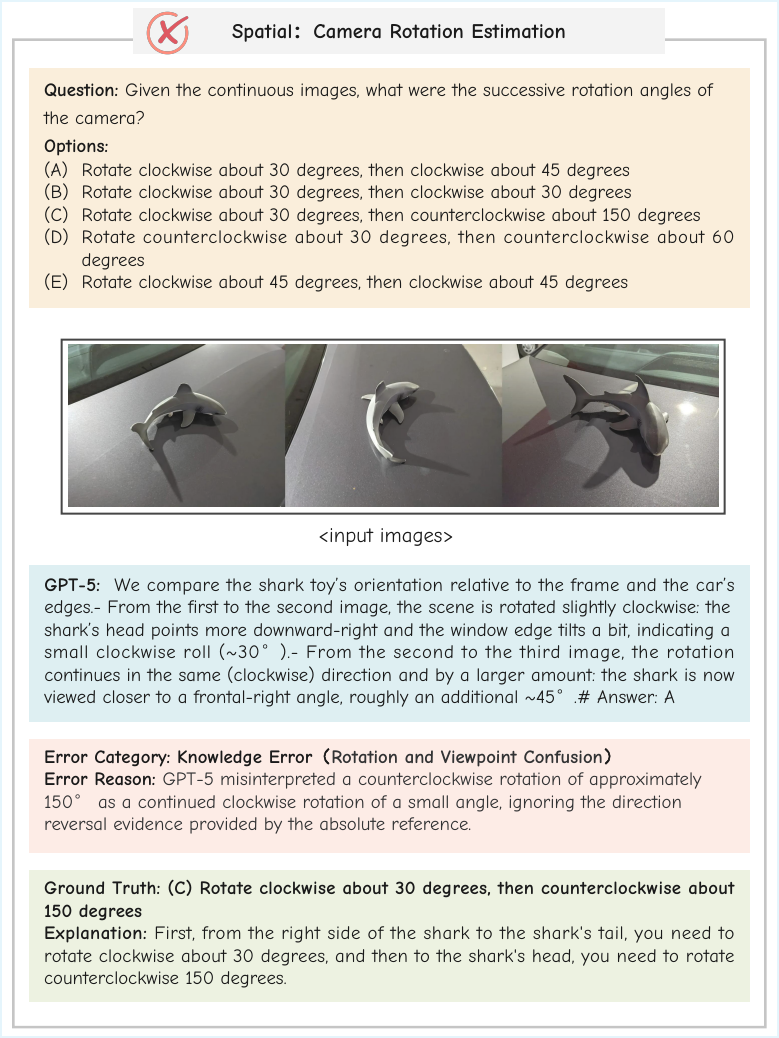}
    \caption{An error example of Camera Rotation Estimation task.} \label{cre_error}
\end{figure}

\begin{figure}[htbp]
  \centering
    \includegraphics[width=\linewidth]{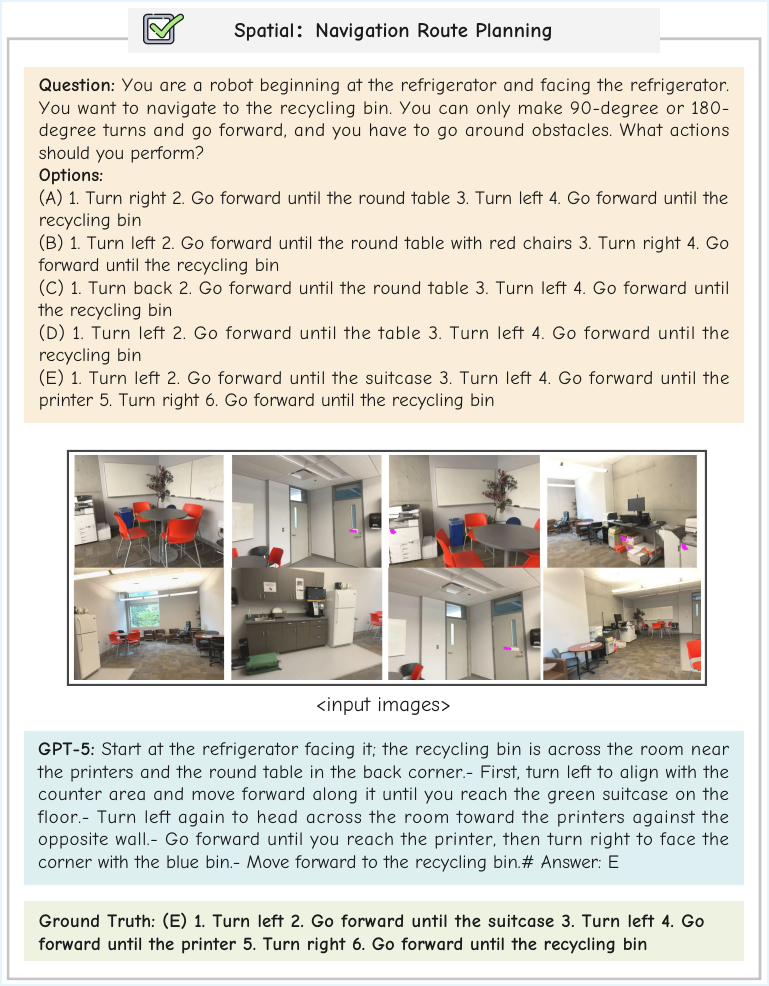}
    \caption{A correct example of Navigation Route Planning task.} \label{nrp_correct}
\end{figure}

\begin{figure}[htbp]
  \centering
    \includegraphics[width=\linewidth]{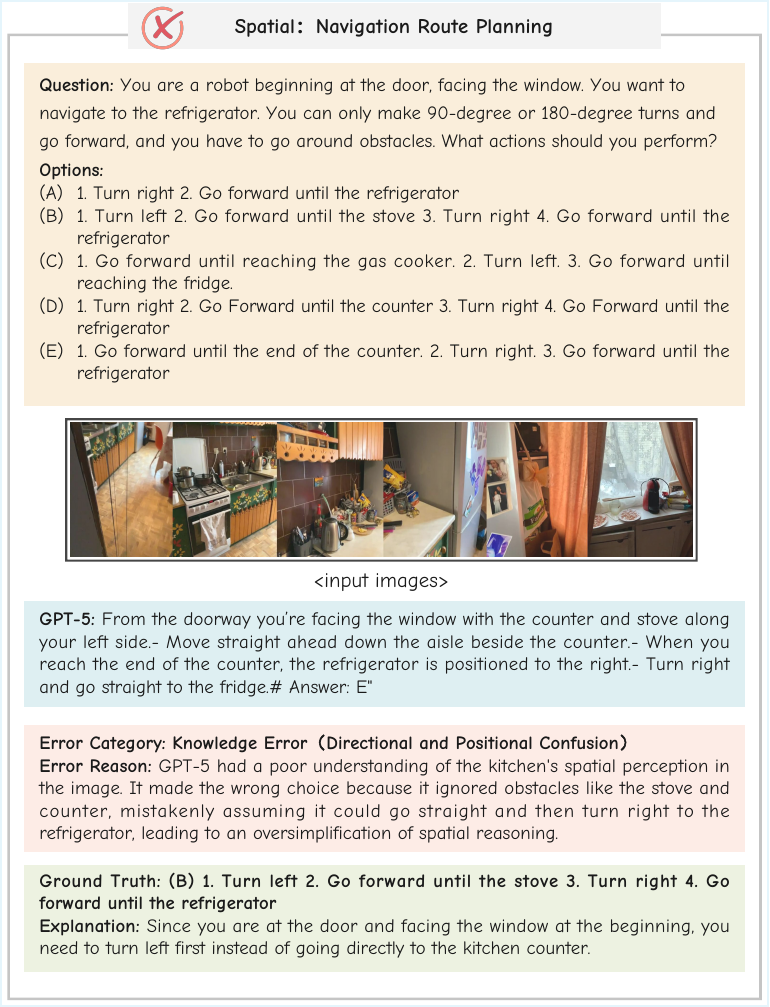}
    \caption{An error example of Navigation Route Planning task.} \label{nrp_error}
\end{figure}

\begin{figure}[htbp]
  \centering
    \includegraphics[width=0.9\linewidth]{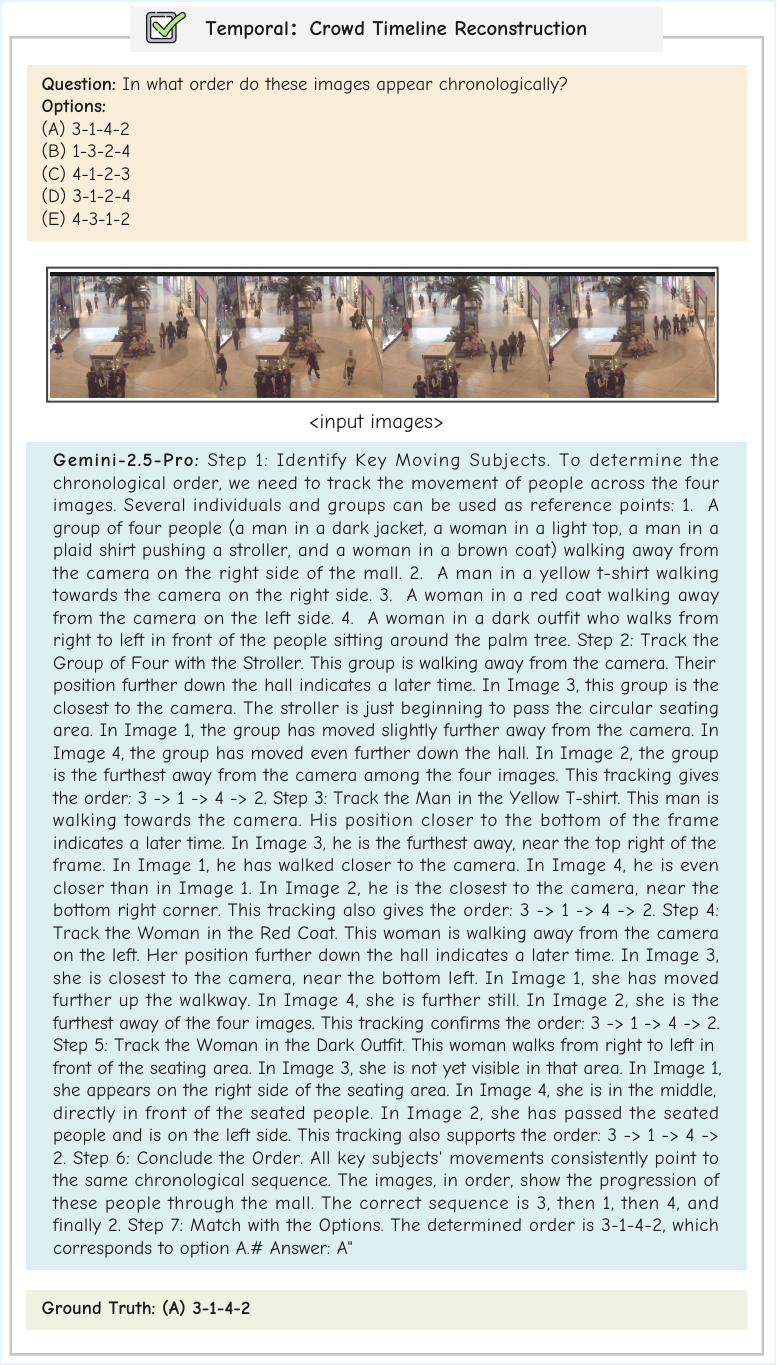}
   \caption{A correct example of Crowd Timeline Reconstruction task.} \label{ctr_correct}
\end{figure}

\begin{figure}[htbp]
  \centering
    \includegraphics[width=0.8\linewidth]{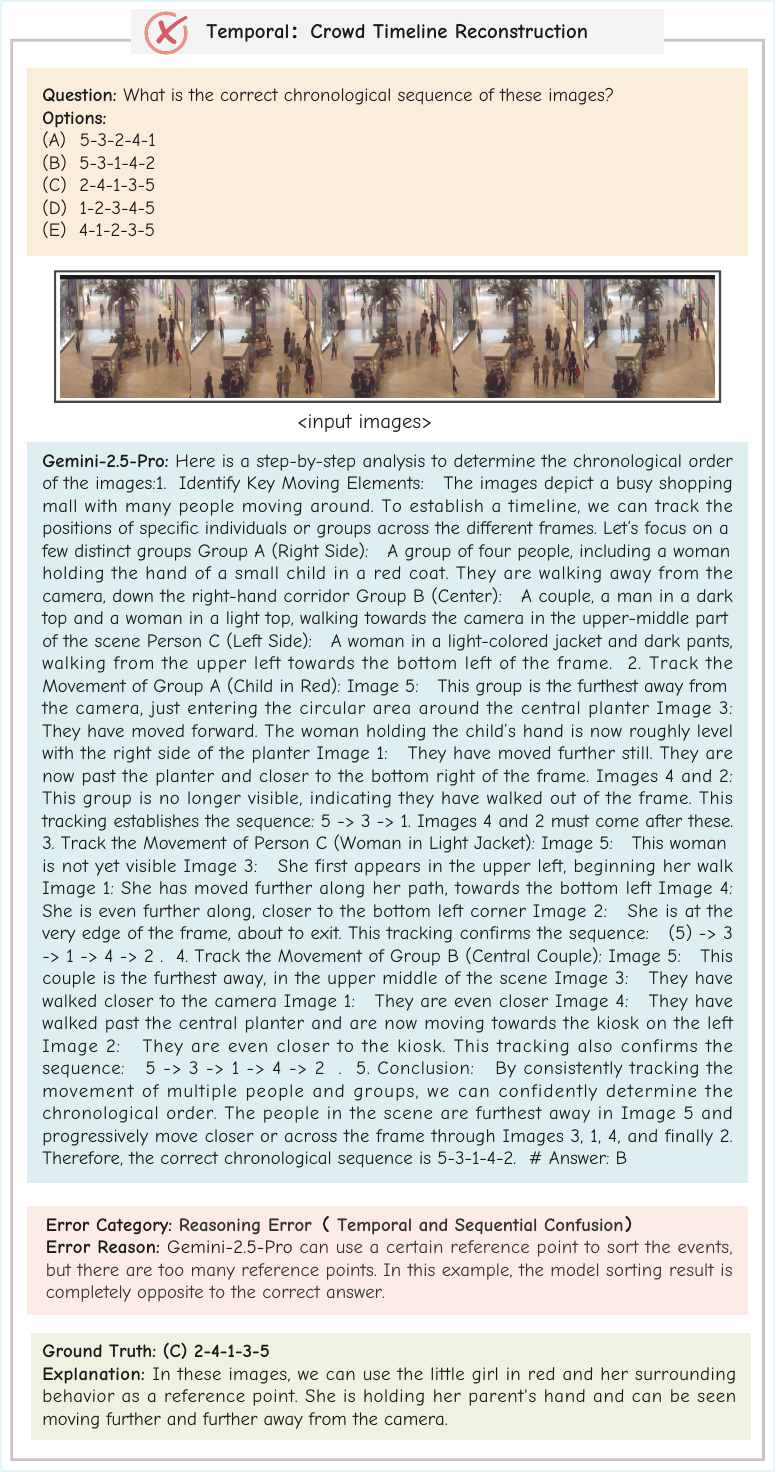}
    \caption{An error example of Crowd Timeline Reconstruction task.} \label{ctr_error}
\end{figure}

\begin{figure}[htbp]
  \centering
    \includegraphics[width=\linewidth]{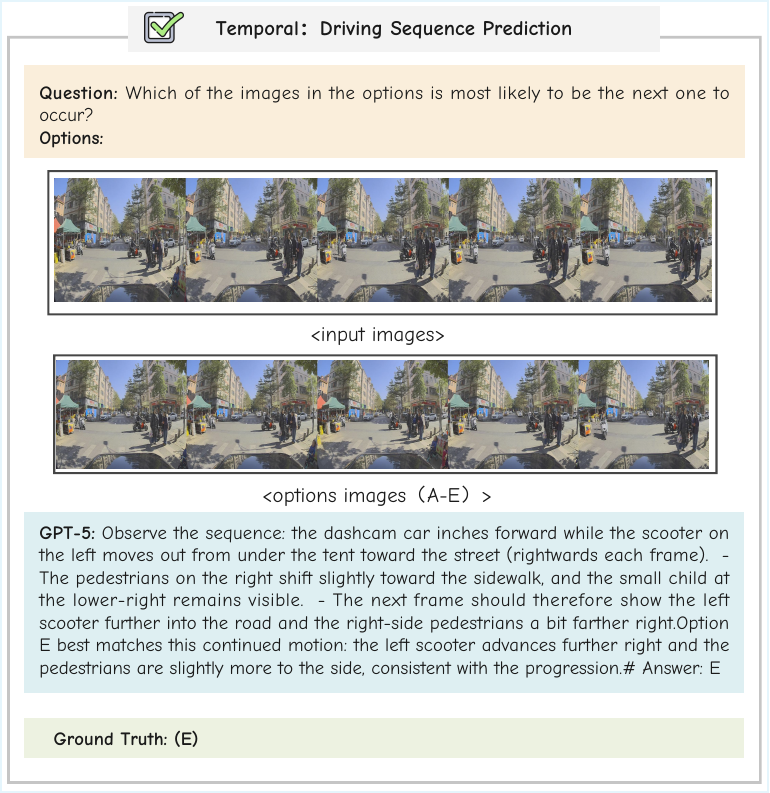}
    \caption{A correct example of Driving Sequence Prediction task.} \label{dsp_correct}
\end{figure}

\begin{figure}[htbp]
  \centering
    \includegraphics[width=\linewidth]{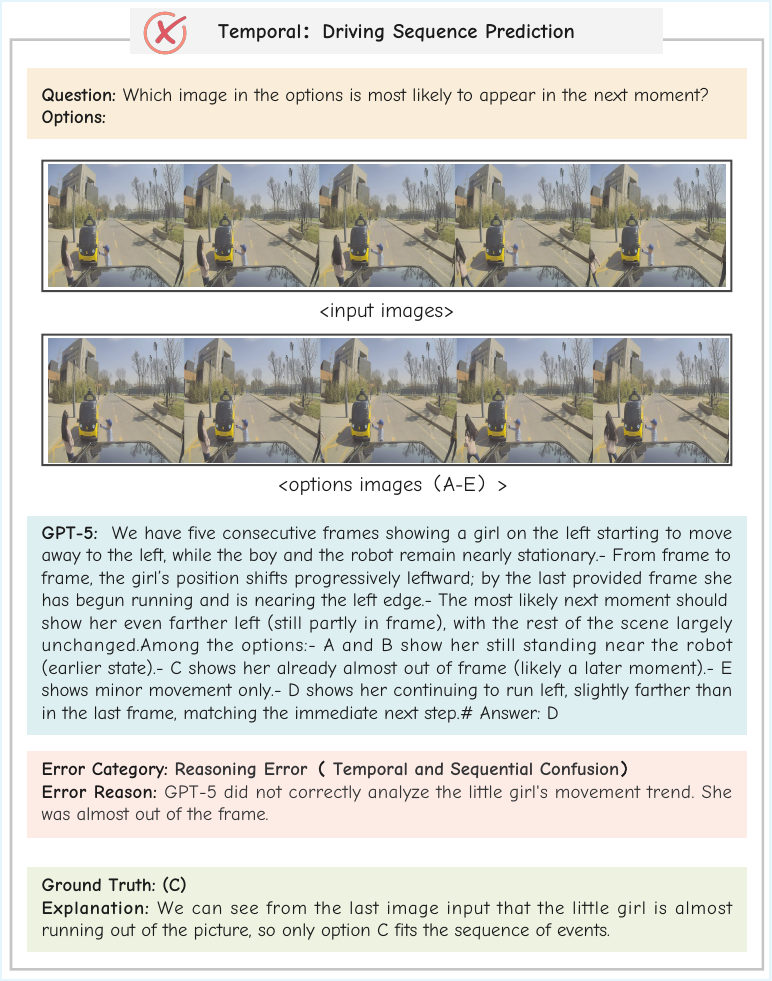}
    \caption{An error example of Driving Sequence Prediction task.} \label{dsp_error}
\end{figure}

\begin{figure}[htbp]
  \centering
    \includegraphics[width=\linewidth]{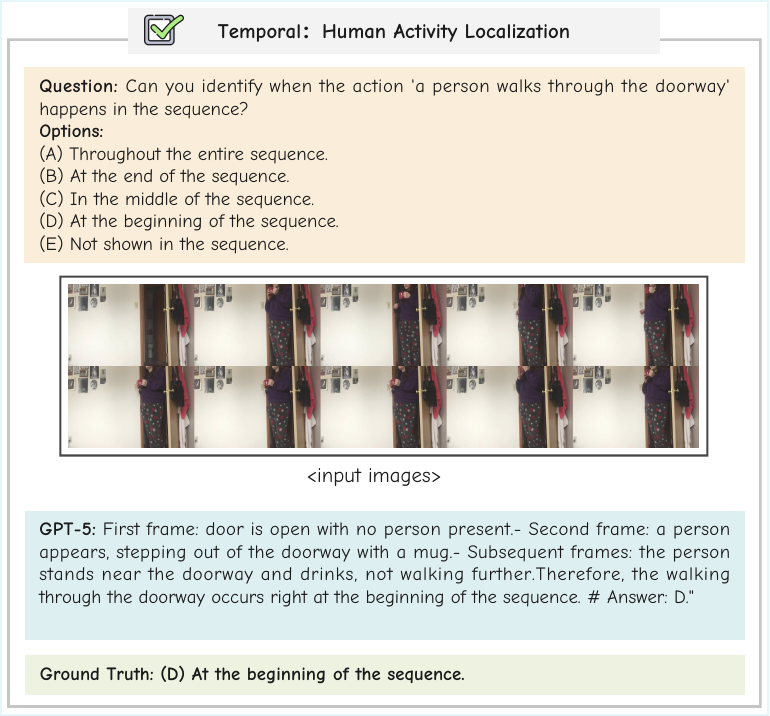}
    \caption{A correct example of Human Activity Localization task.} \label{hal_correct} 
\end{figure}

\begin{figure}[htbp]
  \centering
    \includegraphics[width=\linewidth]{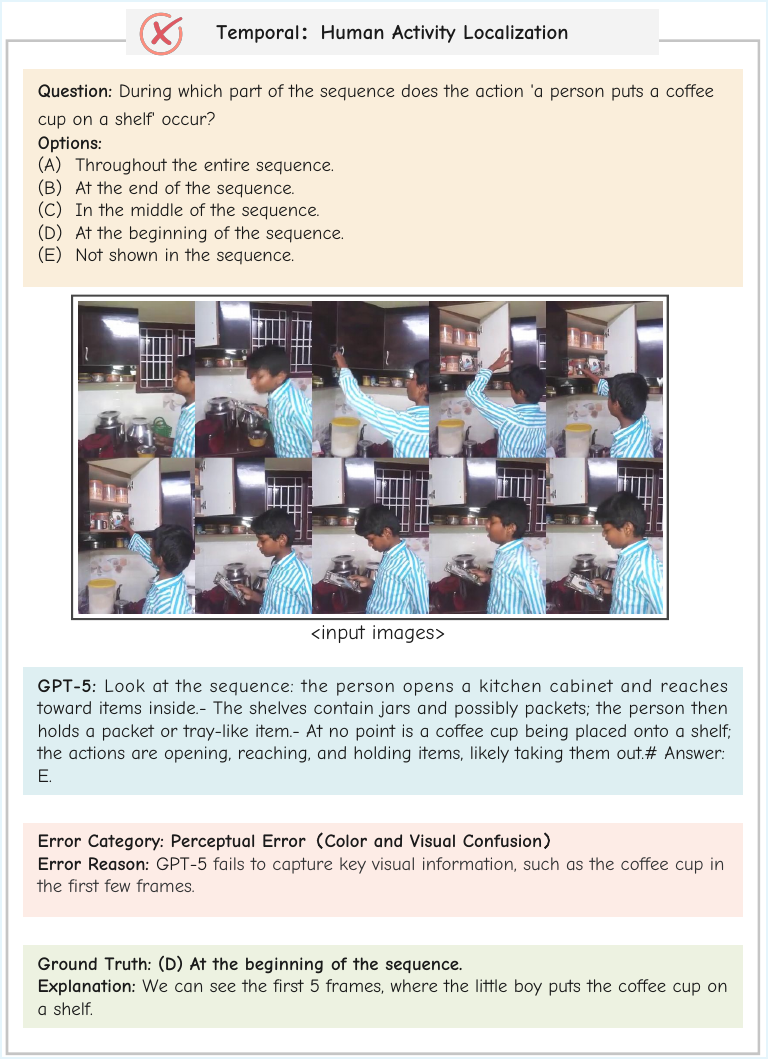}
    \caption{An error example of Human Activity Localization task.} \label{hal_error} 
\end{figure}

We further provide additional case studies as shown from Figure \ref{haa_correct} to Figure \ref{hal_error}, showing both correct and incorrect responses by GPT-5 and Gemini-2.5-Pro.

\begin{table}[h]
\caption{Full performance comparison of SOTA MLLMs on MMR-Life.}
\label{tab:full}
\resizebox{0.99\linewidth}{!}{
\begin{tabular}{lcccccccc}
\toprule
\textbf{Model} & \textbf{Abd} & \textbf{Ana} & \textbf{Cau} & \textbf{Ded} & \textbf{Ind} & \textbf{Spa} & \textbf{Tem} & \textbf{Avg} \\
\midrule
gpt-5 & 53.75 & 78.87 & 41.06 & 80.14 & 78.32 & 17.25 & 41.70 & 58.69 \\
gemini-2.5-pro & 54.40 & 73.77 & 36.99 & 79.43 & 73.66 & 25.10 & 35.79 & 56.86 \\
gemini-2.5-flash & 46.25 & 75.18 & 34.22 & 71.63 & 73.66 & 23.92 & 30.81 & 53.10 \\
o4-mini & 41.37 & 73.59 & 27.38 & 71.28 & 68.07 & 19.22 & 32.66 & 50.49 \\
gpt-5-mini & 44.95 & 69.72 & 32.32 & 75.18 & 68.76 & 12.16 & 29.52 & 49.77 \\
gpt-4.1 & 44.30 & 71.30 & 22.43 & 67.38 & 70.16 & 13.73 & 27.31 & 48.15 \\
claude-sonnet-4-thinking & 36.96 & 60.92 & 44.11 & 67.02 & 56.64 & 15.69 & 28.23 & 45.32 \\
claude-3.7-sonnet & 33.55 & 66.55 & 35.36 & 59.93 & 59.67 & 20.78 & 26.01 & 45.09 \\
gpt-4o & 46.91 & 65.67 & 25.86 & 51.42 & 66.20 & 11.37 & 26.01 & 44.75 \\
gpt-4.1-mini & 32.90 & 61.62 & 30.80 & 52.13 & 65.27 & 16.47 & 30.63 & 44.10 \\
claude-sonnet-4 & 35.50 & 57.22 & 38.02 & 64.89 & 55.71 & 14.51 & 25.83 & 42.82 \\
Qwen2.5-VL-72B & 35.50 & 55.46 & 35.36 & 52.13 & 55.48 & 12.94 & 23.80 & 40.21 \\
doubao-1.5-vision & 37.13 & 53.70 & 31.18 & 59.57 & 54.31 & 12.16 & 23.06 & 39.98 \\
VL-Rethinker-72B & 36.48 & 50.88 & 33.08 & 56.03 & 57.58 & 15.69 & 21.59 & 39.68 \\
Gemma3-27B & 35.18 & 57.92 & 36.88 & 31.21 & 60.61 & 12.94 & 18.27 & 38.32 \\
gpt-4o-mini & 23.45 & 55.28 & 20.15 & 31.21 & 62.00 & 12.55 & 16.42 & 34.54 \\
QVQ-72B-Preview & 31.27 & 41.20 & 38.02 & 47.87 & 31.24 & 14.12 & 16.42 & 31.14 \\
VL-Rethinker-32B & 22.98 & 42.96 & 28.14 & 49.65 & 30.77 & 14.90 & 19.19 & 30.29 \\
MM-Eureka-Qwen-7B & 31.27 & 42.25 & 20.91 & 36.17 & 37.53 & 13.73 & 17.34 & 29.59 \\
Gemma3-12B & 25.08 & 50.70 & 17.11 & 27.30 & 42.42 & 10.20 & 15.87 & 29.52 \\
MM-Eureka-Qwen-32B & 26.06 & 41.02 & 25.10 & 47.52 & 27.97 & 16.08 & 17.34 & 29.02 \\
R1-Onevision-7B & 28.34 & 37.85 & 23.57 & 25.89 & 42.66 & 14.51 & 19.37 & 28.80 \\
MiMo-VL-7B-RL & 38.76 & 25.88 & 28.14 & 60.99 & 24.94 & 14.12 & 19.19 & 28.68 \\
Qwen2.5-VL-32B & 23.45 & 42.78 & 21.29 & 50.00 & 27.27 & 15.69 & 16.24 & 28.61 \\
VL-Rethinker-7B & 30.62 & 40.49 & 21.29 & 28.72 & 43.59 & 13.73 & 11.81 & 28.23 \\
MiMo-VL-7B-SFT & 36.81 & 23.24 & 27.00 & 63.48 & 23.31 & 12.16 & 18.08 & 27.36 \\
Qwen2.5-VL-7B & 26.06 & 35.74 & 20.53 & 20.92 & 38.93 & 9.41 & 12.18 & 24.68 \\
R1-Onevision-7B-RL & 24.76 & 35.04 & 22.05 & 25.53 & 30.30 & 12.16 & 12.73 & 24.00 \\
InternVL3\_5-30B-A3B & 45.60 & 19.19 & 33.46 & 36.52 & 14.45 & 12.16 & 14.39 & 23.09 \\
InternVL3\_5-38B & 46.25 & 15.67 & 26.24 & 41.13 & 5.59 & 14.51 & 18.27 & 21.77 \\
Gemma3-4B & 14.66 & 29.75 & 20.91 & 26.60 & 23.31 & 12.16 & 17.34 & 21.50 \\
Keye-VL-1.5-8B & 19.87 & 21.30 & 23.95 & 14.18 & 20.28 & 13.73 & 23.62 & 20.22 \\
InternVL3\_5-8B & 35.18 & 11.44 & 18.63 & 34.04 & 11.19 & 14.90 & 16.61 & 18.67 \\
Kimi-VL-A3B-Thinking-2506 & 24.10 & 12.85 & 18.25 & 37.23 & 10.96 & 12.94 & 18.08 & 18.07 \\
OpenVLThinker-7B-v1.2 & 16.94 & 19.37 & 20.91 & 12.06 & 18.18 & 17.25 & 18.27 & 17.84 \\
OpenVLThinker-7B-v1.2-sft & 16.29 & 19.19 & 21.29 & 13.83 & 18.41 & 17.65 & 17.16 & 17.80 \\
Skywork-R1V-38B & 22.15 & 10.39 & 16.73 & 23.76 & 11.89 & 9.80 & 11.07 & 14.13 \\
\bottomrule
\end{tabular}}
\end{table}

\begin{table}[h]
\caption{Performance comparison of SOTA MLLMs on MMR-Life mini set.}
\label{tab:mini}
\resizebox{0.99\linewidth}{!}{
\begin{tabular}{lcccccccc}
\toprule
\textbf{Model} & \textbf{Abd} & \textbf{Ana} & \textbf{Cau} & \textbf{Ded} & \textbf{Ind} & \textbf{Spa} & \textbf{Tem} & \textbf{Avg} \\
\midrule
Human & 79.76 & 57.65 & 75.00 & 70.59 & 63.41 & 79.76 & 79.76 & 72.28 \\
gpt-5 & 63.33 & 66.67 & 53.33 & 73.33 & 66.67 & 30.00 & 50.00 & 57.62 \\
gemini-2.5-pro & 63.33 & 80.00 & 60.00 & 76.67 & 50.00 & 30.00 & 36.67 & 56.67 \\
gemini-2.5-flash & 53.33 & 63.33 & 50.00 & 53.33 & 50.00 & 50.00 & 53.33 & 53.33 \\
o4-mini & 46.67 & 73.33 & 43.33 & 73.33 & 50.00 & 23.33 & 43.33 & 50.48 \\
gpt-5-mini & 50.00 & 56.67 & 40.00 & 83.33 & 50.00 & 6.67 & 33.33 & 45.71 \\
claude-sonnet-4 & 40.00 & 66.67 & 60.00 & 66.67 & 30.00 & 16.67 & 56.67 & 48.10 \\
gpt-4.1 & 50.00 & 56.67 & 36.67 & 16.67 & 63.33 & 6.67 & 26.67 & 36.67 \\
claude-3.7-sonnet & 30.00 & 73.33 & 46.67 & 70.00 & 40.00 & 13.33 & 43.33 & 45.24 \\
gpt-4o & 43.33 & 50.00 & 40.00 & 16.67 & 46.67 & 20.00 & 26.67 & 34.76 \\
gpt-4.1-mini & 33.33 & 63.33 & 36.67 & 53.33 & 50.00 & 10.00 & 26.67 & 39.05 \\
doubao-1.5-vision & 50.00 & 50.00 & 36.67 & 63.33 & 40.00 & 13.33 & 23.33 & 39.52 \\
VL-Rethinker-72B & 26.67 & 50.00 & 40.00 & 63.33 & 33.33 & 13.33 & 23.33 & 35.71 \\
QVQ-72B-Preview & 26.67 & 53.33 & 50.00 & 36.67 & 26.67 & 3.33 & 16.67 & 30.48 \\
MM-Eureka-Qwen-32B & 20.00 & 46.67 & 43.33 & 50.00 & 33.33 & 16.67 & 36.67 & 35.24 \\
MiMo-VL-7B-RL & 36.67 & 16.67 & 40.00 & 70.00 & 16.67 & 13.33 & 26.67 & 31.43 \\
VL-Rethinker-7B & 23.33 & 36.67 & 13.33 & 13.33 & 33.33 & 0.00 & 16.67 & 22.78 \\
Keye-VL-1.5-8B & 13.33 & 23.33 & 33.33 & 20.00 & 26.67 & 13.33 & 26.67 & 22.38 \\
Skywork-R1V-38B & 30.00 & 6.67 & 33.33 & 16.67 & 30.00 & 16.67 & 13.33 & 20.95 \\
Qwen2.5-VL-72B & 36.67 & 40.00 & 40.00 & 20.00 & 36.67 & 10.00 & 30.00 & 30.48 \\
Gemma3-27B & 20.00 & 36.67 & 33.33 & 30.00 & 40.00 & 6.67 & 30.00 & 28.10 \\
Gemma3-12B & 30.00 & 50.00 & 16.67 & 33.33 & 20.00 & 6.67 & 23.33 & 25.71 \\
Qwen2.5-VL-32B & 23.33 & 30.00 & 30.00 & 20.00 & 20.00 & 6.67 & 23.33 & 21.90 \\
Qwen2.5-VL-7B & 13.33 & 0.00 & 23.33 & 20.00 & 0.00 & 6.67 & 10.00 & 14.67 \\
InternVL3.5-30B-A3B & 33.33 & 13.33 & 40.00 & 26.67 & 16.67 & 10.00 & 16.67 & 22.38 \\
InternVL3.5-8B & 30.00 & 6.67 & 20.00 & 16.67 & 20.00 & 13.33 & 10.00 & 16.67 \\
\bottomrule
\end{tabular}}
\end{table}

\clearpage


\clearpage





\end{document}